\PassOptionsToPackage{most,breakable}{tcolorbox}
\PassOptionsToPackage{dvipsnames,table}{xcolor}
\PassOptionsToPackage{numbers,compress}{natbib}
\documentclass[11pt]{article}

\usepackage[final]{ropedia}
\usepackage[T1]{fontenc}
\usepackage[utf8]{inputenc}
\usepackage{amsmath,amssymb}
\usepackage{amsfonts}
\usepackage{mathpazo}
\usepackage[scaled]{helvet}
\usepackage{pifont}
\usepackage{booktabs}
\usepackage{multirow}
\usepackage{tabularx}
\usepackage{subcaption}
\usepackage[export]{adjustbox}
\usepackage{makecell}
\usepackage{enumitem}
\usepackage{titletoc}
\usepackage{nicefrac}
\usepackage{microtype}
\usepackage{graphicx}
\usepackage{capt-of}
\usepackage{xspace}
\usepackage{xurl}
\usepackage{placeins}
\setlist[itemize]{leftmargin=2em}
\newcommand{\projecttitle}{Ropedia~W0\xspace}
\newcommand{\projectsubtitle}{Scaling Loco-Manipulation World Models \\ across Embodiments with Human Experience}
\newcommand{\method}{Ropedia~W0\xspace}
\newcommand{\datasetname}{Xp-Data\xspace}
\newcommand{\teleopname}{Xp-Teleop\xspace}
\newcommand{\coremark}{\ensuremath{^{*}}}
\newcommand{\corrmark}{\ensuremath{^{\dagger}}}

\newcommand{\rankone}[1]{\textbf{#1}}
\newcommand{\ranktwo}[1]{\underline{#1}}
\newcommand{\rankthree}[1]{#1}
\newcommand{\papertablestyle}{%
  \fontencoding{T1}\fontfamily{ptm}\selectfont
  \fontsize{8}{10}\selectfont
  \setlength{\tabcolsep}{2pt}%
  \renewcommand{\arraystretch}{1.1}%
}


\firstpagelogos{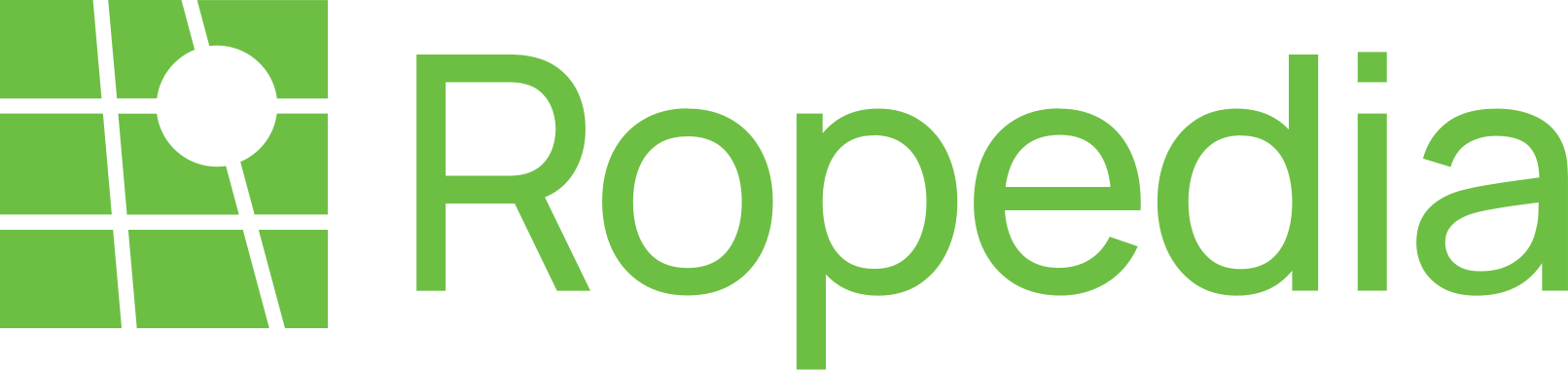}{}{}
\abstractlogo{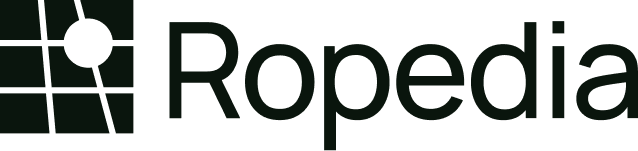}
\title{\projecttitle}
\subtitle{\projectsubtitle}

\newcommand{\authorfootnotes}{%
  \begingroup
  \renewcommand{\thefootnote}{}%
  \footnotetext{$*$ Equal contribution.\quad $\dagger$ Corresponding author.}%
  \endgroup
}

\author{
  \AuthorName{Tianqi Liu}{1,2,\coremark}\quad
  \AuthorName{Hanfei Zhao}{2,3,\coremark}\quad
  \AuthorName{Qifeng Liang}{2,4,\coremark}\quad
  \AuthorName{Hui En Pang}{1,2}\quad
  \AuthorName{Jiarui Xu}{1,2}\\[0.55ex]
  \AuthorName{Sayang Mu}{1,2}\quad
  \AuthorName{Penghao Qiu}{1}\quad
  \AuthorName{YingSheng Zhang}{5}\quad
  \AuthorName{Jun Wang}{2}\quad
  \AuthorName{Ziang Cao}{1,2}\\[0.55ex]
  \AuthorName{Mingze Sun}{2,5}\quad
  \AuthorName{Yukang Cao}{1}\quad
  \AuthorName{Fangzhou Hong}{1,2}\quad
  \AuthorName{Zhaoxi Chen}{1,2}\quad
  \AuthorName{Ziwei Liu}{1,2,\corrmark}\\[1.0ex]
  {\normalfont\footnotesize
  \mbox{\textsuperscript{1}Nanyang Technological University}\quad
  \mbox{\textsuperscript{2}Ropedia}\quad
  \mbox{\textsuperscript{3}Duke University}\\
  \mbox{\textsuperscript{4}University of California, San Diego}\quad
  \mbox{\textsuperscript{5}Tsinghua University}}
}

\begin{document}
\maketitle
\authorfootnotes

\vspace{2pt}
{\centering
\href{https://ropedia.github.io/ropedia-w0/}{\ttfamily\bfseries\color{ropediadeepgreen}https://ropedia.github.io/ropedia-w0/}\par}

\vspace{7pt}
{\centering
\includegraphics[width=1\textwidth]{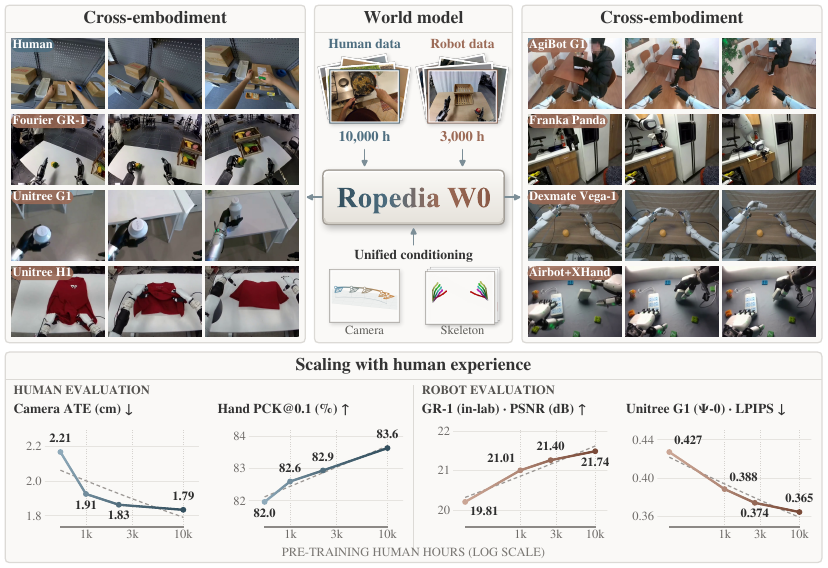}
\par\vspace{6pt}
\begin{minipage}{\textwidth}
\footnotesize
\refstepcounter{figure}\label{fig:teaser}
Figure 1 | \textbf{\method{}: a unified egocentric world model for cross-embodiment loco-manipulation.} Trained on \datasetname{} (14 human and 13 robot public datasets plus in-house recordings; ${\sim}$10k\,h human and ${\sim}$3k\,h robot video), a single shared interface of camera poses and projected kinematic skeletons (\emph{center}) lets one model predict future egocentric video across humans, gripper robots, and dexterous humanoids (\emph{left and right}). Prediction quality on both human and robot evaluations improves as the amount of human-video pre-training is scaled up (\emph{bottom}).
\end{minipage}
\par}

\vspace{8pt}
\begin{abstract}
World models enable embodied agents to anticipate action consequences, yet jointly modeling locomotion and manipulation across diverse embodiments remains challenging due to heterogeneous action spaces and limited multimodal supervision. We present \textbf{\method}, the first egocentric loco-manipulation world foundation model spanning humans and robots with diverse embodiments. We curate \textbf{\datasetname}, comprising 10k hours of egocentric human data and 3k hours of robot data with unified multimodal annotations and systematic quality control. A unified kinematic representation encodes locomotion through camera poses and manipulation through projected skeletons, with robot skeletons combining embodiment-specific forward kinematics with extensive manual keypoint annotation and correction. We also introduce \textbf{\teleopname}, a long-horizon loco-manipulation benchmark based on high-quality dexterous humanoid teleoperation data. Evaluations demonstrate state-of-the-art performance across five cross-embodiment benchmarks (DreamDojo, $\Psi_0$, $\omega$-HOME, \teleopname{}, and AgiBot) and on held-out human egocentric video, from zero-shot inference to few-shot and full-data post-training. Scaling human pre-training data improves performance on both human and robot evaluations. After distillation, our model supports real-time, minute-long rollouts, enabling live VR interaction. Applications in policy evaluation, model-based planning, and world--action modeling further demonstrate the model's utility for embodied decision-making.

\end{abstract}

\section{Introduction}
\label{sec:intro}
World models predict how environments evolve in response to actions~\citep{ha2018worldmodels,bruce2024genie} and support decision-making by imagining future outcomes~\citep{hafner2025dreamer}, giving them considerable potential in embodied intelligence. However, most existing models~\citep{wu2026oscar,xu2026kinema4d,alzayer2026masked} focus on fixed-base, gripper-based manipulation, leaving locomotion-induced viewpoint changes and dexterous hand--object interactions largely underexplored. Moreover, generalizing across embodiments poses a further challenge. DreamDojo~\citep{gao2026dreamdojo} replaces latent-action conditioning with robot-specific actions during post-training, and Cosmos3~\citep{nvidia2026cosmos} relies on domain-specific action projections despite a shared backbone; both therefore require a dedicated control interface or additional adaptation for each new embodiment. Therefore, jointly modeling locomotion and manipulation while seamlessly generalizing across diverse embodiments remains an open problem.

Learning a generalizable world model hinges on large-scale, high-quality egocentric video paired with rich multimodal annotations, such as camera trajectories, hand motion, and task descriptions. While robot demonstrations across diverse tasks and embodiments remain expensive to collect, egocentric human video offers an abundant and complementary source of experience on how humans navigate and interact with their surroundings. However, existing public egocentric datasets~\citep{grauman2021ego4d,grauman2023egoexo4d,hoque2025egodex,wang2023holoassist,damen2020rescaling,buildaiegocentric10k2025,liu2022hoi4d} frequently suffer from incomplete or unreliable annotations. Specifically, many lack metric camera trajectories, provide noisy or missing hand poses, and adopt heterogeneous formats that hinder joint training. Furthermore, the scale of these datasets remains limited, falling short of the data required to train a generalizable world model.

To address these challenges, we present \textbf{\method{}}, a unified egocentric world model that scales learning from human and robot videos for cross-embodiment loco-manipulation. We curate \textbf{\datasetname{}}, comprising 10k hours of egocentric human data and 3k hours of robot data, with unified geometric, kinematic, and semantic annotations produced through automated processing, manual robot-keypoint annotation, and quality checks. To connect human motion and robot actions, we introduce a shared explicit kinematic representation that combines metric camera poses for locomotion-induced viewpoint changes with projected skeletons for manipulation. The skeletons combine human pose annotations with robot forward kinematics (FK) and manually annotated or corrected robot keypoints, preserving embodiment-specific structure in an image-aligned format. This conditioning interface remains consistent across human pre-training, human--robot mid-training, and optional post-training. We further introduce \textbf{\teleopname{}}, a humanoid teleoperation benchmark for evaluating long-horizon prediction during locomotion and dexterous manipulation.

We evaluate \method{} on five cross-embodiment benchmarks (DreamDojo, $\Psi_0$, $\omega$-HOME, \teleopname{}, and AgiBot), alongside held-out human egocentric video. Results demonstrate strong predictive performance across embodiments and post-training budgets, including zero-shot transfer and data-efficient adaptation. Scaling analyses show that increasing the amount of human pre-training data improves both prediction on human videos and transfer to robots. Ablations establish the complementary roles of camera and skeleton conditioning, while distillation enables real-time, minute-long rollouts and live VR interaction across embodiments. Downstream experiments further demonstrate accurate policy ranking, improved trajectory planning, and more robust action learning in simulation and on real robots.

Our contributions are threefold:
\begin{itemize}[topsep=0pt,itemsep=6pt,parsep=0pt,leftmargin=1.4em]
    \item We introduce \textbf{\method{}}, which, to the best of our knowledge, is the first unified egocentric world model for cross-embodiment loco-manipulation. A shared explicit kinematic representation of camera poses and projected skeletons enables scalable learning from human and robot videos.
    \item We design a human-in-the-loop pipeline that unifies collection, annotation, and quality checks, including extensive manual robot-keypoint annotation and correction of inaccurate FK projections. Consequently, we build \textbf{\datasetname{}}, comprising over 10k hours of human data and 3k hours of robot data in a shared schema.
    \item Extensive experiments show that \method{} achieves state-of-the-art cross-embodiment prediction, including on our long-horizon loco-manipulation \textbf{\teleopname{}} benchmark. We further show that \method{}'s performance scales with human data and demonstrate applications in real-time interaction, policy evaluation, model-based planning, and world--action modeling.
\end{itemize}

\section{Related Work}
\label{sec:related}

\textbf{Robotic video world models.}
Early world models learned compact latent dynamics for control~\citep{ha2018worldmodels,hafner2025dreamer}; large video generators yield interactive worlds~\citep{bruce2024genie,parkerholder2024genie2,ball2025genie3,Liu_2025_ICCV,chen20254dnex,liu2025light} and ego-motion-controlled driving simulators~\citep{hu2023gaia1,russell2025gaia2,gao2024vista}.
Distillation and self-rollout training convert video generators into causal few-step models~\citep{yin2025causvid,huang2025selfforcing,cui2025selfforcingpp,liu2025rollingforcing}, augmented with long-horizon memory~\citep{hong2025relic,sun2025worldplay}. Robotic world models learn video dynamics from robot trajectories~\citep{yang2024unisim,zhu2025irasim,jiang2025enerverseac,liao2025genieenvisioner,guo2025ctrlworld,zhen2025tesseract,xu2026kinema4d,ho2025onex}, often adapting a video foundation model~\citep{nvidia2025cosmospredict,wan2025}.
Numerical action tokens require embodiment-specific encoding~\citep{wang2025hma,guo2025ctrlworld}; latent actions inferred from video~\citep{bruce2024genie,ye2025lapa,gao2025adaworld,garrido2026latentwild,gao2026dreamdojo} make unlabeled footage usable, but may entangle actions with other visual changes~\citep{zhang2025whatlam}; spatially aligned renderings provide explicit geometric conditioning~\citep{jiang2025enerverseac,xu2026kinema4d,yang2026orv,wang2025vap,wu2026oscar}.
Cross-embodiment models use embodiment-specific action heads, projections, or adaptation~\citep{wang2025hma,gao2026dreamdojo,nvidia2026cosmos}, a single rendered interface~\citep{wang2025vap,wu2026oscar}, or factorized human-to-robot recomposition~\citep{anyworld2026}. The last, AnyWorld~\citep{anyworld2026}, recomposes human interactions into robot experiences but conditions on source human skeletons rather than target-robot kinematics and is evaluated mainly on manipulation, leaving sustained loco-manipulation untested.
Ego-motion is also studied in navigation and driving~\citep{bar2025nwm,hu2023gaia1} and in egocentric predictors conditioned on or jointly modeling full-body pose~\citep{bai2025peva,pallotta2025egocontrol,xiu2025egotwin}.
\method{} uses an explicit kinematic interface to unify loco-manipulation across humans, gripper robots, and dexterous humanoids.

\textbf{Human and robot datasets.}
Large egocentric human collections cover daily and skilled activity~\citep{grauman2021ego4d,damen2020rescaling,grauman2023egoexo4d}; pose-annotated collections add body or hand pose~\citep{ma2024nymeria,liu2022hoi4d,hoque2025egodex,punamiya2026egoverse}, and portable rigs bring hand pose into everyday scenes~\citep{wang2024dexcap,liu2025egozero}.
Aggregated and in-the-wild demonstrations~\citep{oneill2023open,khazatsky2024droid,walke2023bridgev2,fang2023rh20t,wu2024robomind} and large teleoperation sets~\citep{bu2025agibot} provide robot actions, calibrated views and task captions in varying combinations, with substantial coverage of tabletop arm and gripper manipulation; humanoid sets include locomotion or dexterous hands~\citep{zhao2025humanoid,fourier2025actionnet,qiu2025humanoidpolicy,wei2026psi0,li2026omega0}.
Evaluation data include held-out robot video clips~\citep{zhu2025irasim,gao2026dreamdojo,wu2026oscar}, multimodal world-model benchmarks~\citep{nvidia2026cosmos}, and benchmarks measuring visual quality, physical consistency or embodied utility~\citep{yue2025ewmbench,motamed2026physicsiq}.
\datasetname{} unifies annotations for public human and robot data and self-collected egocentric video; \teleopname{} benchmarks long-horizon dexterous humanoid loco-manipulation with robot-native kinematics.

\textbf{Policy evaluation, planning, and learning.}
Policy evaluation extends real-to-sim protocols~\citep{li2024simpler} to video world models through recorded-action rollouts~\citep{zhu2025irasim,li2025worldeval,wu2026oscar} or closed-loop policy interaction~\citep{quevedo2025worldgym,guo2025ctrlworld,tseng2025scalable,geminirobotics2025veo}.
For planning, world models score or refine action proposals using predicted outcomes~\citep{qi2025predictive,du2025dynaguide,jain2025sailor,gao2026dreamdojo}, in pixel or learned latent spaces~\citep{zhou2025dinowm,assran2025vjepa2}, and support reinforcement fine-tuning and synthetic data generation~\citep{zhu2025wmpo,jiang2025world4rl,li2025vlarft,xiao2025worldenv,jang2025dreamgen}.
Vision-language-action models map observations and language to actions using pretrained vision-language models~\citep{brohan2023rt2,kim2024openvla,black2024pi0,intelligence2025pi05,bjorck2025gr00t}, increasingly trained with human video~\citep{ye2025lapa,bu2025univla,bu2025agibot,yang2025egovla,luo2025beingh0,kareer2025emergence,wei2026psi0}, whereas world--action models couple future video or latent-state prediction with action learning~\citep{cheang2024gr2,zhu2025uwm,hu2025vpp,wang2025hma,liao2025genieenvisioner,ye2026dreamzero,nvidia2026cosmos,motubrain2026,lyu2026lda1b,li2026omega0,wang2026wamsurvey,fastwam2026}.
\method{} supports these applications in Sec.~\ref{sec:experiments}; for world--action modeling, we transfer its pretrained video backbone to initialize such a model rather than adding a dedicated action head.

\begin{figure}[t]
  \centering
  \includegraphics[width=\linewidth]{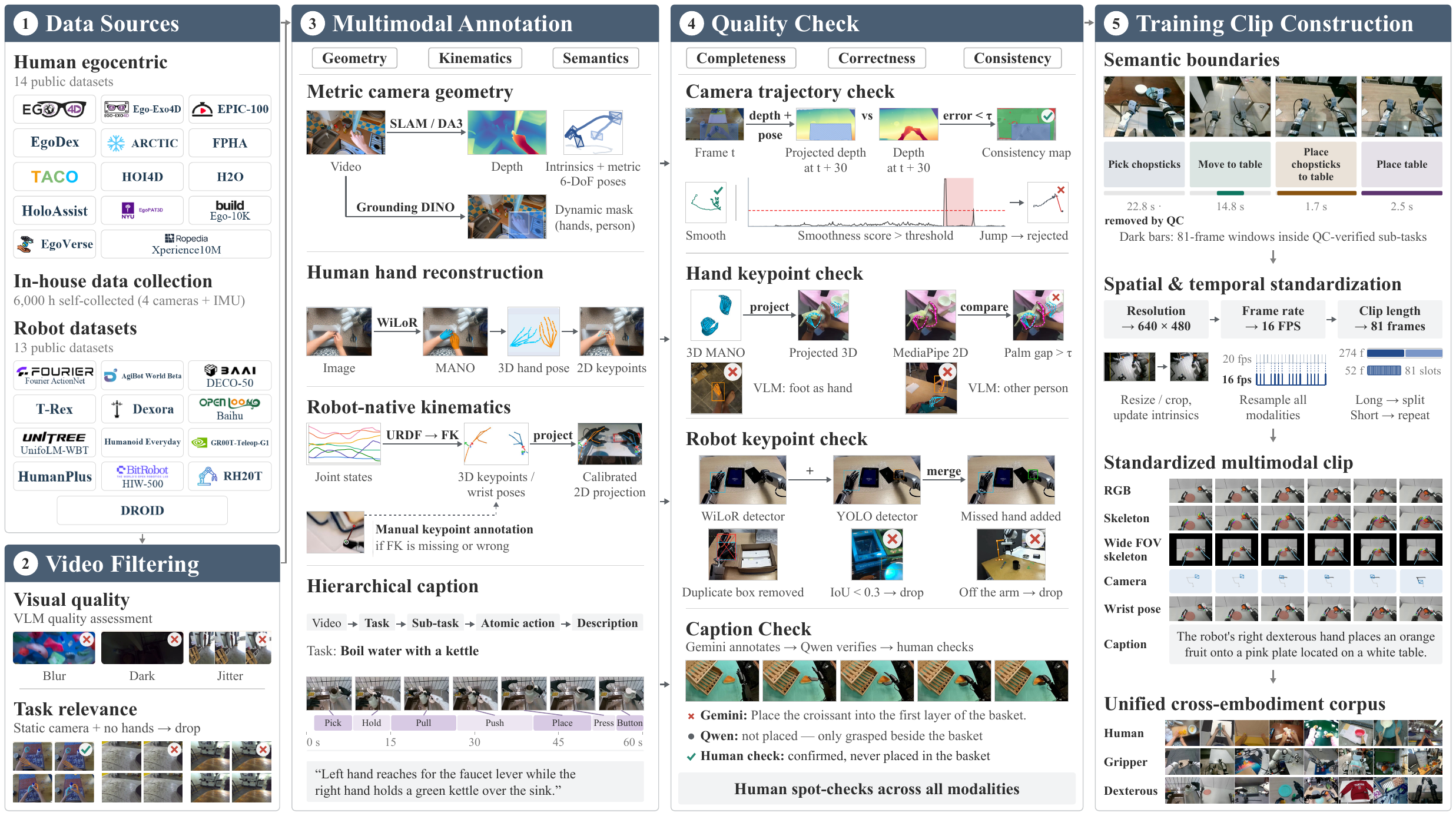}
  \caption{\textbf{\datasetname{} curation pipeline.} Heterogeneous human and robot recordings pass through five stages. (1)~\emph{Data sources}: 14 public human egocentric datasets, 6{,}000\,h of in-house four-camera recordings, and 13 public robot datasets. (2)~\emph{Video filtering} removes low-quality or task-irrelevant clips with a vision-language model. (3)~\emph{Multimodal annotation} produces geometry (metric camera poses via SLAM and DA3~\citep{depthanything3}), kinematics (human hands via WiLoR~\citep{potamias2025wilor} and MANO~\citep{romero2017embodied}, robot keypoints via URDF forward kinematics, manual annotation, and correction), and semantics (hierarchical task, subtask, and atomic-action captions). (4)~\emph{Quality check} verifies completeness, correctness, and cross-modal consistency, pairing automated checks with a caption cascade (Gemini~\citep{comanici2025gemini} annotation, Qwen~\citep{qwen} verification, and human review). (5)~\emph{Training-clip construction} cuts quality-verified subtasks into standardized 81-frame, $640\times480$, 16\,fps multimodal clips, yielding the unified cross-embodiment corpus.}
  \label{fig:data_pipeline}
\end{figure}
\section{Data}
\label{sec:data}
\label{sec:data_extended}

Learning cross-embodiment loco-manipulation requires data that captures both viewpoint changes and manipulator motion under a shared annotation scheme. However, existing datasets~\citep{grauman2021ego4d,grauman2023egoexo4d,hoque2025egodex,wang2023holoassist,damen2020rescaling,buildaiegocentric10k2025,liu2022hoi4d,fang2024egopat3dv2,fan2022arctic,kwon2021h2o,liu2024taco,garciahernando2017firstperson,punamiya2026egoverse,xperience_10m,bu2025agibot,khazatsky2024droid,fang2023rh20t,zhang2026dexora,fourier2025actionnet,bjorck2025gr00t,fu2024humanplus,bitrobot2026hiw,li2026deco,niu2026trex,openloong2025baihu,unitree2026unifolm,zhao2025humanoid} fall short of this requirement: most lack essential geometric information such as camera poses, and their hand annotations are often noisy or inaccurate, making them insufficient for training a general-purpose world model.

To address this gap, our \textbf{human-in-the-loop data pipeline} unifies collection, multimodal annotation, and quality checks (Fig.~\ref{fig:data_pipeline}). We re-annotate public human and robot recordings, with substantial manual robot-keypoint annotation and correction, and collect over 6,000 hours of in-house multi-view video with a four-fisheye-camera rig. This yields \datasetname{}, over 10k human and 3k robot training hours with synchronized RGB, camera trajectories, manipulator skeletons, and captions; wrist poses and wide-FOV skeletons are included where available. Our \teleopname{} benchmark provides about one held-out hour of long-horizon dexterous humanoid loco-manipulation, recorded in the same schema with an off-the-shelf rig~\citep{unitree2025xrteleoperate} (Sec.~\ref{app:teleop}).

Throughout, an \emph{episode} (robot) or \emph{clip}
(human) is a contiguous source recording, a \emph{window} is an 81-frame
training sample extracted from that recording, and a \emph{source} is one
(dataset, end-effector) stream. AgiBot World Beta therefore contributes two
sources: a parallel-gripper stream and a dexterous-hand head-camera stream.

\textbf{Data sources.}
\datasetname{} comprises human and robot data. The human collection combines 14 public egocentric datasets~\citep{grauman2021ego4d,grauman2023egoexo4d,hoque2025egodex,wang2023holoassist,damen2020rescaling,buildaiegocentric10k2025,liu2022hoi4d,fang2024egopat3dv2,fan2022arctic,kwon2021h2o,liu2024taco,garciahernando2017firstperson,punamiya2026egoverse,xperience_10m} with over 6,000 hours of in-house recordings captured using a device equipped with four fisheye cameras, covering diverse daily activities and object interactions. The robot collection draws from 13 public datasets~\citep{bu2025agibot,khazatsky2024droid,fang2023rh20t,zhang2026dexora,fourier2025actionnet,bjorck2025gr00t,fu2024humanplus,bitrobot2026hiw,li2026deco,niu2026trex,openloong2025baihu,unitree2026unifolm,zhao2025humanoid} spanning parallel grippers and dexterous hands. Together, these collections provide complementary coverage of activities, camera motion, and manipulator kinematics.

\textbf{Video filtering.}
We filter the raw recordings for visual quality and task relevance before annotation. A vision-language model screens each segment for severe blur, poor or uneven illumination, over- and under-exposure, and excessive camera jitter, removing clips unsuitable for reliable geometric and kinematic annotation. We discard segments with neither camera motion nor visible hands.

\textbf{Multimodal annotation.}
We annotate geometry, kinematics, and semantics in a unified format. For geometry, we obtain camera intrinsics and metric trajectories from available calibration or estimate them using our in-house SLAM pipeline or DA3~\citep{depthanything3}. Grounding DINO~\citep{liu2024grounding} identifies dynamic regions, such as hands and people, for exclusion from camera-trajectory consistency checks. For human kinematics, we project WiLoR-reconstructed~\citep{potamias2025wilor} MANO~\citep{romero2017embodied} keypoints into images. Captions describe hands, objects, and interactions at task, subtask, and atomic-action levels.

\textbf{Robot keypoint annotation and correction.}
Direct FK projections often misalign with the hands and grippers in the public robot recordings we process, and many releases lack reliable camera calibration. We therefore invest substantial effort in manually annotating image-space keypoints and correcting inaccurate projections. FK provides initial estimates where joint states and calibration are available; manual annotations supply or refine the image-aligned robot skeletons used for training. Wrist poses are derived through FK where available.

\textbf{Quality checks.}
We verify annotation completeness, correctness, and cross-modal consistency. Camera trajectories are checked for depth-reprojection consistency and temporal smoothness, following~\citep{wang2026vggtomega}. Human hand annotations are checked both geometrically and semantically: projected 3D keypoints are compared against independent 2D detections from MediaPipe~\citep{mediapipe} and a fine-tuned MMPose~\citep{mmpose} model, while a vision-language model flags false detections and hands belonging to other people. Robot keypoints are cross-checked by two detectors fine-tuned on our manually annotated robot-hand data, which recover missed hands and remove duplicate or misaligned detections. Qwen~\citep{qwen} verifies Gemini~\citep{comanici2025gemini} captions, with human review resolving discrepancies. Human spot checks across modalities identify systematic annotation errors.

\textbf{Training clip construction.}
\label{sec:data_protocol}
From subtasks that pass quality checks, we build 81-frame clips at $640 \times 480$ (with adjusted intrinsics), each uniformly resampled from a contiguous source interval. A language model~\citep{comanici2025gemini} merges the keyframe captions within a window into one window-level caption. Each clip provides synchronized RGB, projected skeletons, camera parameters, and a caption (a default caption is used for the small fraction of robot windows that lack a generated one), plus wide-FOV skeletons and wrist poses where available, in a format shared across human and robot data.

\begin{figure}[t]
  \centering
  \includegraphics[width=\linewidth]{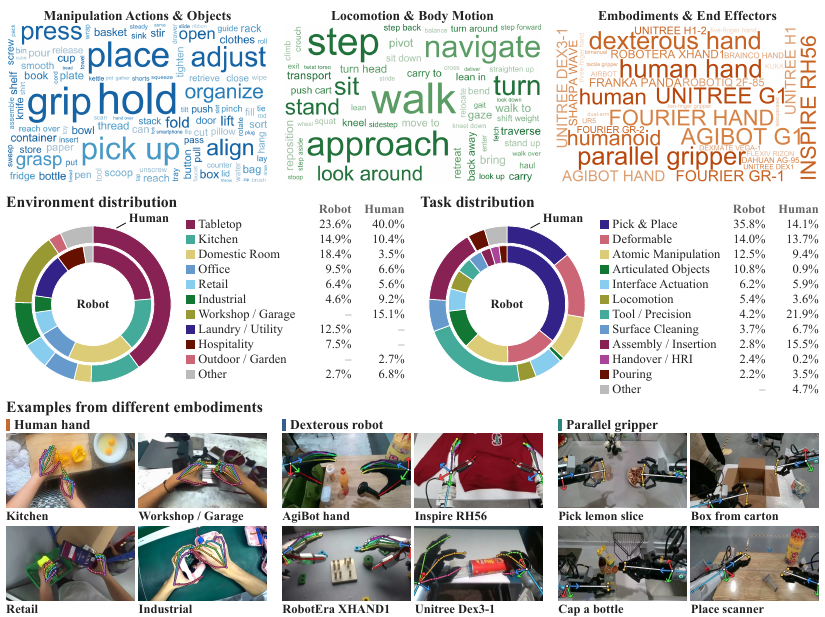}
  \caption{\textbf{Overview of \datasetname{}.} Top: caption terms. Middle: environment and task distributions for robot (inner ring) and human (outer ring) data. Bottom: unified skeleton annotations across human hands, dexterous robot hands, and parallel grippers.}
  \label{fig:training_data_overview}
\end{figure}

The overall curation funnel is summarized in Fig.~\ref{fig:data_funnel} (Appendix~\ref{app:data_figures}). Fig.~\ref{fig:training_data_overview} summarizes the composition of \datasetname{}: the most frequent caption terms, the distribution of environments and tasks for the robot and human portions, and representative unified skeleton annotations across human hands, dexterous robot hands, and parallel grippers. Together they show that the corpus spans diverse scenes and activities while sharing a single annotation format across embodiments.

\section{Model}

\label{sec:method}

\begin{figure}[t]
  \centering
  \includegraphics[width=\linewidth]{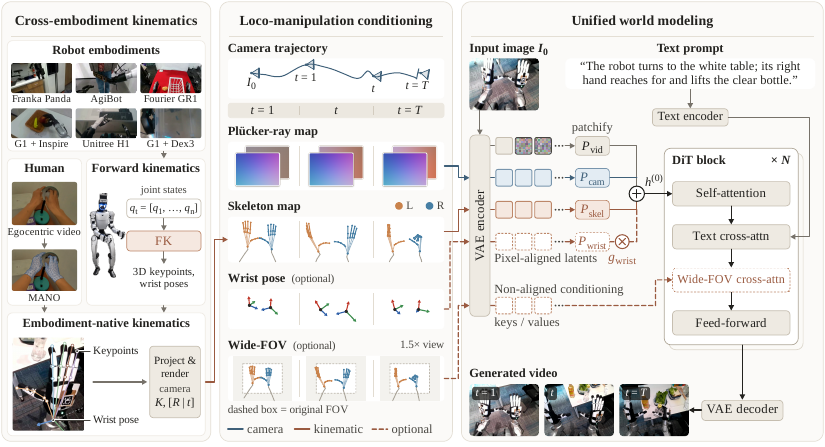}
  \caption{\textbf{Overview of \method{}.} \textbf{Left:} Human pose annotations and robot forward kinematics provide kinematic structure and wrist poses; manual robot-keypoint annotation and correction align robot skeletons with the observed images. \textbf{Middle:} Camera trajectories are encoded as Pl\"ucker-ray maps, while manipulation is represented by skeleton maps, optionally augmented with wrist-pose and wide-FOV renderings. \textbf{Right:} A video diffusion transformer predicts future frames from an initial image, a text prompt, and these control signals. Pixel-aligned conditions are fused through additive projections, while wide-FOV features are injected via cross-attention.}
  \label{fig:method_pipeline}
\end{figure}

\method{} combines a unified kinematic control interface with a video diffusion model for cross-embodiment egocentric loco-manipulation prediction (Fig.~\ref{fig:method_pipeline}). The interface jointly captures viewpoint changes and manipulation through camera poses and projected skeletons, allowing human and robot motion to share a common visual format while preserving embodiment-specific geometry. We first formulate the prediction task and introduce a shared representation of camera and manipulator motion (Sec.~\ref{sec:method-representation}). We then describe the conditioning architecture (Sec.~\ref{sec:method-architecture}), training recipe (Sec.~\ref{sec:method-training}), and distillation (Sec.~\ref{sec:method-distillation}).

\subsection{Representation: Unified Kinematic Control}
\label{sec:method-representation}

Egocentric loco-manipulation requires jointly modeling
locomotion-induced viewpoint changes and manipulator motion.
We encode these complementary factors through camera poses and
projected kinematic skeletons, providing explicit viewpoint control
and spatially aligned manipulation conditioning.
Across embodiments, joint-space actions differ in dimensionality
and semantics, whereas native skeletons can be rendered in a common
image-space format.
Together, these conditions provide a unified interface for humans,
grippers, and dexterous humanoids while preserving
embodiment-specific geometry.

Given an initial image $I_0$, a text condition $c_{\mathrm{txt}}$,
and a per-frame kinematic condition sequence $c_{1:T}$,
\method{} predicts future egocentric frames:
\begin{equation}
\hat{I}_{1:T} \sim
p_\theta\!\left(
I_{1:T} \mid I_0, c_{\mathrm{txt}}, c_{1:T}
\right),
\qquad
c_t = \left(
c_t^{\mathrm{cam}}, c_t^{\mathrm{skel}}
\right),
\label{eq:formulation}
\end{equation}
where $c_{\mathrm{txt}}$ denotes the encoded text prompt,
and $c_t^{\mathrm{cam}}$ and $c_t^{\mathrm{skel}}$ denote
the camera-ray map and projected skeleton rendering, respectively.

\textbf{Camera motion for locomotion.}
We express camera poses relative to the initial frame and encode
them as dense Pl\"ucker-ray maps.
Let $\mathbf{R}_t$ map directions from the current camera frame
to the initial camera frame, and let $\mathbf{o}_t$ denote the
camera center in this reference frame.
Given camera intrinsics $\mathbf{K}_t$ and a homogeneous pixel
coordinate $\tilde{\mathbf{u}}$, we compute
\begin{equation}
\mathbf{d}_t(\mathbf{u}) =
\frac{
\mathbf{R}_t \mathbf{K}_t^{-1} \tilde{\mathbf{u}}
}{
\left\|
\mathbf{R}_t \mathbf{K}_t^{-1} \tilde{\mathbf{u}}
\right\|_2
},
\qquad
c_t^{\mathrm{cam}}(\mathbf{u}) =
\left[
\mathbf{o}_t \times \mathbf{d}_t(\mathbf{u})
\,;\,
\mathbf{d}_t(\mathbf{u})
\right].
\label{eq:plucker}
\end{equation}
This six-channel encoding associates camera geometry with each
image location, providing spatially aligned conditioning for
video generation.
Camera translations are expressed in metric units, maintaining
a consistent physical scale across embodiments.

\textbf{Kinematic skeletons for manipulation.}
Human hand keypoints are obtained from pose annotations.
For curated robot videos, we combine embodiment-native URDF
forward kinematics with manual image-space keypoint annotation
and correction (Sec.~\ref{sec:data}).
Where joint states and calibration are available, the initial FK
projection of keypoint $j$ on robot embodiment $e$ is
\begin{equation}
\tilde{\mathbf{u}}_{t,j}^{(e)} =
\pi\!\left[
\mathbf{K}_t
\left(
\mathbf{R}_t^{\mathrm{c} \leftarrow \mathrm{b}}
\mathcal{F}_j^{(e)}(\mathbf{q}_t)
+
\mathbf{t}_t^{\mathrm{c} \leftarrow \mathrm{b}}
\right)
\right],
\label{eq:fk_projection}
\end{equation}
where $\mathcal{F}_j^{(e)}$ maps joint states $\mathbf{q}_t$
to a keypoint position in the robot base frame,
$(\mathbf{R}_t^{\mathrm{c} \leftarrow \mathrm{b}},
\mathbf{t}_t^{\mathrm{c} \leftarrow \mathrm{b}})$
transforms it into the current camera frame,
and $\pi$ denotes perspective division.
We render the resulting image-aligned keypoints with their native connectivity
into $c_t^{\mathrm{skel}}$, preserving gripper opening geometry
and dexterous finger configurations.
Projection into the camera view at each timestep captures
how these configurations appear under simultaneous camera
and manipulator motion.

When available, wrist-pose renderings $c_t^{\mathrm{wrist}}$
and wide-FOV skeletons $c_t^{\mathrm{wide}}$ supplement the
core representation with explicit wrist orientation and kinematic
context outside the target field of view.

\subsection{Architecture: Kinematics-Conditioned World Model}
\label{sec:method-architecture}

\method{} adopts a latent video diffusion transformer conditioned
on the initial observation, text, and kinematic controls.
We instantiate it at two scales: a dense 5B model and an A14B
mixture-of-experts (MoE) model based on Wan2.2-I2V-A14B~\citep{wan2025}
(approximately 27B total and 14B active parameters).
Spatially aligned conditions are added to the video tokens,
while additional context is incorporated through cross-attention.

\textbf{Spatially aligned conditioning.}
We encode camera-ray maps and skeleton renderings into condition
latents $z^{\mathrm{cam}}$ and $z^{\mathrm{skel}}$, aligned with
the video latents in space and time.
Separate projections map these latents to the transformer's
token dimension:
\begin{equation}
h^{(0)}
=
P_{\mathrm{vid}}(z_\tau)
+
P_{\mathrm{cam}}(z^{\mathrm{cam}})
+
P_{\mathrm{skel}}(z^{\mathrm{skel}}),
\label{eq:condition_injection}
\end{equation}
where $z_\tau$ denotes the video latents at noise level $\tau$,
and $P_{\mathrm{vid}}$ is the pretrained video projection.
The camera and skeleton projections are initialized to zero,
preserving the backbone's behavior at initialization.
Text embeddings $c_{\mathrm{txt}}$ enter through the backbone's
cross-attention layers.

\textbf{Additional kinematic context.}
Wrist-pose renderings follow the same aligned injection scheme,
adding $g_{\mathrm{wrist}}P_{\mathrm{wrist}}(z^{\mathrm{wrist}})$
to $h^{(0)}$, where $g_{\mathrm{wrist}}\in\{0,1\}$ indicates
annotation availability.
Wide-FOV skeletons extend beyond the target view and lack direct
pixelwise correspondence with the video.
Their encoded features therefore enter through an additional
cross-attention branch.
We initialize the wrist projection and the wide-FOV branch's
output projection to zero, and disable each optional branch
when its annotations are unavailable.

\subsection{Training: From Human Experience to Cross-Embodiment}
\label{sec:method-training}

We train \method{} in two stages: human pre-training and
human--robot mid-training.
Both stages share the camera-and-skeleton conditioning interface,
enabling the model to transfer motion-conditioned prediction
learned from human videos to diverse robot embodiments.

\textbf{Human pre-training.}
We first train on large-scale egocentric human videos conditioned
on text, camera trajectories, and projected hand skeletons.
This stage learns the visual consequences of camera and manipulator
motion across diverse environments and interactions, establishing
a foundation for cross-embodiment transfer.

\textbf{Human--robot mid-training.}
We continue training on a balanced mixture of human and robot data,
sampling each with equal probability.
Robot motion is expressed through the same conditioning interface
using embodiment-native kinematics.
This stage introduces robot-specific appearance, manipulator
geometry, and interaction patterns while maintaining exposure
to diverse human experience.
We also incorporate wrist-pose and wide-FOV skeleton conditions
during this stage, using availability gates to accommodate samples
with different annotation coverage.

\subsection{Distillation: Real-Time Interactive Rollouts}
\label{sec:method-distillation}

We first obtain a block-causal, 4-step autoregressive generator using Causal Forcing~\citep{zhu2026causal,zhao2026causalpp}, then extend it to long-horizon streaming generation using LongLive~\citep{yang2025longlive}, retaining image, camera, skeleton, and text conditioning. Causal initialization produces the few-step generator; subsequent asymmetric distribution-matching distillation (DMD) contrasts the clean-latent predictions of a frozen bidirectional real denoiser $\mu_\psi$ (with classifier-free guidance) and an online bidirectional fake denoiser $\mu_\phi$ trained on the generator's samples:
\begin{equation}
\label{eq:dmd}
\nabla_\theta \mathcal{L}_{\mathrm{DMD}}
= \mathbb{E}_{\tau,\epsilon}\!\left[
  \frac{\mu_\phi(z_\tau,\tau)-\mu_\psi(z_\tau,\tau)}
       {\lVert z_0-\mu_\psi(z_\tau,\tau)\rVert_1}
  \frac{\partial z_0}{\partial\theta}
\right].
\end{equation}
Here $z_0$ is the student output and $z_\tau$ its noised version; the update is normalized per sample by the real denoiser's correction magnitude~\citep{yin2024dmd}, and we apply several critic updates per generator update. LongLive subsequently adapts the few-step generator for long-video streaming with a bounded KV cache (a local attention window plus an attention sink). The student streams fixed-size latent blocks at 21.3\,fps at $640\times480$ for minute-long interactive rollouts.

\section{\teleopname{}}
\label{sec:teleop}
\teleopname{} is a humanoid teleoperation benchmark for evaluating long-horizon prediction during locomotion and dexterous manipulation. It comprises 27 long-horizon takes (about one hour of egocentric video at 30\,fps) recorded on a Unitree G1 humanoid with two six-DoF dexterous hands through a whole-body teleoperation rig assembled from off-the-shelf hardware and Unitree's open-source framework~\citep{unitree2025xrteleoperate}. Two operators drive locomotion and manipulation concurrently, so walking and dexterous interaction occur within the same take. Every take is recorded with synchronized RGB-D, camera poses, projected skeletons, robot states, language, tactile, and head-IMU signals in the same schema as \datasetname{}, and is divided into 543 human-annotated atomic segments spanning 30 skill classes (median 18 segments per take). We keep failed grasps and recoveries so that the benchmark also probes whether a world model can simulate off-nominal outcomes, not only successful executions.

\begin{figure}[htbp]
    \centering
    \includegraphics[width=\linewidth]{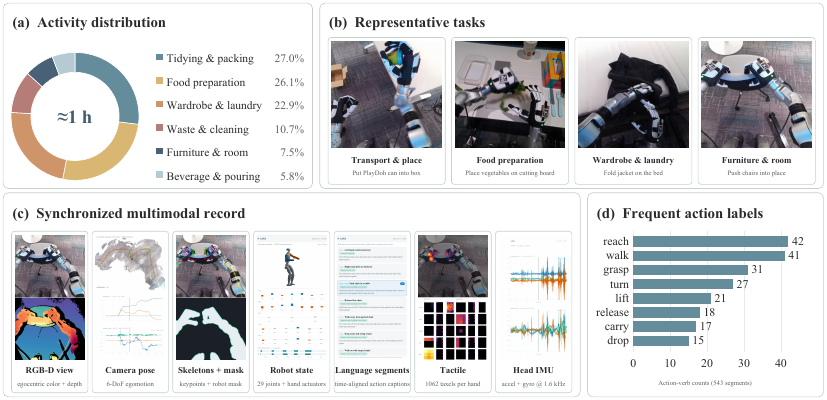}
    \caption{\textbf{Overview of \teleopname{}.} About one hour of dexterous-humanoid teleoperation spanning six everyday household activity groups, from tidying and food preparation to laundry and cleaning (a). Each take is a long-horizon episode that chains locomotion with manipulation, such as putting objects into a box or folding a jacket on the bed (b). Takes are recorded as a single synchronized multimodal stream (egocentric RGB-D, 6-DoF camera pose, projected hand/robot skeletons and masks, 29-joint robot state, time-aligned language segments, per-hand tactile, and head IMU; c), and their 543 annotated segments are dominated by loco-manipulation verbs such as \emph{reach}, \emph{walk}, and \emph{grasp} (d).}
    \label{fig:xteleop_overview_palette}
\end{figure}

Fig.~\ref{fig:xteleop_overview_palette} overviews the activities, representative episodes, synchronized modalities, and action labels, and Table~\ref{tab:teleop_modalities_new} compares \teleopname{} with other humanoid teleoperation datasets in task coverage and available modalities. Full rig, control, recording, and comparison details are provided in Sec.~\ref{app:teleop}.

\begin{table}[t]
    \caption{\textbf{Comparison of humanoid teleoperation datasets} across capabilities (locomotion and dexterity), task horizon (duration, locomotion range, and number of subtasks), and available recording modalities (RGB frame rate, depth, IMU, whole-body joints, tactile, calibration, and keypoints). A check vs.\ cross marks whether each property is provided; datasets are grouped by whether the lower body is used during collection (tabletop vs.\ whole-body). \teleopname{} is the only one offering synchronized long-horizon loco-manipulation with the full modality set.}
    \label{tab:teleop_modalities_new}
    \centering
    \vspace{3pt}
    \begingroup
    \definecolor{yesgreen}{RGB}{25,155,65}
    \definecolor{nored}{RGB}{205,55,55}
    \newcommand{\ym}{\textcolor{yesgreen}{\ensuremath{\checkmark}}}
    \newcommand{\nm}{\textcolor{nored}{\ensuremath{\times}}}
    \newcommand{\rgbm}[2]{#1@#2}
    \scriptsize
    \setlength{\tabcolsep}{2.0pt}
    \renewcommand{\arraystretch}{1.02}
    \resizebox{0.99\linewidth}{!}{%
    \begin{tabular}{@{}ll*{5}{c}@{\hspace{4pt}}*{7}{c}@{}}
    \toprule
    \multirow[c]{2}{*}{Dataset} & \multirow[c]{2}{*}{Robot / hand}
    & \multicolumn{2}{c}{Capabilities}
    & \multicolumn{3}{c}{Task horizon}
    & \multicolumn{7}{c}{Modalities} \\
    \cmidrule(lr){3-4}\cmidrule(lr){5-7}\cmidrule(lr){8-14}
    & & Loco. & Dex.
    & Duration (s) & Loco. range (m) & Subtasks
    & RGB@fps & Dpt. & IMU & WB joints
    & Tac. & Cal. & Kpts. \\
    \midrule
    \multicolumn{14}{@{}l@{}}{\textbf{Tabletop teleoperation}} \\
    RoboMIND (Tien Kung) & Tien Kung / Inspire
    & \nm & \ym & 21.7 & 0 & 2.2 & \rgbm{$640{\times}480$}{30}
    & \ym & \nm & \nm & \nm & \nm & \nm \\
    GR00T-Teleop-GR1 & GR-1 / Fourier
    & \nm & \ym & 14.3 & 0.21 & 1.0 & \rgbm{$1280{\times}800$}{20}
    & \nm & \nm & \ym & \nm & \nm & \nm \\
    Unitree G1 Dex3 & G1 / Dex3
    & \nm & \ym & 27.4 & 0 & 1.0 & \rgbm{$640{\times}480$}{30}
    & \nm & \nm & \nm & \nm & \nm & \nm \\
    Unitree G1 Dex1 & G1 / Dex1 gripper
    & \nm & \nm & 28.7 & 0 & 1.1 & \rgbm{$640{\times}480$}{30}
    & \nm & \nm & \ym & \nm & \nm & \nm \\
    PH2D (robot) & H1 / Inspire
    & \nm & \ym & 12.6 & 0 & 2.6 & \rgbm{$1280{\times}720$}{30}
    & \nm & \nm & \nm & \nm & \nm & \nm \\
    \midrule
    \multicolumn{14}{@{}l@{}}{\textbf{Loco-manipulation teleoperation (legs commanded)}} \\
    Humanoid Everyday & G1 / Dex3; H1 / Inspire
    & \ym & \ym & 20.4 & 0.84 & 2.0 & \rgbm{$640{\times}480$}{30}
    & \ym & \ym & \ym & \ym & \nm & \nm \\
    $\Psi_0$ (real) & G1 / Dex3
    & \ym & \ym & 40.1 & 0.40 & 2.6 & \rgbm{$640{\times}480$}{30}
    & \nm & \nm & \nm & \nm & \nm & \nm \\
    $\omega$-HOME ($\omega_0$) & G1 / Inspire
    & \ym & \ym & 30.4 & 0.33 & 1.1 & \rgbm{$640{\times}360$}{30}
    & \nm & \nm & \ym & \nm & \nm & \nm \\
    UnifoLM-WBT (Inspire) & G1 / Inspire
    & \ym & \ym & 39.8 & 0.45 & 1.7 & \rgbm{$640{\times}480$}{30}
    & \nm & \nm & \ym & \nm & \nm & \nm \\
    UnifoLM-WBT (Dex1) & G1 / Dex1 gripper
    & \ym & \nm & 39.8 & 0.45 & 1.7 & \rgbm{$640{\times}480$}{30}
    & \nm & \nm & \ym & \nm & \nm & \nm \\
    \midrule
    \teleopname{} (ours) & G1 / Inspire
    & \ym & \ym & 118.6 & 1.48 & 3.6
    & \rgbm{$1280{\times}720$}{30}
    & \ym & \ym & \ym & \ym & \ym & \ym \\
    \bottomrule
    \end{tabular}
    }\par
    \endgroup
\end{table}

\section{Experiments}
\label{sec:experiments}

\subsection{Setup}
\label{sec:setup}\label{app:exp_details}

\paragraph{Benchmarks.}
We evaluate action-conditioned video prediction on five cross-embodiment benchmarks spanning distinct embodiments and data sources:
(i) the DreamDojo benchmarks~\citep{gao2026dreamdojo}, three Fourier GR-1 robot evaluation sets from DreamDojo: In-lab (100 clips), EgoDex (65 clips), and DreamDojo-HV (68 clips);
(ii) $\Psi_0$~\citep{wei2026psi0}, 100 evaluation windows of dexterous humanoid manipulation recorded with the $\Psi_0$ vision-language-action (VLA) system (50 seen-task, 50 unseen-task);
(iii) $\omega$-HOME~\citep{li2026omega0}, 100 evaluation windows of household loco-manipulation on a Unitree G1 humanoid with dexterous five-finger hands across 24 tasks (50 seen, 50 from five held-out tasks);
(iv) \teleopname{} (Sec.~\ref{sec:teleop}), 100 evaluation windows of long-horizon dexterous humanoid teleoperation, split by scene (50 windows from held-out takes of training scenes, 50 from unseen scenes);
and (v) AgiBot~\citep{bu2025agibot}, 50 held-out windows of gripper-based manipulation, where all compared models include AgiBot data in training.
In addition, we report zero-shot prediction on 200 held-out human egocentric clips (EgoSuite and OpenAoE; Sec.~\ref{sec:benchmark_results}).

\paragraph{Metrics.}
We report visual quality with PSNR, SSIM~\citep{wang2004image}, and LPIPS~\citep{zhang2018unreasonable}. For benchmarks with camera motion we additionally evaluate camera-trajectory accuracy on poses recovered from the generated video, namely absolute trajectory error (ATE), relative pose error (RPE), and relative rotation error (RRE), as well as hand-following accuracy via keypoint PCK@0.1 (PCKm on matched hands, with detection rate Det). Where noted, we also report video-quality metrics (VBench~\citep{huang2024vbench} subject/background consistency, dynamic degree, and aesthetic quality) and FVD~\citep{unterthiner2019fvd}. Full metric definitions and the per-benchmark evaluation protocol are given in Sec.~\ref{app:ego200}.

\paragraph{Post-training protocol.}
For each benchmark, we evaluate three adaptation settings: \emph{no post-training} (direct transfer), \emph{few-shot} post-training using a small number of episodes per training task, and \emph{full-data} post-training using the entire training split.
Throughout, \emph{zero-shot} (equivalently \emph{no post-training} / direct transfer) means the model is evaluated on a target dataset it was never trained on. The target \emph{embodiment} may still appear in the multi-embodiment mid-training mixture through \emph{other} datasets, but the target evaluation dataset itself is never used for training. ``Seen'' vs.\ ``unseen'' splits within a benchmark refer to seen vs.\ held-out \emph{tasks/scenes} of that benchmark, not to the embodiment.
We evaluate both training stages of \method{}: \emph{pretrain} rows are initialized from the human-only pre-training stage, and \emph{midtrain} rows from the subsequent mid-training stage that mixes human and multi-embodiment robot data.
Checkpoints are always selected on a held-out validation split, never on the benchmark itself.

\paragraph{Baselines.}
We compare against generalist robot world models with different action interfaces:
DreamDojo (2B and 14B), which conditions on continuous latent actions pretrained from large-scale human video, and
Cosmos3 (Nano, 16B; Super, 64B), which treats action as a first-class modality with per-domain action encoders.
On the egocentric human-video and AgiBot benchmarks we additionally report two recent action-conditioned world models where available, OSCAR~\citep{wu2026oscar} and RynnWorld-Teleop~\citep{rynnworld_teleop}.
For data-efficiency comparisons, we additionally post-train the same architecture as ours initialized from the public video-generation weights (``official base''), isolating the contribution of our pre-training.
All baselines are post-trained with their official recipes.

\subsection{Benchmark Results}
\label{sec:benchmark_results}

\textbf{Human egocentric video.} We evaluate on 200 clips from EgoSuite~\citep{lightwheel2026egosuite} and OpenAoE~\citep{li2026open}, two datasets unseen during training by both \method{} and all compared methods. \method{} achieves the lowest camera-trajectory errors, the highest hand-following accuracy, and the best visual quality (PSNR, SSIM, LPIPS) (Table~\ref{tab:human_eval}). These results demonstrate strong generalization to unseen human egocentric datasets, in both motion following and visual quality.

\begin{table}[t]
  \centering
  \caption{\textbf{Results on a 200-clip EgoSuite~\citep{lightwheel2026egosuite} and OpenAoE~\citep{li2026open} evaluation set} (100 clips each). Zero-shot egocentric human-video prediction, reported as (a)~camera- and hand-action following (trajectory errors ATE/RPE/RRE, hand position error HandL2, and PCK-based hand-keypoint accuracy with detection rate Det) and (b)~video prediction and generation quality (PSNR/SSIM/LPIPS, VBench~\citep{huang2024vbench} consistency/dynamics/aesthetics, and FVD~\citep{unterthiner2019fvd}). ATE/RPE are in cm, RRE in degrees, HandL2 in pixels, and PSNR in dB; PCK/PCKm/Det and Cons./Dyn/Aes are reported as percentages. Best bold, second-best underlined; full protocol in Sec.~\ref{app:ego200}.}
  \label{tab:ego200_overall}\label{tab:human_eval}
  \begingroup
  \papertablestyle
  \renewcommand{\arraystretch}{1.04}
  \newcommand{\egomodelsep}{\noalign{\vskip 2pt\hbox to\linewidth{\xleaders\hbox{\rule{2pt}{0.3pt}\hskip 2pt}\hfill}\vskip 2pt}}
  \begin{tabular*}{\linewidth}{@{\extracolsep{\fill}}lccccccc@{}}
    \toprule
    \multirow{2}{*}{Method} & \multicolumn{7}{c}{(a) Camera and hand action following} \\
    \cmidrule(lr){2-8}
    & ATE$\downarrow$ & RPE$\downarrow$ & RRE$\downarrow$ & HandL2$\downarrow$ & PCK@.1$\uparrow$ & PCKm@.1$\uparrow$ & Det$\uparrow$ \\
    \midrule
    Cosmos3-Nano (16B)  & 1.931 & 0.170 & 1.584 & 46.9 & 78.94 & 84.03 & 92.08 \\
    Cosmos3-Super (64B)  & 2.031 & 0.178 & 2.045 & 48.3 & 77.89 & 83.32 & 91.83 \\
    OSCAR (2B)  & 3.480 & 0.273 & 6.258 & 42.1 & 77.60 & 84.46 & 90.55 \\
    RynnWorld-Teleop (5B)  & 3.543 & 0.274 & 6.109 & 49.1 & 74.67 & 82.37 & 88.78 \\
    DreamDojo (2B)  & 2.652 & 0.224 & 4.716 & 69.8 & 58.84 & 70.14 & 81.44 \\
    DreamDojo (14B)  & 2.652 & 0.227 & 3.730 & 65.9 & 60.25 & 72.61 & 80.00 \\
    \egomodelsep
    \method{} (5B), pretrain  & \rankthree{1.793} & \rankthree{0.165} & \rankthree{1.078} & \rankone{32.7} & \ranktwo{83.63} & \rankone{88.66} & \ranktwo{93.15} \\
    \method{} (5B), midtrain  & 1.884 & 0.170 & 1.131 & \ranktwo{33.2} & \rankone{83.85} & \ranktwo{88.56} & \rankone{93.78} \\
    \method{} (A14B), pretrain  & \rankone{1.542} & \rankone{0.152} & \ranktwo{0.980} & \rankthree{36.0} & 81.27 & \rankthree{86.85} & 92.20 \\
    \method{} (A14B), midtrain  & \ranktwo{1.750} & \ranktwo{0.164} & \rankone{0.970} & 36.6 & \rankthree{81.38} & 86.67 & \rankthree{92.62} \\
  \end{tabular*}

  \vspace{2pt}
  \hrule height 0.4pt
  \vspace{3pt}
  \begin{tabular*}{\linewidth}{@{\extracolsep{\fill}}lcccccccc@{}}
    \multirow{2}{*}{Method} & \multicolumn{8}{c}{(b) Video prediction and generation quality} \\
    \cmidrule(lr){2-9}
    & PSNR$\uparrow$ & SSIM$\uparrow$ & LPIPS$\downarrow$ & Cons.$\uparrow$ & Dyn$\uparrow$ & Aes$\uparrow$ & FVD$_{16}$$\downarrow$ & FVD$_{\mathrm{full}}$$\downarrow$ \\
    \midrule
    Cosmos3-Nano (16B)  & 18.66 & 0.6244 & 0.2753 & 93.66 & 74.00 & 40.68 & 71.9 & 178.9 \\
    Cosmos3-Super (64B)  & 18.36 & 0.6119 & 0.2884 & \ranktwo{93.77} & 72.50 & \rankthree{40.89} & \ranktwo{68.5} & \rankthree{175.3} \\
    OSCAR (2B)  & 16.21 & 0.5473 & 0.3818 & 93.15 & 68.50 & \rankone{41.32} & 121.3 & 286.4 \\
    RynnWorld-Teleop (5B)  & 16.94 & 0.5765 & 0.3649 & \rankone{94.26} & 68.00 & 40.05 & 114.0 & 277.5 \\
    DreamDojo (2B)  & 16.75 & 0.5711 & 0.3813 & 92.01 & 67.00 & 38.85 & 128.1 & 322.5 \\
    DreamDojo (14B)  & 17.76 & 0.6014 & 0.3435 & 91.99 & 62.00 & 39.11 & 121.4 & 317.8 \\
    \egomodelsep
    \method{} (5B), pretrain  & 18.90 & \rankthree{0.6409} & \ranktwo{0.2444} & 93.64 & \rankthree{75.00} & 40.73 & 73.0 & 189.9 \\
    \method{} (5B), midtrain  & \rankthree{18.91} & 0.6402 & \rankthree{0.2453} & \rankthree{93.71} & 72.00 & \ranktwo{41.15} & \rankthree{70.1} & 190.1 \\
    \method{} (A14B), pretrain  & \rankone{19.77} & \rankone{0.6749} & \rankone{0.2381} & 93.55 & \ranktwo{76.50} & 40.20 & \rankone{64.0} & \rankone{155.9} \\
    \method{} (A14B), midtrain  & \ranktwo{19.34} & \ranktwo{0.6691} & 0.2462 & 93.45 & \rankone{79.00} & 40.23 & 72.4 & \ranktwo{168.0} \\
    \bottomrule
  \end{tabular*}
  \endgroup
\end{table}

\textbf{DreamDojo benchmark.} Following DreamDojo's protocol, we post-train on its GR-1 teleoperation data and evaluate on three test sets: In-lab, EgoDex, and DreamDojo-HV. \method{} outperforms DreamDojo across all evaluated post-training budgets (1\%--100\%), with the largest gains in the low-data regime, demonstrating data-efficient transfer to the target embodiment (Table~\ref{tab:fewshot_dreamdojo}).

\begin{table}[t]
  \centering
  \caption{\textbf{Comparison on the DreamDojo~\citep{gao2026dreamdojo} benchmarks under different post-training settings.} Despite having only \textbf{5B} parameters, our model consistently outperforms the larger DreamDojo (14B) baseline across all metrics. Best results are in \textbf{bold}, and second best are \underline{underlined}.}
  \label{tab:fewshot_dreamdojo}
  \begingroup
  \papertablestyle
  \renewcommand{\arraystretch}{1.05}
  \begin{tabular*}{\linewidth}{@{\extracolsep{\fill}}ll ccc@{\hspace{7pt}}ccc@{\hspace{7pt}}ccc@{}}
    \toprule
    \multirow{2}{*}{Data} & \multirow{2}{*}{Model} & \multicolumn{3}{c}{In-lab} & \multicolumn{3}{c}{EgoDex} & \multicolumn{3}{c}{DreamDojo-HV} \\
    \cmidrule(lr){3-5}\cmidrule(lr){6-8}\cmidrule(lr){9-11}
    & & PSNR$\uparrow$ & SSIM$\uparrow$ & LPIPS$\downarrow$
    & PSNR$\uparrow$ & SSIM$\uparrow$ & LPIPS$\downarrow$
    & PSNR$\uparrow$ & SSIM$\uparrow$ & LPIPS$\downarrow$ \\
    \midrule
    \multirow{4}{*}{1\%}
       & DreamDojo (2B)    & 18.840 & 0.731 & 0.254 & \rankthree{18.584} & \rankthree{0.749} & \rankthree{0.243} & \rankthree{16.996} & \rankthree{0.724} & \rankthree{0.256} \\
       & DreamDojo (14B)   & \rankthree{18.983} & \rankthree{0.743} & \rankthree{0.244} & 18.070 & 0.747 & 0.246 & 16.643 & 0.716 & 0.262 \\
       & \method{} (5B)    & \ranktwo{20.854} & \ranktwo{0.764} & \ranktwo{0.213} & \ranktwo{20.569} & \ranktwo{0.778} & \ranktwo{0.208} & \ranktwo{19.045} & \ranktwo{0.760} & \ranktwo{0.217} \\
       & \method{} (A14B)  & \rankone{23.028} & \rankone{0.826} & \rankone{0.190} & \rankone{22.354} & \rankone{0.830} & \rankone{0.188} & \rankone{20.788} & \rankone{0.810} & \rankone{0.198} \\
    \cmidrule(lr){1-11}
    \multirow{4}{*}{10\%}
       & DreamDojo (2B)    & 19.238 & 0.744 & 0.244 & \rankthree{19.091} & 0.768 & 0.230 & 17.210 & 0.729 & 0.250 \\
       & DreamDojo (14B)   & \rankthree{19.348} & \rankthree{0.751} & \rankthree{0.235} & 19.057 & \rankthree{0.770} & \rankthree{0.225} & \rankthree{17.511} & \rankthree{0.739} & \rankthree{0.241} \\
       & \method{} (5B)    & \ranktwo{21.021} & \ranktwo{0.786} & \ranktwo{0.206} & \ranktwo{20.776} & \ranktwo{0.801} & \ranktwo{0.199} & \ranktwo{19.290} & \ranktwo{0.787} & \ranktwo{0.207} \\
       & \method{} (A14B)  & \rankone{23.151} & \rankone{0.827} & \rankone{0.185} & \rankone{22.485} & \rankone{0.835} & \rankone{0.181} & \rankone{20.929} & \rankone{0.816} & \rankone{0.191} \\
    \cmidrule(lr){1-11}
    \multirow{5}{*}{100\%}
       & Cosmos-Predict2.5~\citep{nvidia2025cosmospredict}  & 20.576 & 0.774 & 0.222 & 19.952 & \rankthree{0.787} & 0.219 & 18.274 & \rankthree{0.754} & 0.236 \\
       & DreamDojo (2B)    & 21.114 & 0.774 & 0.222 & 20.411 & 0.775 & 0.226 & 18.813 & 0.747 & 0.238 \\
       & DreamDojo (14B)   & \rankthree{21.413} & \rankthree{0.788} & \rankthree{0.208} & \rankthree{20.525} & \rankthree{0.787} & \rankthree{0.213} & \rankthree{18.924} & 0.751 & \rankthree{0.228} \\
       & \method{} (5B)    & \ranktwo{21.738} & \ranktwo{0.797} & \ranktwo{0.202} & \ranktwo{21.230} & \ranktwo{0.810} & \ranktwo{0.196} & \ranktwo{19.814} & \ranktwo{0.795} & \ranktwo{0.204} \\
       & \method{} (A14B)  & \rankone{23.217} & \rankone{0.828} & \rankone{0.186} & \rankone{22.731} & \rankone{0.845} & \rankone{0.177} & \rankone{21.086} & \rankone{0.826} & \rankone{0.187} \\
    \bottomrule
  \end{tabular*}
  \endgroup
\end{table}

\textbf{$\Psi_0$ and $\omega$-HOME.} On these dexterous-humanoid benchmarks (Table~\ref{tab:main_results}; breakdowns in Table~\ref{tab:seen_unseen}), \method{} leads across all adaptation settings. On both, an A14B midtrain variant already matches or exceeds every baseline's full-data result with little or no post-training. Without post-training, the A14B midtrain variant also yields the smallest seen--unseen gap on both benchmarks (0.06/0.69\,dB on $\Psi_0$/$\omega$-HOME); this narrowing does not persist after few-shot or full-data post-training (Table~\ref{tab:seen_unseen}). Qualitative comparisons on $\omega$-HOME (Figs.~\ref{fig:additional_omega_1} and~\ref{fig:additional_omega_2}) illustrate the generated hand motion and object interactions against ground-truth frames.

\begin{table}[t]
  \centering
  \caption{\textbf{World-model evaluation on $\Psi_0$~\citep{wei2026psi0}, $\omega$-HOME~\citep{li2026omega0}, and \teleopname{}} under three post-training settings: no post-training (zero-shot transfer), few-shot, and full-data, reporting PSNR, SSIM, and LPIPS. For \method{} we report both the pretrain (human-only) and midtrain (human\,+\,robot) stages at the 5B and A14B scales. Best bold, second-best underlined.}
  \label{tab:main_benchmarks}\label{tab:main_results}\label{tab:psi0_benchmarks_full}\label{tab:omega_home_benchmarks}\label{tab:teleop_benchmarks}
  \begingroup
  \papertablestyle
  \renewcommand{\arraystretch}{0.95}
  \begin{tabular*}{\linewidth}{@{\extracolsep{\fill}}llccc@{\hspace{6pt}}ccc@{\hspace{6pt}}ccc@{}}
    \toprule
    \multirow{2}{*}{Benchmark} & \multirow{2}{*}{Model / initialization} & \multicolumn{3}{c}{No post-training} & \multicolumn{3}{c}{Few-shot} & \multicolumn{3}{c}{Full-data} \\
    \cmidrule(lr){3-5}\cmidrule(lr){6-8}\cmidrule(lr){9-11}
    & & PSNR$\uparrow$ & SSIM$\uparrow$ & LPIPS$\downarrow$
    & PSNR$\uparrow$ & SSIM$\uparrow$ & LPIPS$\downarrow$
    & PSNR$\uparrow$ & SSIM$\uparrow$ & LPIPS$\downarrow$ \\
    \midrule
    \multirow{7}{*}{$\Psi_0$}
    & DreamDojo (14B) & 14.45 & 0.580 & 0.432 & 14.43 & 0.579 & 0.438 & 13.78 & 0.569 & 0.449 \\
    & Cosmos3-Nano (16B) & 13.74 & 0.563 & 0.465 & 14.71 & 0.584 & 0.409 & \rankthree{16.55} & \rankthree{0.636} & \rankthree{0.320} \\
    & Cosmos3-Super (64B) & 13.78 & 0.562 & 0.461 & 14.94 & 0.591 & 0.400 & 15.73 & 0.613 & 0.352 \\
    & \method{} (5B), pretrain & 15.17 & 0.587 & 0.394 & \rankthree{15.59} & \rankthree{0.606} & \rankthree{0.364} & 16.26 & 0.625 & 0.330 \\
    & \method{} (5B), midtrain & \rankthree{15.22} & \rankthree{0.594} & \rankthree{0.385} & \rankthree{15.59} & 0.603 & 0.369 & 16.41 & 0.628 & 0.325 \\
    & \method{} (A14B), pretrain & \ranktwo{15.87} & \ranktwo{0.614} & \ranktwo{0.368} & \ranktwo{17.83} & \ranktwo{0.664} & \ranktwo{0.272} & \ranktwo{18.94} & \ranktwo{0.692} & \ranktwo{0.241} \\
    & \method{} (A14B), midtrain & \rankone{17.36} & \rankone{0.656} & \rankone{0.291} & \rankone{18.28} & \rankone{0.674} & \rankone{0.260} & \rankone{19.07} & \rankone{0.698} & \rankone{0.232} \\
    \midrule
    \multirow{7}{*}{$\omega$-HOME}
    & DreamDojo (14B) & 14.70 & 0.642 & 0.436 & 14.44 & 0.636 & 0.448 & 16.16 & 0.676 & 0.344 \\
    & Cosmos3-Nano (16B) & 16.24 & 0.686 & 0.336 & 17.24 & 0.712 & 0.293 & 18.84 & \rankthree{0.744} & 0.235 \\
    & Cosmos3-Super (64B) & 15.90 & 0.675 & 0.353 & 16.84 & 0.700 & 0.309 & 18.46 & 0.735 & 0.247 \\
    & \method{} (5B), pretrain & 17.35 & 0.698 & 0.295 & \rankthree{18.35} & \rankthree{0.728} & \rankthree{0.250} & 19.02 & 0.738 & 0.233 \\
    & \method{} (5B), midtrain & \rankthree{17.82} & \rankthree{0.714} & \rankthree{0.277} & 18.30 & 0.721 & 0.259 & \rankthree{19.25} & 0.742 & \rankthree{0.228} \\
    & \method{} (A14B), pretrain & \ranktwo{18.40} & \ranktwo{0.723} & \ranktwo{0.265} & \ranktwo{20.43} & \ranktwo{0.768} & \ranktwo{0.197} & \ranktwo{21.49} & \ranktwo{0.785} & \ranktwo{0.175} \\
    & \method{} (A14B), midtrain & \rankone{19.55} & \rankone{0.755} & \rankone{0.217} & \rankone{20.71} & \rankone{0.775} & \rankone{0.190} & \rankone{21.77} & \rankone{0.788} & \rankone{0.168} \\
    \midrule
    \multirow{7}{*}{\teleopname{}}
    & DreamDojo (14B) & 13.49 & 0.445 & 0.392 & 13.00 & 0.423 & 0.448 & 12.91 & 0.433 & 0.406 \\
    & Cosmos3-Nano (16B) & 13.69 & 0.461 & 0.375 & 14.51 & 0.489 & 0.320 & 14.98 & 0.508 & 0.293 \\
    & Cosmos3-Super (64B) & 13.85 & 0.462 & 0.364 & 14.40 & 0.482 & 0.324 & 14.81 & 0.496 & 0.300 \\
    & \method{} (5B), pretrain & 14.52 & 0.465 & \rankthree{0.336} & 15.50 & \rankthree{0.518} & 0.296 & \rankthree{15.90} & \rankthree{0.520} & \rankthree{0.277} \\
    & \method{} (5B), midtrain & \ranktwo{14.98} & \rankthree{0.489} & \ranktwo{0.320} & \rankthree{15.52} & 0.509 & \rankthree{0.293} & 15.89 & 0.517 & 0.279 \\
    & \method{} (A14B), pretrain & \rankthree{14.95} & \ranktwo{0.496} & 0.343 & \ranktwo{17.98} & \ranktwo{0.600} & \ranktwo{0.215} & \ranktwo{18.26} & \ranktwo{0.601} & \ranktwo{0.209} \\
    & \method{} (A14B), midtrain & \rankone{16.42} & \rankone{0.537} & \rankone{0.259} & \rankone{18.25} & \rankone{0.601} & \rankone{0.205} & \rankone{18.74} & \rankone{0.618} & \rankone{0.196} \\
    \bottomrule
  \end{tabular*}
  \endgroup
\end{table}

\textbf{\teleopname{}.} On long-horizon loco-manipulation split into seen and unseen scenes, A14B midtrain leads all three settings over the best baseline, and its zero-shot result exceeds every baseline's full-data result (Table~\ref{tab:main_results}). Post-training lifts unseen scenes in step with seen ones (Table~\ref{tab:seen_unseen}), consistent with adaptation rather than pure memorization. \method{} also leads on long-horizon autoregressive rollouts ($\approx$32\,s on $\omega$-HOME and $\approx$30\,s on \teleopname{}), with full tables in the appendix (Tables~\ref{tab:omega_home_long} and~\ref{tab:teleop_long30s}).

\textbf{AgiBot.} All compared checkpoints include AgiBot World data; on held-out gripper-manipulation windows, A14B midtrain leads all baselines and 5B is on par with the strongest (OSCAR~\citep{wu2026oscar}), so the advantage is not limited to transfer settings (Table~\ref{tab:agibot_main}).
\begin{table}[t]
  \centering
  \caption{\textbf{Comparison on AgiBot}~\citep{bu2025agibot} (50 held-out gripper windows). VBench~\citep{huang2024vbench} consistency, aesthetics, and dynamic degree. Best bold, second-best underlined.}
  \label{tab:agibot_benchmarks}\label{tab:agibot_main}
  \vspace{2pt}
  \begingroup
  \newcommand{\agibotmodelsep}{\noalign{\vskip 2pt\hbox to\linewidth{\xleaders\hbox{\rule{2pt}{0.3pt}\hskip 2pt}\hfill}\vskip 2pt}}
  \papertablestyle
  \renewcommand{\arraystretch}{0.95}
  \begin{tabular*}{\linewidth}{@{\extracolsep{\fill}}lccc@{\hspace{7pt}}cccc@{}}
    \toprule
    \multirow{2}{*}{Model / initialization} & \multicolumn{3}{c}{Fidelity} & \multicolumn{4}{c}{Video quality} \\
    \cmidrule(lr){2-4}\cmidrule(lr){5-8}
    & PSNR$\uparrow$ & SSIM$\uparrow$ & LPIPS$\downarrow$
    & S-Cons.$\uparrow$ & B-Cons.$\uparrow$ & Aes.$\uparrow$ & Dyn. \\
    \midrule
    DreamDojo (2B), AgiBot post-trained  & 18.26 & \rankthree{0.750} & 0.218 & \rankthree{0.9383} & 0.9520 & 0.4318 & 0.54  \\
    DreamDojo (14B), AgiBot post-trained  & 18.30 & \ranktwo{0.755} & \ranktwo{0.210} & \ranktwo{0.9416} & \ranktwo{0.9526} & 0.4374 & 0.48  \\
    Cosmos3-Nano (16B)  & 17.70 & 0.742 & 0.230 & 0.9159 & 0.9262 & 0.4373 & 0.72  \\
    Cosmos3-Super (64B)  & 17.28 & 0.732 & 0.241 & 0.9181 & 0.9274 & \rankthree{0.4396} & 0.68  \\
    OSCAR (2B)  & \rankthree{18.56} & 0.746 & 0.217 & 0.9281 & 0.9458 & \rankone{0.4477} & 0.66  \\
    \agibotmodelsep
    \method{} (5B), midtrain  & \ranktwo{18.60} & \rankthree{0.750} & \rankthree{0.214} & \rankone{0.9437} & \rankone{0.9573} & 0.4378 & 0.60  \\
    \method{} (A14B), midtrain  & \rankone{20.88} & \rankone{0.790} & \rankone{0.154} & 0.9368 & \rankthree{0.9525} & \ranktwo{0.4407} & 0.64  \\
    \bottomrule
  \end{tabular*}
  \endgroup
\end{table}

\textbf{Single-checkpoint cross-embodiment prediction.} A single midtrain checkpoint predicts all eight embodiments on the cross-embodiment set with no post-training on these evaluation datasets (\emph{zero-shot}; the embodiments appear in mid-training only through other datasets), reaching 19.83\,PSNR at 5B and 21.98 at A14B (Table~\ref{tab:ablation_main}). For a matched comparison, we fine-tune Cosmos3 on the same robot mixture; even so it reaches only 19.02\,PSNR, below our 5B and $\approx$3\,dB below A14B. Mid-training drives this transfer, improving prediction most before any target-specific adaptation (pretrain/midtrain contrast, Table~\ref{tab:main_benchmarks}).

\subsection{Ablation Studies and Analysis}
\label{sec:ablation}

\textbf{Effectiveness of control conditions.}
We train separate variants without skeleton or camera conditioning and evaluate them on the cross-embodiment set (Table~\ref{tab:ablation_main}). Removing skeleton conditioning reduces PSNR by 3.8/1.8\,dB for A14B/5B, respectively, highlighting its importance for visual fidelity. Removing camera conditioning yields smaller average drops of 0.3/0.1\,dB, as six of the eight embodiments exhibit little camera motion. On the embodiments with moving viewpoints, however, removing camera conditioning reduces PSNR by 1.3\,dB and leads to less generated motion. These results demonstrate the complementary roles of the two control conditions, with camera conditioning particularly important for loco-manipulation.

\begin{figure}[t]
  \centering
  \includegraphics[width=\linewidth]{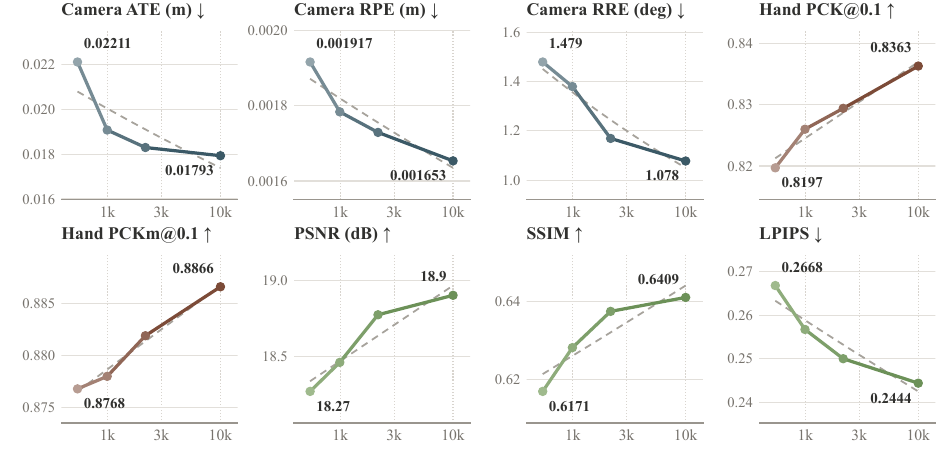}
  \caption{\textbf{Scaling on held-out human video.} Zero-shot evaluation of the 5B model on 200 held-out egocentric clips versus human-video pre-training hours (log scale). The eight panels report camera ATE, RPE, and RRE; hand PCK@0.1 and PCKm@0.1; and PSNR, SSIM, and LPIPS. PCK counts missed hand detections as failures, whereas PCKm scores matched hands only; see Sec.~\ref{app:ego200} for the evaluation protocol. Dashed lines indicate fitted trends.}
  \label{fig:scaling_human200_conf03}
\end{figure}

\begin{figure}[t]
  \centering
  \includegraphics[width=\linewidth]{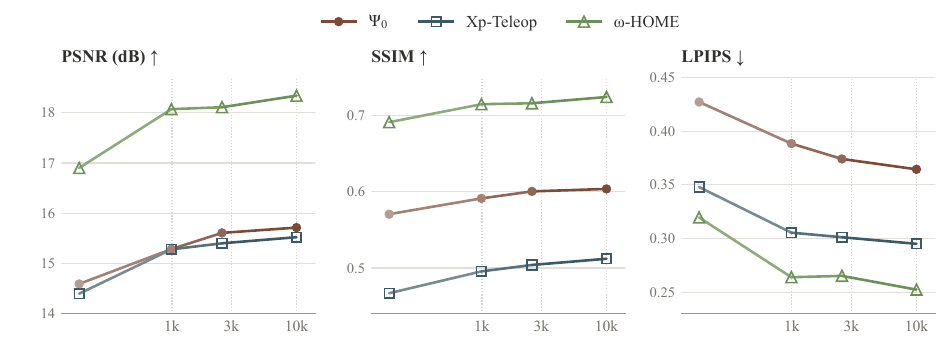}
  \caption{\textbf{Human-to-robot transfer scaling.} PSNR, SSIM, and LPIPS on $\Psi_0$~\citep{wei2026psi0}, \teleopname{}, and $\omega$-HOME~\citep{li2026omega0} versus human-video pre-training hours (log scale); each point is a pre-training checkpoint post-trained on the target benchmark before evaluation. Colors and marker shapes distinguish benchmarks; lines connect measured checkpoint results.}
  \label{fig:scaling_ours3}
\end{figure}

\begin{figure}[t]
  \centering
  \includegraphics[width=\linewidth]{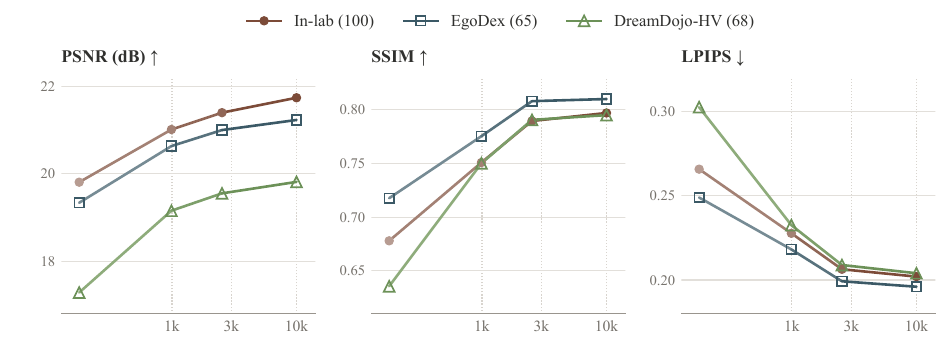}
  \caption{\textbf{Cross-evaluation scaling on the DreamDojo~\citep{gao2026dreamdojo} benchmark.} PSNR, SSIM, and LPIPS on the In-lab, EgoDex~\citep{hoque2025egodex}, and DreamDojo-HV evaluation sets versus human-video pre-training hours (log scale), under an identical GR-1 post-training budget; the largest-scale point uses \method{} (5B). Colors and marker shapes distinguish evaluation sets; lines connect measured checkpoint results.}
  \label{fig:scaling_gr1}
\end{figure}

\textbf{Human pre-training and scaling.}
To isolate the effect of our pre-training, we post-train the same architecture initialized from either public video-generation weights (``official base'') or our pretrained weights (Table~\ref{tab:pretrain_ablation}). With 1\% of the post-training data, \method{} 5B improves PSNR by 4.8/7.1/6.6\,dB over the official base and already outperforms the official base post-trained on 100\% of the data. It retains an approximately 2.3\,dB advantage under full-data post-training. The gains are largest in the low-data regime and remain substantial under full-data post-training. Prediction quality also improves as we scale the amount of human-video pre-training data, showing a consistent, monotonic improvement on held-out human video (Fig.~\ref{fig:scaling_human200_conf03}), in human-to-robot transfer to $\Psi_0$, \teleopname{}, and $\omega$-HOME (Fig.~\ref{fig:scaling_ours3}), and on the DreamDojo GR-1 cross-evaluation sets (Fig.~\ref{fig:scaling_gr1}); see Sec.~\ref{app:scaling} for the scaling protocol.

\textbf{Model scaling.}
Beyond data, increasing the model from 5B to A14B improves prediction consistently across every regime: $+2.2$\,dB PSNR on the cross-embodiment set ($19.83\!\to\!21.98$, Table~\ref{tab:ablation_main}), $+2.3$\,dB in-distribution on AgiBot ($18.60\!\to\!20.88$, Table~\ref{tab:agibot_main}), and, without any post-training, $+2.1/1.7/1.4$\,dB on $\Psi_0$/$\omega$-HOME/\teleopname{} (Table~\ref{tab:main_results}). Model-size and data scaling are thus complementary, jointly driving the gains reported above.

\textbf{Distillation.}
The distilled 4-step 5B student reaches 21.3\,fps on a single GPU ($7\times$ the 50-step teacher, past the 20\,fps real-time line), while matching the teacher's fidelity from short clips (6.3\,s) out to minute-long rollouts (Table~\ref{tab:distillation}; Sec.~\ref{app:distillation}).

\subsection{Downstream Applications}
\label{sec:downstream}

We study four downstream uses of \method{}, spanning simulation and real robots: policy evaluation, model-based planning, world--action modeling, and live teleoperation.

\begin{figure}[!t]
  \centering
  \includegraphics[width=\linewidth]{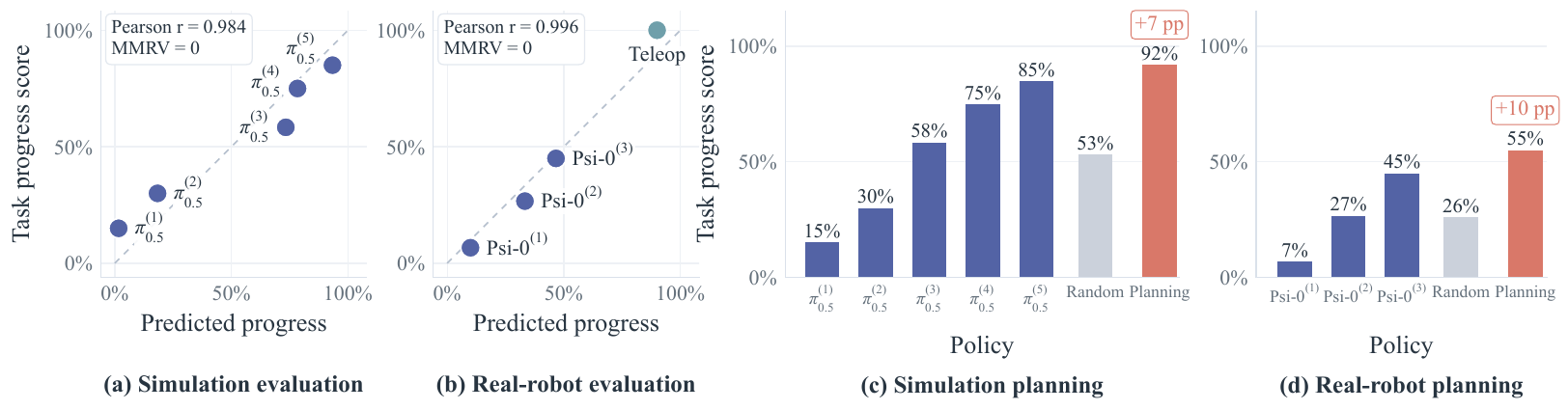}

  \caption{\textbf{Policy evaluation and world-model-based planning.} (a,b) Predicted vs.\ ground-truth task-progress scores in simulation and on the real robot (dashed: perfect agreement). (c,d) Task-progress scores for individual policies and world-model-based candidate selection; \emph{Random} selects one candidate rollout uniformly at random, so its expected score equals the mean over candidates.}

  \label{fig:downstream_results}
  \label{fig:policy_eval}
  \label{fig:policy_eval_real}
  \label{fig:planning}
\end{figure}

\textbf{Policy evaluation.}
Used as a proxy environment, \method{} evaluates policy rollouts using a task-progress rubric following WorldEval~\citep{li2025worldeval} and SIMPLER~\citep{li2024simpler}. Its scores closely match ground-truth evaluations, achieving Pearson correlations of $r=0.984$ across five $\pi_{0.5}$~\citep{intelligence2025pi05} checkpoints in RoboTwin~2.0~\citep{chen2025robotwin} and $r=0.996$ across three $\Psi_0$ checkpoints and an expert-teleoperation reference on a real Unitree~G1 loco-manipulation task. Both evaluations yield MMRV $=0$ and preserve the ground-truth checkpoint rankings (Fig.~\ref{fig:policy_eval}(a,b)), supporting the use of \method{} for policy comparison and checkpoint selection.

\textbf{Model-based planning.}
Following DreamDojo~\citep{gao2026dreamdojo}, \method{} is used for closed-loop planning: it rolls out multiple candidate action sequences, an external value model scores the predicted rollouts, and the selected plan is executed in the environment. We evaluate planning on fixed-view manipulation tasks in simulation and on the real robot. The executed plan reaches 92\% task progress vs.\ 53\% (random selection) / 85\% (best fixed checkpoint) in RoboTwin~2.0, and 55\% vs.\ 26\% / 45\% on the real G1 (Fig.~\ref{fig:planning}(c,d)).

\begin{figure}[t]
  \centering
  \includegraphics[width=\linewidth]{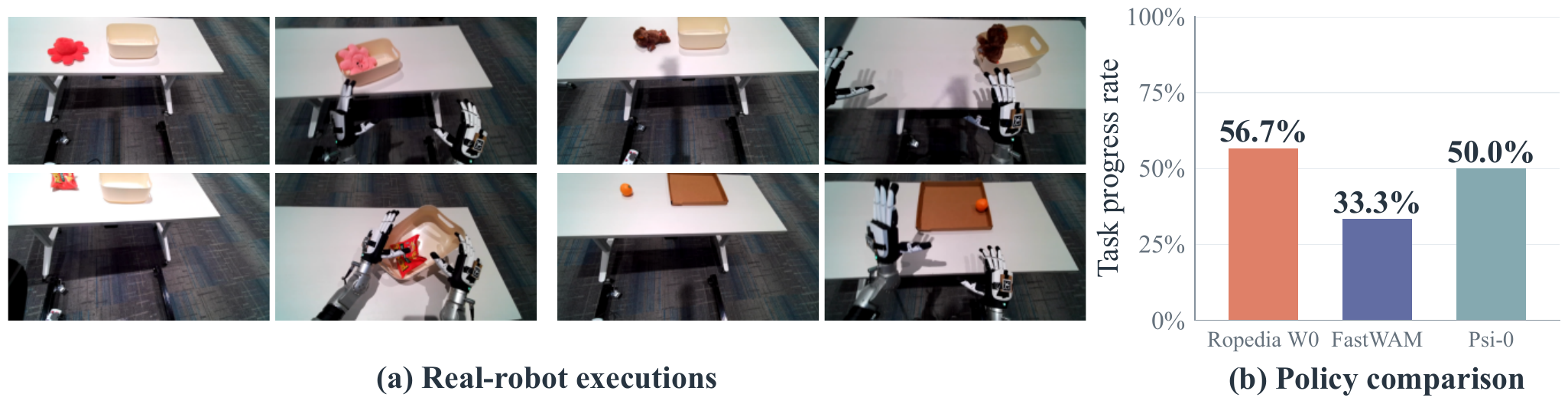}
  \caption{\textbf{World--action modeling.}
  (a) Real-robot executions of the policy initialized from \method{}, shown as four chronological pairs of recorded keyframes with different objects, containers, and placements.
  (b) Aggregate real-robot task progress for \method{} initialization, FastWAM~\citep{fastwam2026}, and the $\Psi_0$~\citep{wei2026psi0} policy.}
  \label{fig:wam_real}
\end{figure}

\textbf{World--action modeling.}
A world--action model jointly predicts future frames and the actions that produce them, so a stronger world model should be a better initialization. We test this by initializing the video backbone of FastWAM~\citep{fastwam2026} from our pretrained 5B world model and leaving its action head and training recipe unchanged. This improves action learning: trained on LIBERO~\citep{liu2023libero} and evaluated directly on both LIBERO and the held-out LIBERO-Plus~\citep{fei2025libero}, it raises LIBERO-Plus success from 50.06\% to 56.82\% and improves every few-shot setting (Table~\ref{tab:wam_libero}), and on a real G1 loco-manipulation task it reaches 56.7\% task progress vs.\ 33.3\% for FastWAM, competitive with the $\Psi_0$ VLA policy itself (50.0\%; Fig.~\ref{fig:wam_real}).

\begin{table}[!t]
  \centering
  \caption{\textbf{World--action modeling on LIBERO~\citep{liu2023libero} and LIBERO-Plus~\citep{fei2025libero}.}
  Success rates (\%) with FastWAM~\citep{fastwam2026} or \method{} init (LIBERO-Plus held out); best bold. Improvement is the absolute gain, relative gain over FastWAM in gray.}
  \label{tab:wam_libero}
  \begingroup
  \definecolor{wamgaincolor}{gray}{0}
  \definecolor{wamlosscolor}{gray}{0.40}
  \definecolor{wamrelativecolor}{gray}{0.40}
  \newcommand{\wampair}[3]{{\renewcommand{\arraystretch}{0.88}\begin{tabular}[c]{@{}c@{}}\textcolor{#1}{#2}\\\textcolor{wamrelativecolor}{(#3)}\end{tabular}}}
  \newcommand{\wampairbold}[3]{{\bfseries\wampair{#1}{#2}{#3}}}
  \papertablestyle
  \renewcommand{\arraystretch}{0.98}
  \setlength{\aboverulesep}{1.5pt}
  \setlength{\belowrulesep}{1.5pt}
  \begin{tabular*}{\linewidth}{@{\extracolsep{\fill}}ll*{9}{c}@{}}
    \toprule
    \multirow{2}{*}{Demos/task} & \multirow{2}{*}{Initialization} & \multirow{2}{*}{LIBERO} & \multicolumn{8}{c}{LIBERO-Plus} \\
    \cmidrule(lr){4-11}
    & & & \multicolumn{1}{c}{Camera} & \multicolumn{1}{c}{Robot} & \multicolumn{1}{c}{Language} & \multicolumn{1}{c}{Lighting} & \multicolumn{1}{c}{Background} & \multicolumn{1}{c}{Noise} & \multicolumn{1}{c}{Layout} & \multicolumn{1}{c}{\textbf{Overall}} \\
    \midrule
    \multirow{3}{*}{full ($\approx$43)}
      & FastWAM & 96.95 & 17.32 & 44.32 & 67.47 & 79.96 & 52.04 & 38.66 & 60.92 & 50.06 \\
      & \method{} & \rankone{97.00} & \rankone{20.14} & \rankone{51.68} & \rankone{73.52} & \rankone{88.05} & \rankone{61.43} & \rankone{48.09} & \rankone{66.30} & \rankone{56.82} \\
      & Improvement & \wampair{wamgaincolor}{+0.05}{0.1\%} & \wampair{wamgaincolor}{+2.82}{16.3\%} & \wampair{wamgaincolor}{+7.36}{16.6\%} & \wampair{wamgaincolor}{+6.05}{9.0\%} & \wampair{wamgaincolor}{+8.09}{10.1\%} & \wampair{wamgaincolor}{+9.39}{18.0\%} & \wampair{wamgaincolor}{+9.43}{24.4\%} & \wampair{wamgaincolor}{+5.38}{8.8\%} & \wampairbold{wamgaincolor}{+6.76}{13.5\%} \\
    \midrule
    \multirow{3}{*}{10}
      & FastWAM & 87.30 & \rankone{18.32} & 23.10 & 43.59 & 55.27 & 29.46 & \rankone{28.36} & 42.36 & 33.58 \\
      & \method{} & \rankone{92.20} & 10.94 & \rankone{24.26} & \rankone{50.94} & \rankone{70.30} & \rankone{40.43} & 21.99 & \rankone{49.84} & \rankone{36.71} \\
      & Improvement & \wampair{wamgaincolor}{+4.90}{5.6\%} & \wampair{wamlosscolor}{-7.38}{-40.3\%} & \wampair{wamgaincolor}{+1.16}{5.0\%} & \wampair{wamgaincolor}{+7.35}{16.9\%} & \wampair{wamgaincolor}{+15.03}{27.2\%} & \wampair{wamgaincolor}{+10.97}{37.2\%} & \wampair{wamlosscolor}{-6.37}{-22.5\%} & \wampair{wamgaincolor}{+7.48}{17.7\%} & \wampairbold{wamgaincolor}{+3.13}{9.3\%} \\
    \midrule
    \multirow{3}{*}{3}
      & FastWAM & 60.00 & 7.57 & 12.58 & 25.31 & \rankone{51.32} & \rankone{19.05} & 15.12 & 27.15 & 21.44 \\
      & \method{} & \rankone{62.65} & \rankone{9.76} & \rankone{13.87} & \rankone{34.74} & 46.31 & 17.38 & \rankone{20.17} & \rankone{30.23} & \rankone{23.97} \\
      & Improvement & \wampair{wamgaincolor}{+2.65}{4.4\%} & \wampair{wamgaincolor}{+2.19}{28.9\%} & \wampair{wamgaincolor}{+1.29}{10.3\%} & \wampair{wamgaincolor}{+9.43}{37.3\%} & \wampair{wamlosscolor}{-5.01}{-9.8\%} & \wampair{wamlosscolor}{-1.67}{-8.8\%} & \wampair{wamgaincolor}{+5.05}{33.4\%} & \wampair{wamgaincolor}{+3.08}{11.3\%} & \wampairbold{wamgaincolor}{+2.53}{11.8\%} \\
    \bottomrule
  \end{tabular*}
  \endgroup
\end{table}

\textbf{Live teleoperation.}
A Quest~3 headset maps the user's head and hand motion to camera poses and skeletons, and the distilled 5B student generates egocentric video in real time, letting users interactively drive viewpoint and manipulation across embodiments (Fig.~\ref{fig:live_teleop}; videos in the supplement).

\begin{figure}[t]
  \centering
  \includegraphics[width=\linewidth]{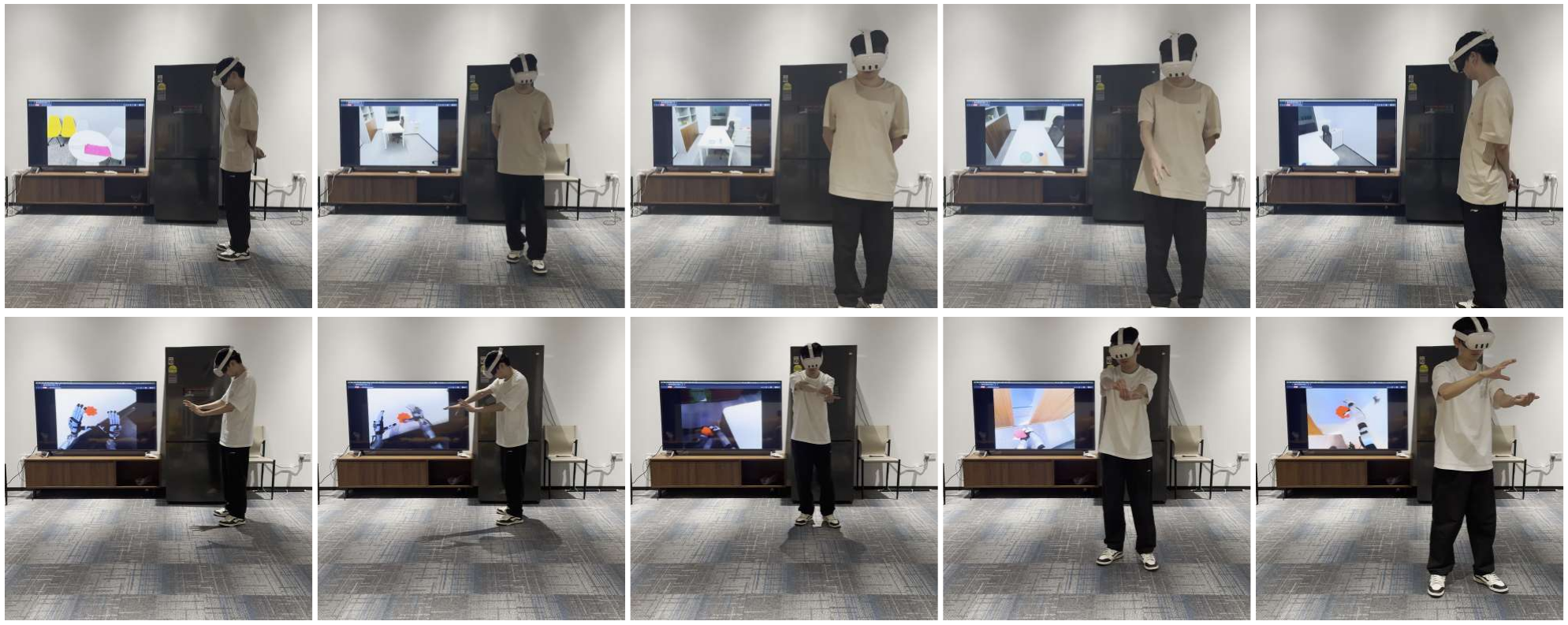}
  \caption{\textbf{Live teleoperation with \method{}.}
  On-site recordings show the headset user and the monitor displaying model-generated feedback.
  Five chronological keyframes show interactions with human (top) and robot (bottom) embodiments.}
  \label{fig:live_teleop}
\end{figure}

\begingroup
\makeatletter
\setlength{\@fptop}{0pt}
\makeatother
\FloatBarrier
\endgroup

\section{Conclusion}
\label{sec:conclusion}

We presented \method{}, a unified egocentric world model for loco-manipulation across humans and robots with grippers or dexterous hands. Its explicit kinematic representation combines camera poses for viewpoint motion with projected skeletons for manipulation, preserving embodiment-native geometry within a shared visual format. This interface enables learning from human and robot data and remains consistent from pre-training to post-training. Trained on Xp-Data with 10k hours of human video and 3k hours of robot data, \method{} demonstrates strong zero-shot generalization, data-efficient adaptation, and competitive prediction across diverse benchmarks, including our new \teleopname{}. Our scaling analysis further shows that increasing human-video pre-training improves prediction on held-out human video and transfer to robot benchmarks, highlighting human experience as a scalable source of supervision for cross-embodiment world modeling. Beyond video prediction, the model supports policy evaluation, model-based planning, and world--action modeling, while distillation enables real-time interactive simulation.

\newpage
{\small\bibliography{references}}

\clearpage
\appendix
\startcontents[appendix]
\printcontents[appendix]{l}{1}{%
  \section*{Appendix Contents}%
  \vspace{-0.5em}%
}
\clearpage
\setcounter{figure}{0}
\setcounter{table}{0}
\renewcommand{\thefigure}{S\arabic{figure}}
\renewcommand{\thetable}{S\arabic{table}}
\renewcommand{\thesection}{{\Alph{section}}}
\renewcommand{\theHfigure}{app.fig.\arabic{figure}}
\renewcommand{\theHtable}{app.tab.\arabic{table}}
\renewcommand{\theHsection}{app.\Alph{section}}

\section{Data and Benchmark Details}
\label{app:data_details}

We first document the curated training corpus and its unified annotations, then describe the held-out \teleopname{} benchmark, including its collection setup and dataset-comparison criteria.

\subsection{Data Curation and Annotation Examples}
\label{app:data_figures}

\begin{figure}[!htbp]
  \centering
  \includegraphics[width=0.92\linewidth]{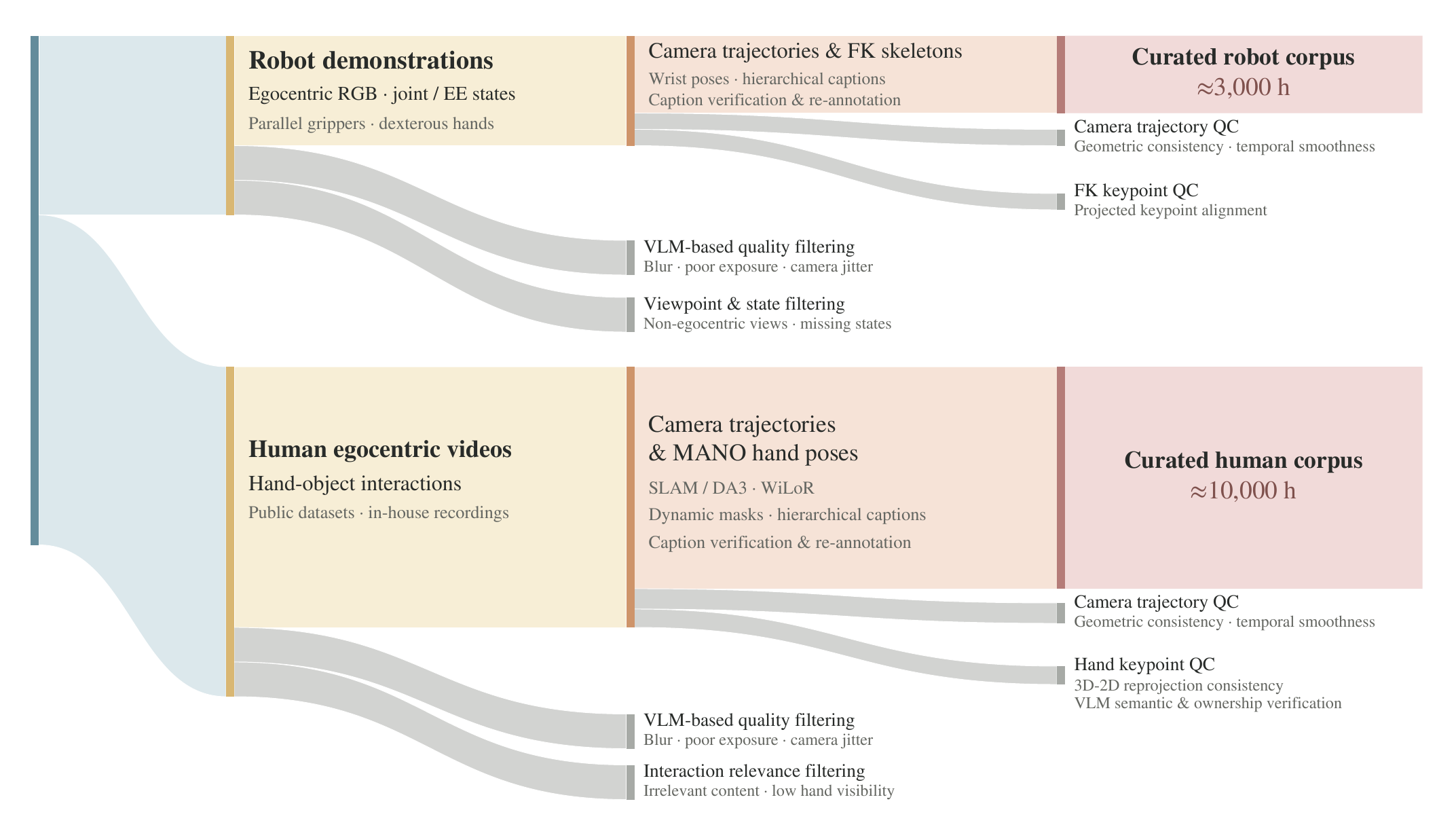}
  \caption{\textbf{Data curation funnel for \datasetname{}.} Robot and human egocentric recordings pass through source screening, motion annotation, and quality control; gray bands mark data removed at each step, and the surviving streams form the unified corpus of $\sim$3,000 robot hours and $\sim$10,000 human hours.}
  \label{fig:data_funnel}
\end{figure}

Figure~\ref{fig:data_funnel} summarizes the curation pipeline. The examples below illustrate the resulting scene coverage and unified motion annotations.

\label{app:data_gallery}

\begin{figure}[!htbp]
  \centering
  \includegraphics[width=0.90\linewidth]{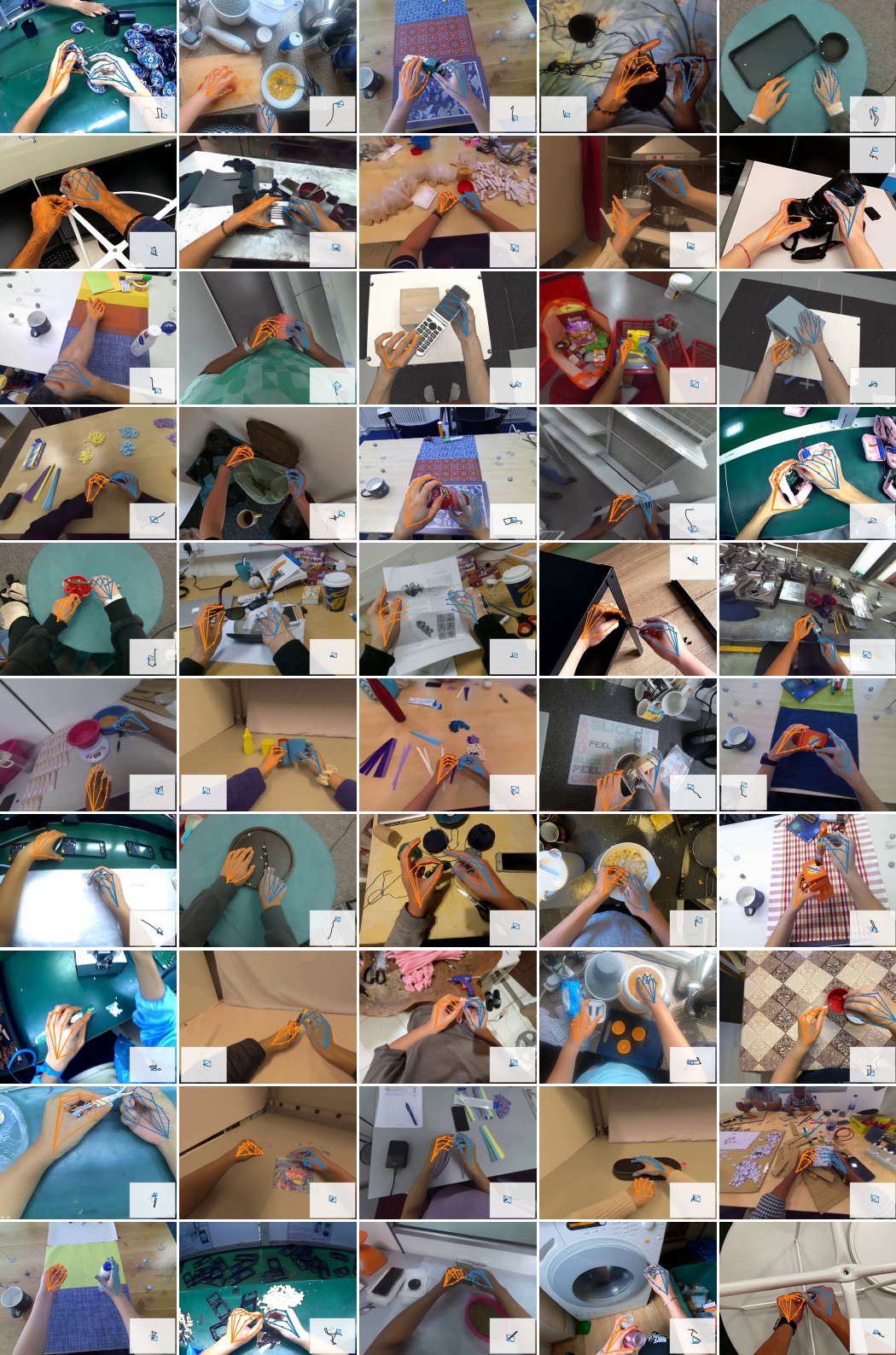}
  \caption{\textbf{Motion annotations on the egocentric human data.} Projected 21-point hand skeletons are overlaid on each frame, with the two hands drawn in different colors; the inset shows the camera trajectory of the surrounding window.}
  \label{fig:human_modalities}
\end{figure}

\begin{figure}[!htbp]
  \centering
  \includegraphics[width=0.90\linewidth]{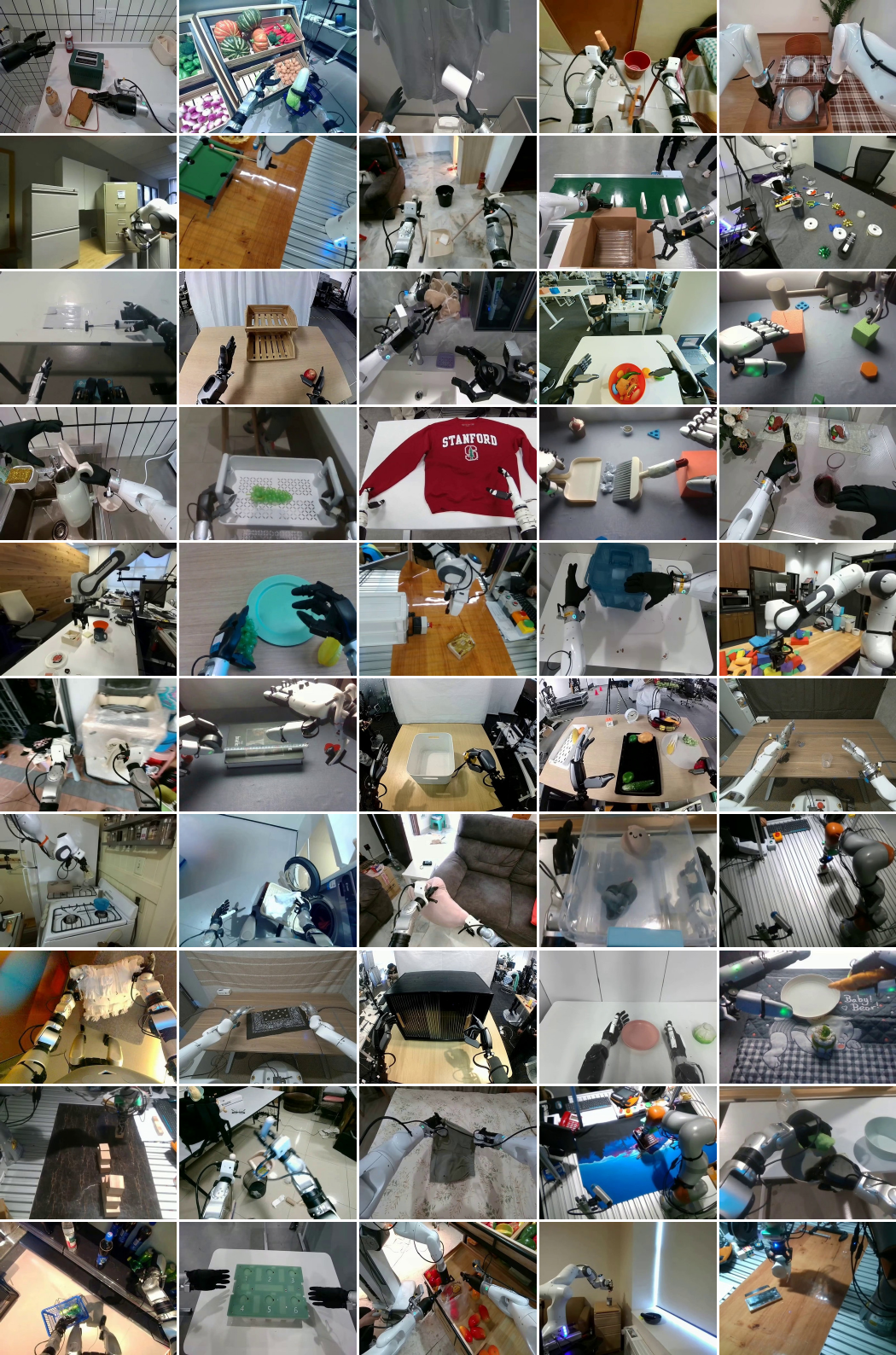}
  \caption{\textbf{Example frames from the robot data in \datasetname{}.} The grid spans dexterous five-finger hands, three-finger hands, and parallel grippers on single-arm, dual-arm, and humanoid platforms, across household, retail, industrial, and tabletop scenes.}
  \label{fig:robot_diversity}
\end{figure}

\begin{figure}[!htbp]
  \centering
  \includegraphics[width=0.88\linewidth]{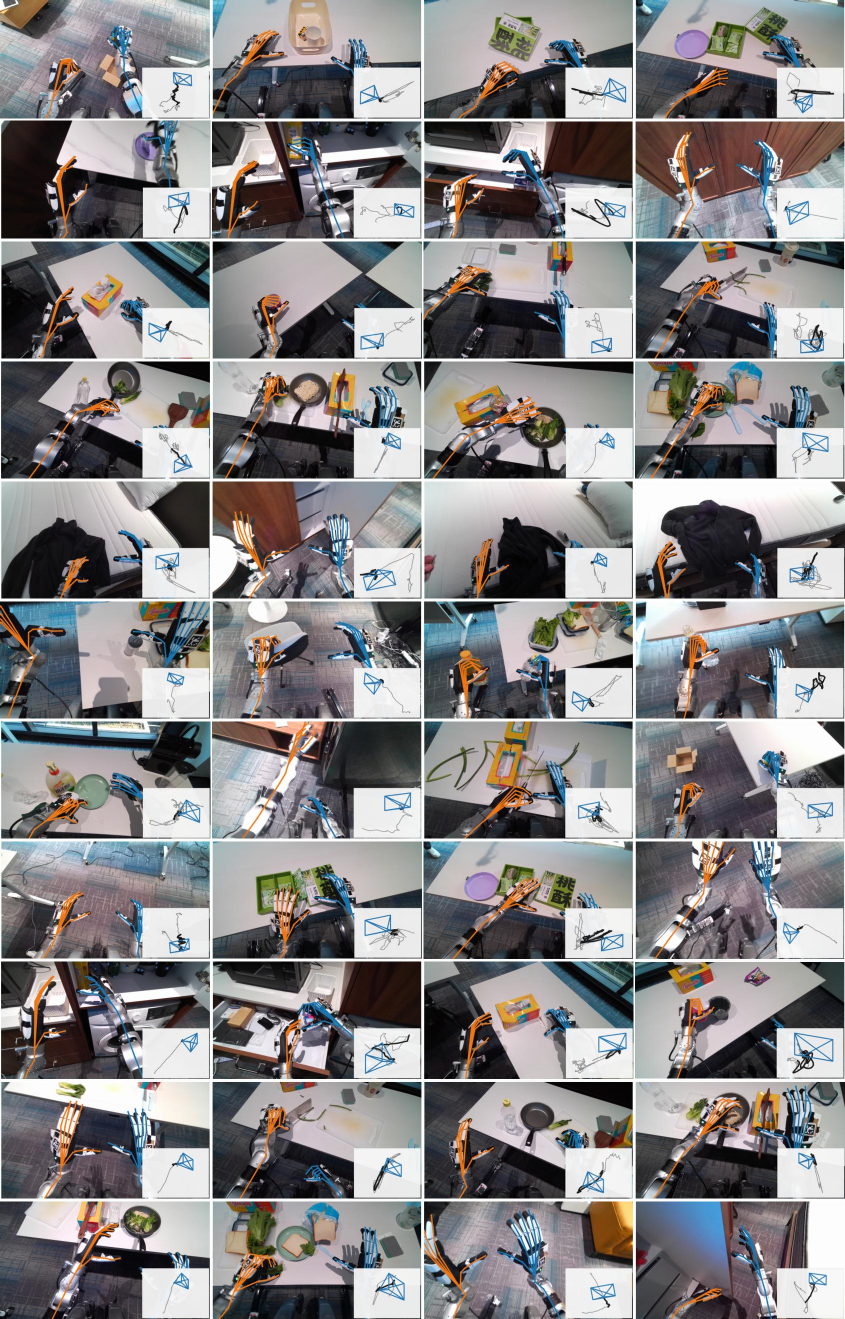}
  \caption{\textbf{Motion annotations on the robot data.} Image-aligned skeletons combine embodiment-native FK projections with manual keypoint annotation and correction (Sec.~\ref{sec:data}), covering dexterous hands and grippers; the inset shows the camera trajectory of the surrounding window.}
  \label{fig:robot_modalities}
\end{figure}

Figures~\ref{fig:human_modalities}--\ref{fig:robot_modalities} visualize the corpus across human and robot recordings.
Figure~\ref{fig:robot_diversity} shows example robot frames spanning different environments, tasks, objects, and embodiments.
Figures~\ref{fig:human_modalities} and \ref{fig:robot_modalities} overlay the unified motion annotations on the same kinds of frames: projected hand or manipulator skeletons in the image, with the per-window camera trajectory shown in the inset.

\FloatBarrier
\subsection{\teleopname{} Collection and Benchmark Details}
\label{app:teleop}

The benchmark overview (Fig.~\ref{fig:xteleop_overview_palette}) and the dataset comparison (Table~\ref{tab:teleop_modalities_new}) appear in the main text (Sec.~\ref{sec:teleop}); here we provide the full rig, control, recording, and comparison details.

\paragraph{Rig.}
A Unitree G1 humanoid with 29 actuated degrees of freedom is equipped with two Inspire RH56DFX dexterous hands, each with six actuated degrees of freedom.
An Azure Kinect DK mounted on a custom 3D-printed helmet rigidly attached to the robot's head records the egocentric RGB-D view.
Because the G1 has no actuated neck joints, the camera moves only with the waist and mobile base, matching the motion represented by our camera-pose control channel (Sec.~\ref{sec:method}).

\paragraph{Control.}
Two operators control the robot concurrently, allowing locomotion and manipulation to occur within the same take.
The manipulation operator wears a Meta Quest 3, whose controller-free hand tracking provides targets to a Pinocchio-based inverse-kinematics solver for the two 7-DoF arms and to a dexterous-hand retargeting module.
The locomotion operator independently commands the robot's base velocity using Unitree's R3 remote.

\paragraph{Recording.}
Each take records the robot and hand joint states, commanded targets, R3 commands, headset tracking, and Kinect stream on a common clock.
From these signals, we derive the two action channels described in Sec.~\ref{sec:method}: camera poses and hand skeletons.
The hand skeletons are computed through forward kinematics and projected into the image using the calibrated helmet-to-camera extrinsics.
We retain every take initiated by the operators, including failed grasps and subsequent recoveries, because \teleopname{} is intended to evaluate whether a world model can simulate off-nominal outcomes as well as successful executions.

\paragraph{Dataset composition and annotation granularity.}
\teleopname{} contains 27 long-horizon teleoperation takes, totaling approximately one hour of egocentric video recorded at 30\,fps.
The takes are divided into 543 human-annotated atomic segments, with each take containing between 8 and 42 segments (median 18).
This annotation is substantially finer-grained than the common convention of assigning a single instruction to an entire episode.
The resulting segments cover 30 skill classes.
The segment captions also identify the manipulated objects and describe the temporal progression of each task, providing language supervision aligned with the recorded observations and actions.

\paragraph{Scope of the dataset comparison.}
Table~\ref{tab:teleop_modalities_new} compares publicly described dexterous-humanoid teleoperation datasets in terms of robot capabilities, task horizon, and available recording modalities.
We divide the datasets into two groups according to how the lower body is used during data collection.
In the tabletop-teleoperation group, the robot's legs remain locked even if the release provides base commands or whole-body state.
In the loco-manipulation group, locomotion is actively commanded and can occur within the same episode as manipulation.

\paragraph{Task-horizon metrics.}
The three task-horizon columns report episode-level means over the analyzed portion of each dataset.
\emph{Duration} is the wall-clock length of one episode, computed as the number of recorded frames divided by the dataset's frame rate (30\,Hz for all datasets except GR00T-Teleop, 20\,Hz). Frame counts are taken from each dataset's own metadata. For our dataset we report the released task windows, not the raw recordings. RoboMIND does not document its frame rate; we assume 30\,Hz.

\emph{Loco. Range} measures how much space an episode covers. For each episode we take the 3-D position trajectory of the robot base (or, when only a camera trajectory is available, the head camera), compute its axis-aligned bounding box, and report the box diagonal in metres. We use the bounding-box diagonal rather than integrated path length, because path length at 30\,Hz is dominated by body sway and pose-estimation jitter (on our data it inflates the value by roughly 7\ensuremath{\times}).

Pose sources. Ours: visual-inertial SLAM (metric). Humanoid-Everyday: on-board odometry. UnifoLM-WBT (new schema): the robot state estimator's base pose. All other datasets provide no metric base pose; we run DA3-Streaming with its metric configuration on the egocentric video (10\,fps, 504\,px input) to obtain camera poses. Because monocular metric SLAM has a systematic scale error, we calibrate it once against ground truth: on the 879 Humanoid-Everyday locomotion episodes the DA3 box diagonal exceeds the odometry value by a median factor of 1.39 (IQR 1.28--1.55), and on our 27 episodes the median factor against SLAM is 1.03; we divide all DA3-derived values by the pooled median 1.39. After calibration the estimator reproduces ground truth within 10\% on both datasets (ours 1.48 vs 1.61\,m; Humanoid-Everyday 0.84 vs 0.82\,m). For AgiBot-Beta we estimate on a length-stratified sample of 30 episodes. Unitree G1 Dex1/Dex3, PH2D, and RoboMIND Tien Kung are reported as 0.

\emph{Subtasks} counts how many sequential sub-goals an episode's task instruction specifies. We apply one rule to every dataset: the instruction is lower-cased and split into clauses at commas, semicolons, periods and the connectives "and", "then", "until", "upon", "after"; a clause counts as one subtask if its first or second content word is an action verb from a fixed list (pick, place, put, take, grasp, carry, walk, go, enter, turn, open, close, push, pull, pour, wipe, fold, locate, find, hang, press, insert, \ldots{}). Locomotion clauses count the same as manipulation clauses. Every instruction counts at least 1. Instruction sources: the per-episode task string for LeRobot datasets, the HDF5 description attribute for PH2D and RoboMIND, the task name for $\omega$-HOME, and the human-written task instruction of each take for our dataset (one 19-step instruction excluded as an outlier). AgiBot-Beta ships no instruction text. Because subtask boundaries follow the wording and granularity of each release's instructions, this count is an annotation-level measure of task horizon rather than a platform-independent action count.

\paragraph{Modality criteria.}
The modality columns describe information available in the analyzed dataset release. For \emph{Kpts.}, we also count image-space keypoints that can be reconstructed from released joint states and camera calibration using a publicly available URDF.
\emph{RGB@fps} reports the resolution and frame rate of the principal egocentric RGB stream.
\emph{Dpt.} requires a usable released depth stream; placeholder or zero-filled channels do not qualify.
\emph{IMU} requires explicit inertial-sensor measurements.
Estimated root poses, torso poses, or camera trajectories alone are not counted as IMU data.
\emph{Tac.} denotes released tactile measurements.

\emph{WB joints} requires recorded joint states covering both the upper and lower body, including the legs.
Arm-only or hand-only states, estimated root or torso poses, and commanded base velocities do not satisfy this criterion.
This column describes the coverage of the recorded state and does not require the leg joints to exhibit substantial motion during the released episodes.
Consequently, a dataset may provide whole-body joint states while still belonging to the tabletop group.

\emph{Cal.} requires sufficient camera calibration to relate robot kinematics to the image observations, including both camera intrinsics and the relevant extrinsics.
Camera trajectories or extrinsics provided without intrinsics do not satisfy this criterion.

\emph{Kpts.} denotes temporally aligned image-space keypoints, either provided directly or obtainable by projecting the relevant robot kinematics using released joint states, sufficient camera calibration, and a publicly available URDF.

\paragraph{Dataset-specific qualifications.}
For Unitree Dex1, whole-body joint states are available only in the v3 release; v2 provides arm and gripper states but not the corresponding lower-body joint states.
Although v3 supports base commands, the observed variation of its leg joints is below 0.03\,rad in the analyzed recordings.
We therefore treat its measured locomotion range as zero and place it in the tabletop group.

$\Psi_0$ supports locomotion, but the analyzed release does not provide leg-joint states.
Its released depth channel is zero-filled and is therefore not counted as usable depth.
RoboMIND entries refer specifically to its Xsens dexterous-hand subset.
This subset includes depth observations but does not provide hand-joint states.
Humanoid Everyday task-horizon statistics are computed from its loco-manipulation subset, and its tactile measurements are available only for the G1/Dex3 embodiment.
The analyzed $\omega$-HOME files do not contain a depth stream.

\FloatBarrier

\section{Model and Implementation Details}
\label{app:model_details}

This section describes the video backbones, distillation and interactive inference, and the optional wide-FOV skeleton-conditioning branch. Figures and measurements are grouped with the component they illustrate.

\subsection{Backbone and Conditioning}
\label{app:preliminaries}

\method{} predicts future egocentric video from an initial visual observation and factorized camera and manipulator controls (Sec.~\ref{sec:method-representation}). The following details describe the pretrained backbones used by this shared conditioning interface.

\paragraph{Latent video diffusion backbones.}
We instantiate \method{} at two scales using the Wan2.2 family of pretrained video diffusion models~\citep{wan2025}. Wan2.2-TI2V-5B uses a dense diffusion transformer with approximately 5B parameters. Wan2.2-I2V-A14B uses two experts specialized for different noise regimes: a high-noise expert and a low-noise expert. Only one expert is active at each denoising step, yielding approximately 14B active parameters. In our experiments, the two experts are adapted separately and used in their respective noise regimes during inference.

Both backbones operate on spatiotemporal latents produced by a video VAE and follow the diffusion transformer architecture~\citep{peebles2023dit}. Text embeddings are incorporated through cross-attention, while sinusoidal timestep embeddings provide adaptive normalization and residual modulation within each transformer block. We train them with the latent flow-matching objective~\citep{lipman2023flow}: the video VAE encodes each clip into latents, the first temporal latent is retained as clean visual context, and noise and the loss are applied only to future latents. The camera-ray and skeleton conditioning branches introduced in Sec.~\ref{sec:method-architecture} augment these backbones through zero-initialized projections, preserving the pretrained model's behavior at initialization.

\FloatBarrier
\clearpage
\subsection{Distillation and Interactive Simulation}
\label{app:distillation}

\paragraph{Distillation recipe.}
We first obtain a 4-step autoregressive generator using Causal Forcing~\citep{zhu2026causal,zhao2026causalpp}, then extend it to long-video streaming using LongLive~\citep{yang2025longlive}, retaining camera--skeleton conditioning.

\emph{Few-step distillation.} An autoregressive diffusion model provides causal initialization. Asymmetric DMD then trains the causal generator using a frozen bidirectional real-score teacher (with classifier-free guidance) and an online bidirectional fake-score model. The generator starts from the causal initialization, while both score networks initialize from the bidirectional camera--skeleton model. The fake-score model learns a denoising objective on generated samples.

\emph{Long-video extension.} LongLive adapts the few-step generator for streaming with within-block bidirectional and cross-block causal attention. A bounded KV cache retains a local window and an attention sink.

\begin{figure}[!htbp]
  \centering
  \includegraphics[width=\linewidth]{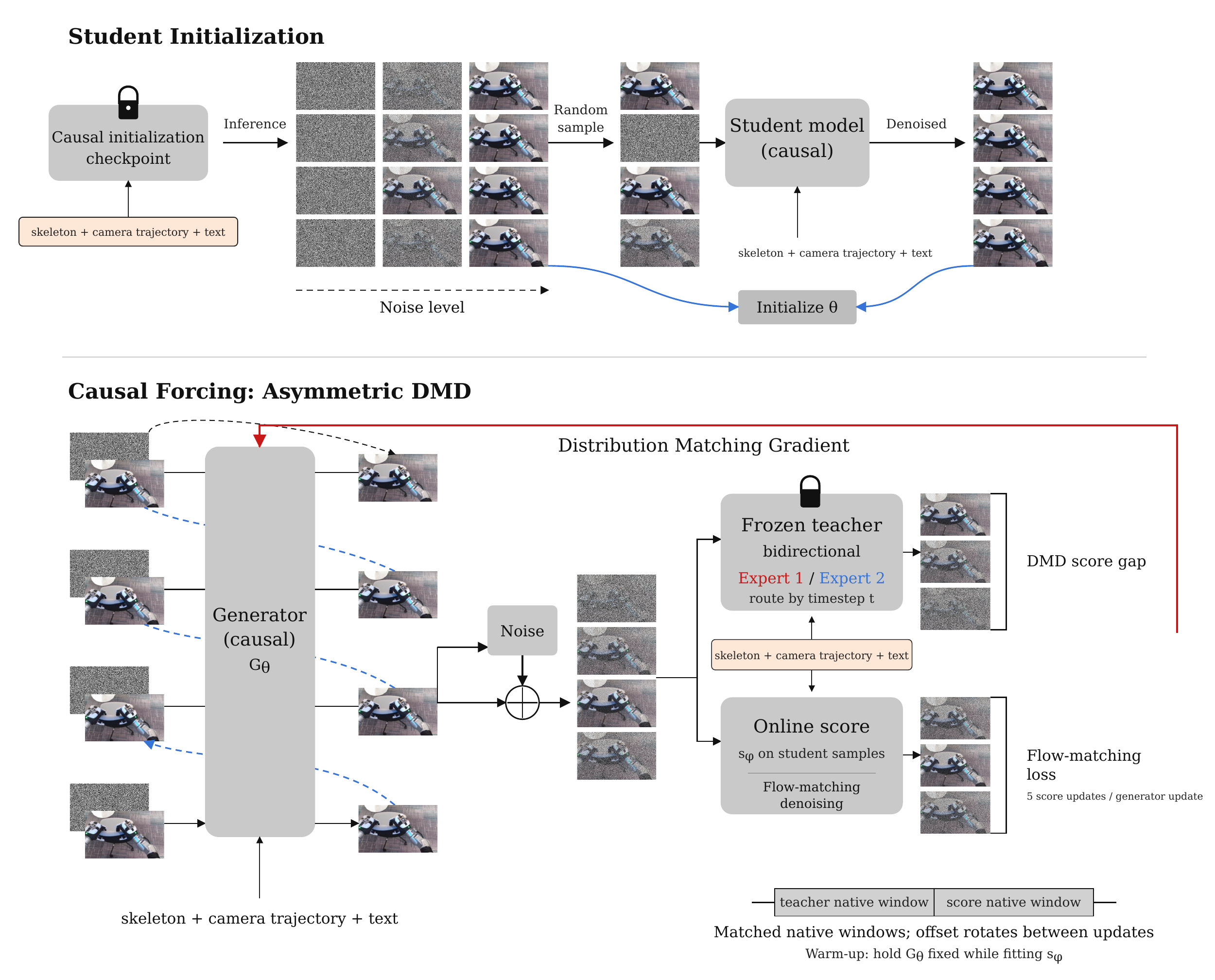}
  \caption{\textbf{Distillation of \method{} into a few-step causal student.}
  Top: causal distillation initializes the few-step autoregressive generator.
  Bottom: asymmetric DMD updates the causal generator $G_\theta$ using a frozen bidirectional real-score teacher and an online-trained bidirectional fake-score model. The diagram illustrates the A14B MoE backbone; the 5B model follows an analogous procedure with a dense backbone. LongLive subsequently extends the few-step model for long-video streaming (Fig.~\ref{fig:interactive_rollout}).}
  \label{fig:distillation_training}
\end{figure}

\begin{figure}[!htbp]
  \centering
  \includegraphics[width=\linewidth]{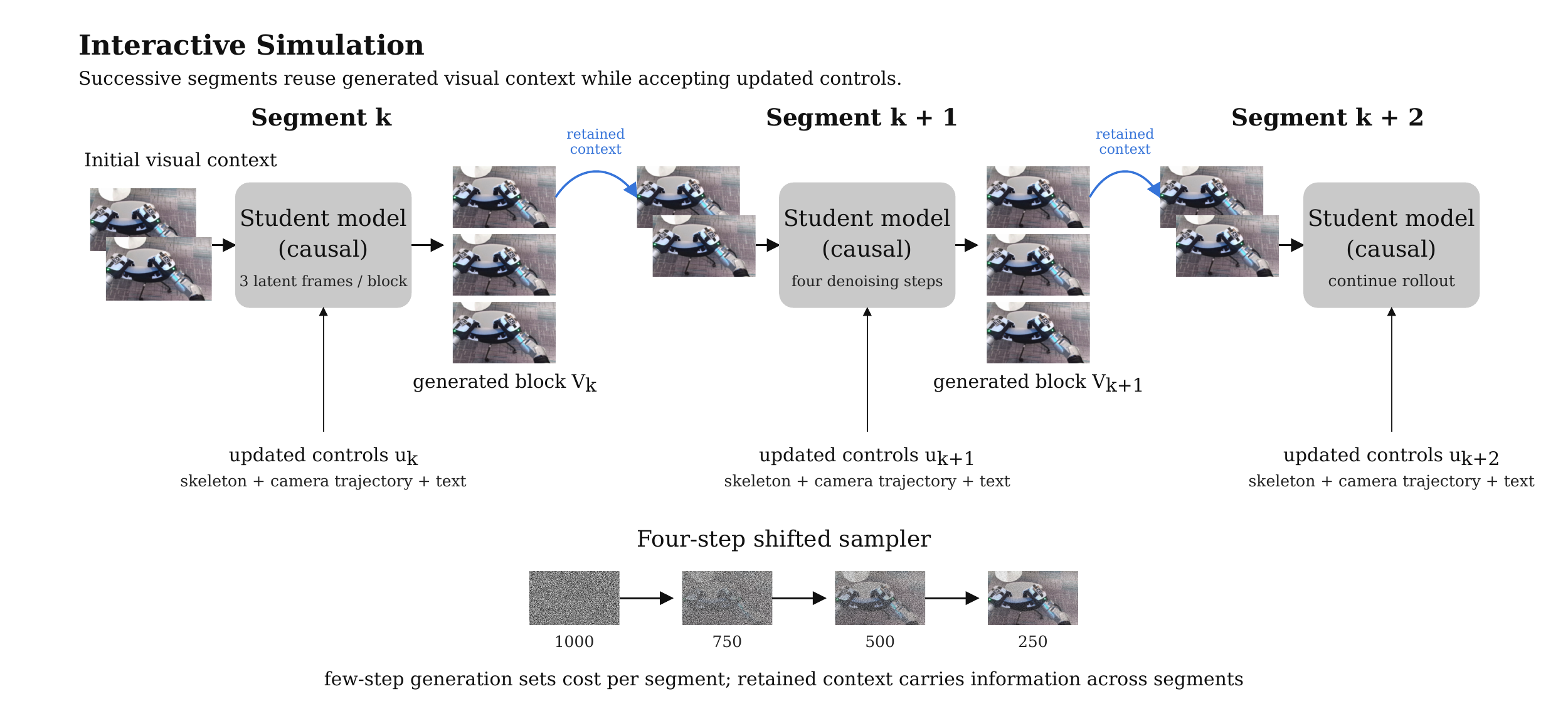}
  \caption{\textbf{LongLive-based streaming with the distilled student.}
  Successive segments reuse the previously generated visual context while accepting updated controls (skeleton, camera trajectory, and text). Each block is produced by the few-step causal student, so the per-segment cost is bounded while retained context carries information across segments, enabling real-time long-horizon rollout.}
  \label{fig:interactive_rollout}
\end{figure}

Figure~\ref{fig:distillation_training} illustrates distillation; Fig.~\ref{fig:interactive_rollout} shows subsequent long-video streaming.

\FloatBarrier

\paragraph{Interactive inference and throughput.}
On a single GPU at $640\times480$, the 4-step student generates a 125-frame clip in about 2\,s of autoregressive generation and about 6\,s including VAE decoding ($\approx$21\,fps), compared with about 41\,s for the 50-step teacher (a $7\times$ end-to-end speed-up), and streams its first block within roughly 1.4\,s.

Figure~\ref{fig:interactive_rollout} illustrates the streaming rollout mechanism; Table~\ref{tab:distillation} reports quality and throughput across rollout lengths.

\begin{table}[!htbp]
  \centering
  \caption{\textbf{Distillation quality--speed comparison} (GR1, 5B). The 4-step student matches or exceeds the 50-step teacher at every rollout length (bold) while running $7\times$ faster.}
  \label{tab:distillation}
  \begingroup
  \papertablestyle
  \setlength{\tabcolsep}{6pt}
  \renewcommand{\arraystretch}{1.0}
  \begin{tabular*}{\linewidth}{@{\extracolsep{\fill}}ll cccc@{}}
    \toprule
    Model & Rollout & FPS$\uparrow$ & PSNR$\uparrow$ & SSIM$\uparrow$ & LPIPS$\downarrow$ \\
    \midrule
    \multirow{4}{*}{Teacher (50-step)} & 6.3\,s & \multirow{4}{*}{3.0} & 15.95 & 0.528 & 0.248 \\
      & 19\,s & & 14.34 & 0.454 & 0.321 \\
      & 32\,s & & 14.13 & 0.438 & 0.348 \\
      & 61\,s & & 13.64 & 0.414 & 0.385 \\
    \cmidrule(lr){1-6}
    \multirow{4}{*}{Student (4-step)} & 6.3\,s & \multirow{4}{*}{\textbf{21.3}} & \rankone{16.12} & \rankone{0.538} & \rankone{0.247} \\
      & 19\,s & & \rankone{14.70} & \rankone{0.468} & \rankone{0.312} \\
      & 32\,s & & \rankone{14.44} & \rankone{0.451} & \rankone{0.338} \\
      & 61\,s & & \rankone{13.98} & \rankone{0.428} & \rankone{0.373} \\
    \bottomrule
  \end{tabular*}
  \endgroup
\end{table}

\FloatBarrier

\FloatBarrier
\clearpage
\subsection{Wide-FOV Skeleton Conditioning}
\label{app:widefov}

Head cameras often crop the manipulator. On the WorldArena~\citep{shang2026worldarena} (RoboTwin~2.0) test set, the narrow skeleton touches the image border in 99.4\% of frames; a $1.5\times$ field-of-view render raises in-frame manipulator coverage from 0.37 to 0.56. The optional wide-FOV branch (Sec.~\ref{sec:method-architecture}) feeds this render through per-block cross-attention, providing arm configuration that is absent from the narrow view.

Figures~\ref{fig:widefov} and~\ref{fig:widefov2} compare predictions without and with this branch against ground truth when the arm works near or beyond the frame edge. Columns identify the variants; the yellow box marks the head-camera field of view within the $1.5\times$ canvas. Both examples report per-episode border-band keypoint error.

\newcommand{\widefovhead}{\makebox[\linewidth]{\small
  \makebox[0.333\linewidth][c]{w/o wide-FOV}%
  \makebox[0.333\linewidth][c]{with wide-FOV}%
  \makebox[0.333\linewidth][c]{ground truth}}}

\begin{figure}[!htbp]
  \centering
  \begin{minipage}{0.88\linewidth}
  \widefovhead\\[1.5pt]
  \includegraphics[width=\linewidth]{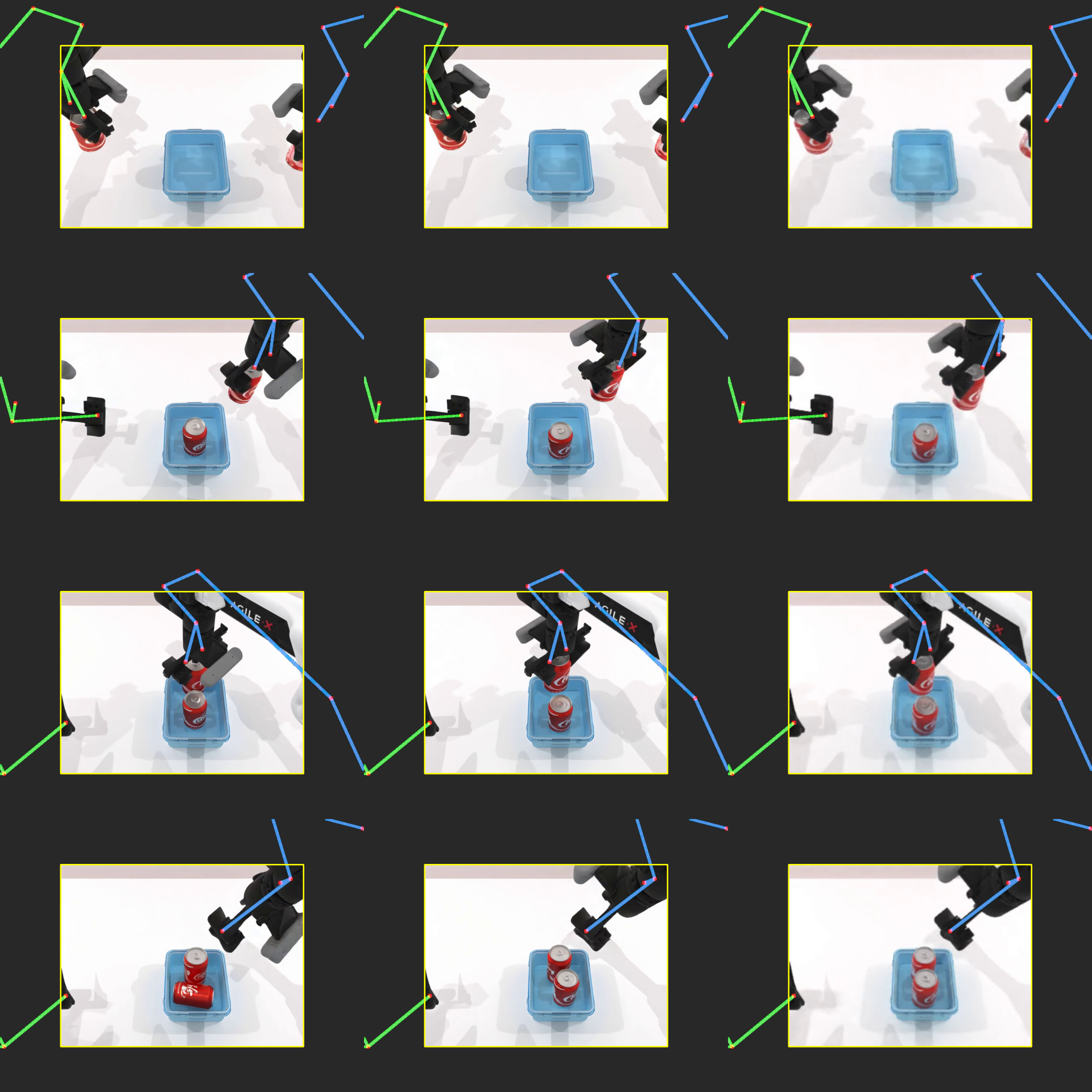}
  \end{minipage}
  \caption{\textbf{Wide-FOV skeleton branch, example~1} (four moments in time order; columns labeled above). Wide-FOV skeleton conditioning improves the alignment of the visible arm near the image boundary. Here the outer border-band keypoint error drops from 8.9 to 6.9\,px, and the wide branch is closer to the ground truth on 91\% of frames.}
  \label{fig:widefov}
\end{figure}

\begin{figure}[!htbp]
  \centering
  \widefovhead\\[1.5pt]
  \includegraphics[width=\linewidth]{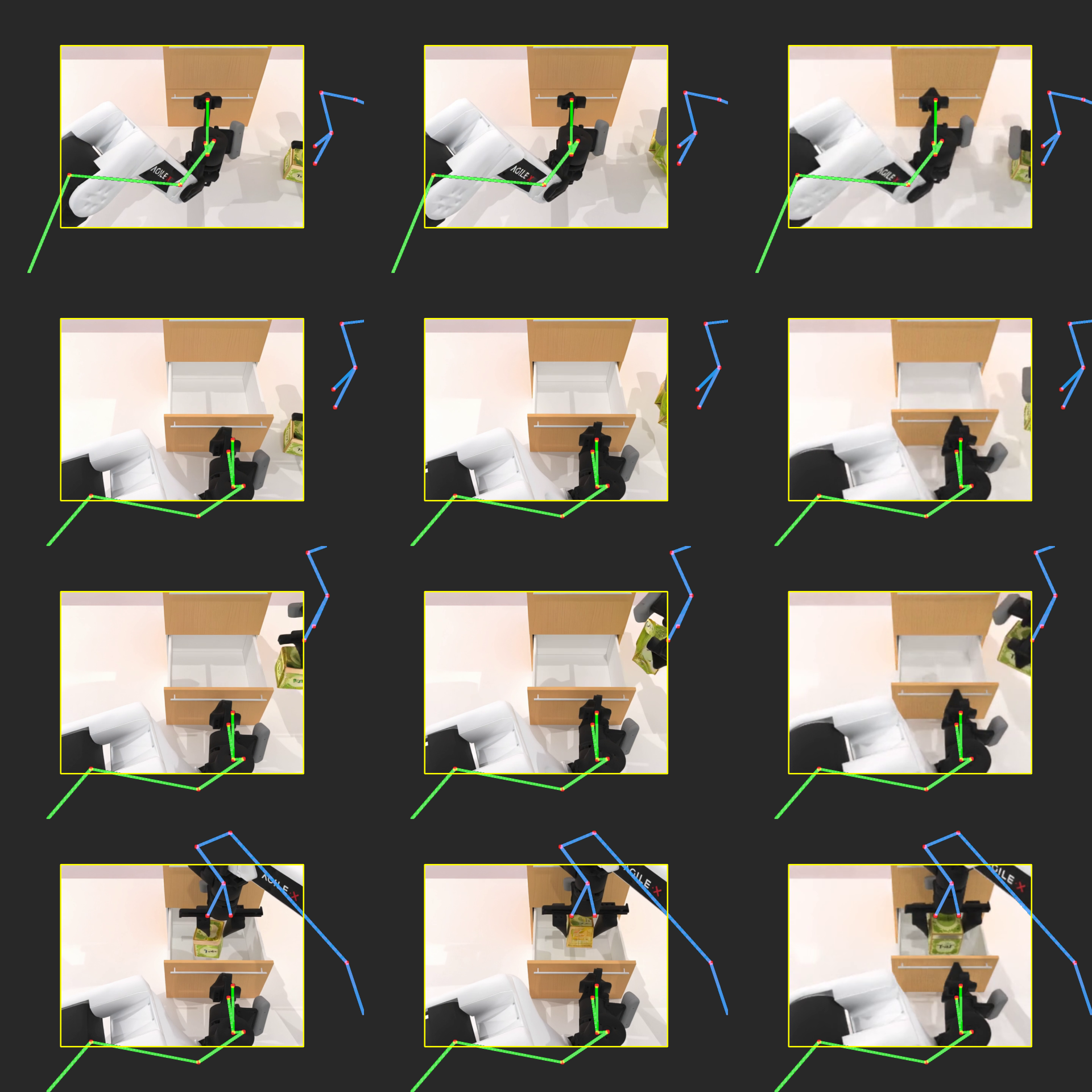}
  \caption{\textbf{Wide-FOV skeleton branch, example~2} (four moments in time order; columns labeled above). When the arm extends beyond the camera view, wide-FOV skeleton conditioning improves the prediction of its visible portion. Here the border-band keypoint error drops from 14.0 to 11.7\,px, with the wide branch closer to the ground truth on 93\% of frames.}
  \label{fig:widefov2}
\end{figure}

\FloatBarrier

\clearpage
\section{Evaluation Protocols}
\label{app:evaluation_protocols}

We specify the human-video evaluation procedure and the protocol used for the scaling curves. Supplementary benchmark results are collected in Sec.~\ref{app:full_results}.

\subsection{Human Egocentric Video Evaluation}
\label{app:ego200}

The pooled 200-clip results (100 per dataset from EgoSuite~\citep{lightwheel2026egosuite} and OpenAoE~\citep{li2026open}) are reported in the main text (Table~\ref{tab:human_eval}). Here we specify the evaluation-set composition, protocol, and metric definitions; the per-dataset breakdown is reported in Sec.~\ref{app:human_results}.

\paragraph{Evaluation set.}
We evaluate on 100 clips per dataset (200 total), sampled across high, medium, and low camera-motion strata.

\paragraph{Evaluation protocol.}
Clips contain 81 frames at $640\times480$ and 16\,fps. Methods receive the same initial frame and, when supported, the same caption, with seed 0. Camera and hand conditions derive from the same camera annotations and projected hand keypoints, converted to each method's native interface. Each baseline is run under its native conditioning interface: DreamDojo with latent actions, Cosmos3 with joint camera-and-hand conditioning.

\paragraph{Metrics.}
ATE is absolute trajectory error after rigid SE(3) alignment without scale fitting; RPE measures stepwise translation error, and RRE measures rotation error relative to the first frame. Camera poses are estimated from generated videos with DA3~\citep{depthanything3}. Hand metrics use MediaPipe~\citep{mediapipe} detections at confidence 0.3, matched to projected annotations. PCK@.1 uses a threshold of 10\% of the image diagonal and penalizes missed detections; PCKm@.1 evaluates matched hands only, and Det is the detection rate. Cons., Dyn, and Aes denote VBench subject consistency, dynamic degree, and aesthetic quality. FVD$_{16}$ uses non-overlapping 16-frame windows, while FVD$_{\mathrm{full}}$ uses the full sequences. FVD is recomputed for each evaluation set rather than averaged across datasets.

In the human evaluation tables, ATE and RPE are reported in centimeters, RRE in degrees, HandL2 in pixels, and PSNR in dB. PCK, PCKm, Det, Cons., Dyn, and Aes are reported as percentages; SSIM, LPIPS, and FVD have no physical units.

\subsection{Scaling Protocol}
\label{app:scaling}

The scaling figures appear in the main text: held-out human video at hand-detection confidence 0.3 (Fig.~\ref{fig:scaling_human200_conf03}), human-to-robot transfer to $\Psi_0$/\teleopname{}/$\omega$-HOME (Fig.~\ref{fig:scaling_ours3}), and the DreamDojo cross-evaluation sets (Fig.~\ref{fig:scaling_gr1}). We detail the shared protocol here. Each point is a checkpoint taken at the indicated amount of human-video pre-training data (hours of video processed), with the data mixture held fixed. Because training stays within a single epoch, the hours processed at each point equal the amount of \emph{unique} human data seen, so these curves measure data scaling rather than repeated exposure. For held-out human video the checkpoint is evaluated zero-shot, whereas for the DreamDojo and transfer benchmarks it is post-trained on the target with the same recipe before evaluation.

For transfer scaling, post-training uses one clip for 50 steps on $\Psi_0$ and 144 clips for five epochs on \teleopname{}.

\clearpage
\section{Additional Quantitative Results}
\label{app:full_results}

The main benchmark summaries appear in Tables~\ref{tab:main_benchmarks}, \ref{tab:agibot_benchmarks}, and~\ref{tab:human_eval}. Here we provide per-dataset human results, seen/unseen robot breakdowns, long-horizon comparisons, ablations, and the WorldArena comparison against BWM. Distillation quality and throughput are reported alongside the implementation in Sec.~\ref{app:distillation}.

\subsection{Human Evaluation by Dataset}
\label{app:human_results}

Table~\ref{tab:ego200_by_dataset} separates the pooled human-video results in Table~\ref{tab:human_eval} into EgoSuite and OpenAoE. The evaluation protocol and metric units are specified in Sec.~\ref{app:ego200}.
\begin{table}[!htbp]
  \centering
  \caption{\textbf{Per-dataset results on EgoSuite~\citep{lightwheel2026egosuite} and OpenAoE~\citep{li2026open}.} Each set contains 100 clips from Table~\ref{tab:ego200_overall}. ATE is in cm and PSNR in dB. Best and second-best bold and underlined.}
  \label{tab:ego200_by_dataset}
  \begingroup
  \papertablestyle
  \renewcommand{\arraystretch}{1.04}
  \newcommand{\egomodelsep}{\noalign{\vskip 2pt\hbox to\linewidth{\xleaders\hbox{\rule{2pt}{0.3pt}\hskip 2pt}\hfill}\vskip 2pt}}
  \begin{tabular*}{\linewidth}{@{\extracolsep{\fill}}lcccccc@{}}
    \toprule
    \multirow{2}{*}{Method} & \multicolumn{3}{c}{EgoSuite (100)} & \multicolumn{3}{c}{OpenAoE (100)} \\
    \cmidrule(lr){2-4}\cmidrule(lr){5-7}
    & ATE$\downarrow$ & PSNR$\uparrow$ & FVD$_{16}$$\downarrow$
    & ATE$\downarrow$ & PSNR$\uparrow$ & FVD$_{16}$$\downarrow$ \\
    \midrule
    Cosmos3-Nano (16B)  & 1.122 & 19.75 & 111.5 & 2.740 & \rankthree{17.56} & 139.9 \\
    Cosmos3-Super (64B)  & 1.322 & 19.55 & 110.4 & 2.740 & 17.16 & \ranktwo{132.7} \\
    OSCAR (2B)  & 2.586 & 16.87 & 173.7 & 4.374 & 15.54 & 221.8 \\
    RynnWorld-Teleop (5B)  & 2.642 & 17.98 & 182.4 & 4.444 & 15.90 & 200.3 \\
    DreamDojo (2B)  & 1.892 & 17.94 & 173.8 & 3.412 & 15.57 & 241.1 \\
    DreamDojo (14B)  & 1.890 & 18.95 & 178.8 & 3.414 & 16.56 & 207.9 \\
    \egomodelsep
    \method{} (5B), pretrain  & \rankthree{1.012} & \rankthree{20.45} & \rankthree{109.8} & \rankthree{2.574} & 17.35 & 140.0 \\
    \method{} (5B), midtrain  & 1.038 & 20.38 & \ranktwo{103.0} & 2.729 & 17.43 & 139.4 \\
    \method{} (A14B), pretrain  & \rankone{0.872} & \rankone{21.48} & \rankone{94.8} & \rankone{2.212} & \ranktwo{18.07} & \rankone{125.8} \\
    \method{} (A14B), midtrain  & \ranktwo{0.988} & \ranktwo{20.51} & 114.4 & \ranktwo{2.511} & \rankone{18.17} & \rankthree{135.8} \\
    \bottomrule
  \end{tabular*}
  \endgroup
\end{table}

\FloatBarrier

\subsection{Seen and Unseen Robot Evaluation}
\label{app:seen_unseen_results}

Table~\ref{tab:seen_unseen} breaks down the main dexterous-humanoid results (Table~\ref{tab:main_results}) into seen and unseen subsets for $\Psi_0$, $\omega$-HOME, and \teleopname{}, under no post-training, few-shot, and full-data adaptation.
\begin{table}[!htbp]
  \centering
  \caption{\textbf{Seen/unseen breakdown on $\Psi_0$~\citep{wei2026psi0}, $\omega$-HOME~\citep{li2026omega0}, and \teleopname{} (PSNR$\uparrow$).} $\Delta$ is the seen$-$unseen gap. Best bold, second-best underlined.}
  \label{tab:seen_unseen}\label{tab:psi0_seen_unseen}\label{tab:omega_home_seen_unseen}\label{tab:teleop_seen_unseen}
  \begingroup
  \newcommand{\susep}{\noalign{\vskip 1.5pt\hbox to\linewidth{\xleaders\hbox{\rule{2pt}{0.3pt}\hskip 2pt}\hfill}\vskip 1.5pt}}
  \papertablestyle
  \renewcommand{\arraystretch}{0.95}
  \begin{tabular*}{\linewidth}{@{\extracolsep{\fill}}lccc@{\hspace{6pt}}ccc@{\hspace{6pt}}ccc@{}}
    \toprule
    \multirow{2}{*}{Model / initialization} & \multicolumn{3}{c}{No post-training} & \multicolumn{3}{c}{Few-shot} & \multicolumn{3}{c}{Full-data} \\
    \cmidrule(lr){2-4}\cmidrule(lr){5-7}\cmidrule(lr){8-10}
    & Seen & Unseen & $\Delta$
    & Seen & Unseen & $\Delta$
    & Seen & Unseen & $\Delta$ \\
    \midrule
    \multicolumn{10}{@{}l}{\textbf{$\Psi_0$}} \\
    \addlinespace[1pt]
    DreamDojo (14B)  & 14.90 & 13.99 & 0.91  & 14.85 & 14.02 & 0.83  & 14.13 & 13.43 & 0.70  \\
    Cosmos3-Nano (16B)  & 13.83 & 13.65 & 0.18  & 15.04 & 14.37 & 0.67  & \rankthree{17.10} & 16.00 & 1.10  \\
    Cosmos3-Super (64B)  & 13.82 & 13.73 & 0.09  & 15.27 & 14.60 & 0.67  & 16.23 & 15.22 & 1.01  \\
    \susep
    \method{} (5B), pretrain  & 15.34 & \rankthree{15.01} & 0.33  & \rankthree{15.78} & 15.39 & 0.39  & 16.67 & 15.85 & 0.82  \\
    \method{} (5B), midtrain  & \rankthree{15.45} & 14.99 & 0.46  & 15.75 & \rankthree{15.44} & 0.31  & 16.68 & \rankthree{16.13} & 0.55  \\
    \method{} (A14B), pretrain  & \ranktwo{15.96} & \ranktwo{15.78} & 0.18  & \ranktwo{17.90} & \ranktwo{17.77} & 0.13  & \ranktwo{19.24} & \ranktwo{18.64} & 0.60  \\
    \method{} (A14B), midtrain  & \rankone{17.39} & \rankone{17.33} & 0.06  & \rankone{18.58} & \rankone{17.98} & 0.60  & \rankone{19.25} & \rankone{18.90} & 0.35  \\
    \midrule
    \multicolumn{10}{@{}l}{\textbf{$\omega$-HOME}} \\
    \addlinespace[1pt]
    DreamDojo (14B)  & 15.41 & 13.98 & 1.43  & 15.14 & 13.74 & 1.40  & 17.16 & 15.17 & 1.99  \\
    Cosmos3-Nano (16B)  & 16.79 & 15.69 & 1.10  & 17.88 & 16.60 & 1.28  & 19.79 & 17.89 & 1.90  \\
    Cosmos3-Super (64B)  & 16.45 & 15.35 & 1.10  & 17.46 & 16.22 & 1.24  & 19.25 & 17.67 & 1.58  \\
    \susep
    \method{} (5B), pretrain  & 18.18 & 16.52 & 1.66  & 18.87 & \rankthree{17.83} & 1.04  & 19.79 & 18.26 & 1.53  \\
    \method{} (5B), midtrain  & \rankthree{18.52} & \rankthree{17.13} & 1.39  & \rankthree{18.89} & 17.71 & 1.18  & \rankthree{20.00} & \rankthree{18.50} & 1.50  \\
    \method{} (A14B), pretrain  & \ranktwo{18.87} & \ranktwo{17.93} & 0.94  & \ranktwo{21.06} & \ranktwo{19.81} & 1.25  & \ranktwo{22.97} & \ranktwo{20.01} & 2.96  \\
    \method{} (A14B), midtrain  & \rankone{19.90} & \rankone{19.21} & 0.69  & \rankone{21.50} & \rankone{19.92} & 1.58  & \rankone{23.29} & \rankone{20.25} & 3.04  \\
    \midrule
    \multicolumn{10}{@{}l}{\textbf{\teleopname{}}} \\
    \addlinespace[1pt]
    DreamDojo (14B)  & 14.60 & 12.39 & 2.21  & 13.98 & 12.01 & 1.97  & 13.93 & 11.90 & 2.03  \\
    Cosmos3-Nano (16B)  & 14.72 & 12.66 & 2.06  & 15.14 & 13.89 & 1.25  & 15.74 & 14.22 & 1.52  \\
    Cosmos3-Super (64B)  & 14.90 & 12.80 & 2.10  & 15.15 & 13.64 & 1.51  & 15.63 & 13.99 & 1.64  \\
    \susep
    \method{} (5B), pretrain  & 15.40 & 13.64 & 1.76  & \rankthree{16.37} & 14.63 & 1.74  & \rankthree{16.76} & \rankthree{15.05} & 1.71  \\
    \method{} (5B), midtrain  & \rankthree{15.84} & \ranktwo{14.11} & 1.73  & \rankthree{16.37} & \rankthree{14.68} & 1.69  & 16.75 & 15.03 & 1.72  \\
    \method{} (A14B), pretrain  & \ranktwo{16.09} & \rankthree{13.80} & 2.29  & \ranktwo{18.85} & \ranktwo{17.11} & 1.74  & \ranktwo{19.18} & \ranktwo{17.34} & 1.84  \\
    \method{} (A14B), midtrain  & \rankone{17.25} & \rankone{15.59} & 1.66  & \rankone{19.15} & \rankone{17.35} & 1.80  & \rankone{19.61} & \rankone{17.87} & 1.74  \\
    \bottomrule
  \end{tabular*}
  \endgroup
\end{table}

\FloatBarrier

\subsection{Long-Horizon Rollout Evaluation}
\label{app:long_horizon_results}

Tables~\ref{tab:omega_home_long} and~\ref{tab:teleop_long30s} compare autoregressive rollouts against the baselines: approximately 32\,s zero-shot on $\omega$-HOME and 30\,s on \teleopname{}. The tables report aggregate quality and how prediction quality changes over the rollout.
\begin{table}[!htbp]
  \centering
  \caption{\textbf{Long-horizon rollout on $\omega$-HOME~\citep{li2026omega0}} ($\approx$32\,s, zero-shot; 12 chained windows). Overall metrics are frame-averaged; per-segment PSNR at $\approx$1/14/31\,s shows degradation over the rollout. Best \textbf{bold}, second-best \underline{underlined}.}
  \label{tab:omega_home_long}
  \begingroup
  \newcommand{\olmodelsep}{\noalign{\vskip 2pt\hbox to\linewidth{\xleaders\hbox{\rule{2pt}{0.3pt}\hskip 2pt}\hfill}\vskip 2pt}}
  \papertablestyle
  \begin{tabular*}{\linewidth}{@{\extracolsep{\fill}}lccc@{\hspace{7pt}}ccc@{}}
    \toprule
    \multirow{2}{*}{Model / initialization} & \multicolumn{3}{c}{Overall} & \multicolumn{3}{c}{Per-segment PSNR$\uparrow$} \\
    \cmidrule(lr){2-4}\cmidrule(lr){5-7}
    & PSNR$\uparrow$ & SSIM$\uparrow$ & LPIPS$\downarrow$ & seg.\,1 & seg.\,6 & seg.\,12 \\
    \midrule
    DreamDojo (14B)  & \rankthree{14.17} & 0.641 & 0.579 & 22.32  & 13.36  & 13.02  \\
    Cosmos3-Nano (16B)  & 13.91 & 0.573 & 0.576 & 25.57  & 12.75  & 11.11  \\
    Cosmos3-Super (64B)  & 13.74 & 0.579 & 0.570 & 25.13  & 12.58  & 11.14  \\
    \olmodelsep
    \method{} (5B), pretrain  & 13.64 & 0.651 & 0.516 & 23.92  & 12.57  & 10.35  \\
    \method{} (5B), midtrain  & 13.93 & \rankthree{0.664} & \rankthree{0.506} & 26.33  & 12.74  & 10.43  \\
    \method{} (A14B), pretrain  & \ranktwo{15.20} & \ranktwo{0.696} & \ranktwo{0.478} & 25.03  & 14.03  & 12.40  \\
    \method{} (A14B), midtrain  & \rankone{16.48} & \rankone{0.738} & \rankone{0.376} & 29.33  & 15.20  & 12.35  \\
    \bottomrule
  \end{tabular*}
  \endgroup
\end{table}

\begin{table}[!htbp]
  \centering
  \caption{\textbf{Long-horizon rollout on \teleopname{}} ($\approx$30\,s; 11 autoregressively chained windows, each restarting from the model's own last frame). Best \textbf{bold}, second-best \underline{underlined}; Dyn.\ is best when closest to the ground-truth value (0.70).}
  \label{tab:teleop_long30s}
  \begingroup
  \newcommand{\longmodelsep}{\noalign{\vskip 2pt\hbox to\linewidth{\xleaders\hbox{\rule{2pt}{0.3pt}\hskip 2pt}\hfill}\vskip 2pt}}
  \papertablestyle
  \begin{tabular*}{\linewidth}{@{\extracolsep{\fill}}lccc@{\hspace{7pt}}ccc@{\hspace{7pt}}ccc@{}}
    \toprule
    \multirow{2}{*}{Model / initialization} & \multicolumn{3}{c}{Hand following} & \multicolumn{3}{c}{Fidelity} & \multicolumn{3}{c}{Video quality} \\
    \cmidrule(lr){2-4}\cmidrule(lr){5-7}\cmidrule(lr){8-10}
    & Err.\,(px)$\downarrow$ & Miss$\downarrow$ & Err.@120$\downarrow$
    & PSNR$\uparrow$ & SSIM$\uparrow$ & LPIPS$\downarrow$
    & B-Cons. & Aes.$\uparrow$ & Dyn. \\
    \midrule
    DreamDojo (14B)  & 62.6 & 77.1\% & 106.9 & 10.00 & 0.301 & 0.620 & \rankthree{0.891} & 0.449 & 0.55  \\
    Cosmos3-Nano (16B)  & 55.8 & \rankthree{50.9\%} & \rankthree{88.5} & \ranktwo{11.10} & \ranktwo{0.338} & \rankthree{0.579} & \ranktwo{0.892} & \ranktwo{0.500} & 0.55  \\
    Cosmos3-Super (64B)  & \rankthree{52.5} & 61.5\% & 94.0 & 10.76 & 0.302 & 0.587 & 0.884 & \rankthree{0.478} & 0.50  \\
    \longmodelsep
    \method{} (5B), midtrain  & \ranktwo{19.8} & \ranktwo{45.3\%} & \ranktwo{65.2} & \rankthree{10.83} & \rankthree{0.336} & \ranktwo{0.573} & \ranktwo{0.892} & 0.469 & 0.70  \\
    \method{} (A14B), midtrain  & \rankone{12.0} & \rankone{6.9\%} & \rankone{19.5} & \rankone{13.47} & \rankone{0.433} & \rankone{0.444} & \rankone{0.906} & \rankone{0.510} & 0.70  \\
    \bottomrule
  \end{tabular*}
  \endgroup
\end{table}

\FloatBarrier

\subsection{Conditioning and Pretraining Ablations}
\label{app:ablation_results}

Table~\ref{tab:ablation_main} evaluates the camera and skeleton conditioning channels on the cross-embodiment set. Table~\ref{tab:pretrain_ablation} compares our pretrained initialization with public video-generation weights at matched post-training data budgets. Qualitative examples of the initialization comparison are grouped in Sec.~\ref{app:qual_posttraining}.
\begin{table}[!htbp]
  \centering
  \caption{\textbf{Cross-embodiment set} (400 evaluation windows over eight embodiments). Cosmos3~\citep{nvidia2026cosmos} is fine-tuned on the same robot mixture used for \method{}'s mid-training; neither model is post-trained on the evaluation windows. Each ablation variant is trained without one conditioning channel. Best bold, second-best underlined.}
  \label{tab:ablation_main}\label{tab:baseline_adaptation}
  \vspace{2pt}
  \begingroup
  \papertablestyle
  \setlength{\tabcolsep}{6pt}
  \renewcommand{\arraystretch}{0.95}
  \begin{tabular}{ll ccc}
    \toprule
    Model & Variant & PSNR$\uparrow$ & SSIM$\uparrow$ & LPIPS$\downarrow$ \\
    \midrule
    \multirow{2}{*}{Cosmos3}
       & Nano (16B)  & 19.02 & 0.739 & 0.197 \\
       & Super (64B)  & 18.77 & 0.736 & 0.202 \\
    \cmidrule(lr){1-5}
    \multirow{3}{*}{Ours 5B}
       & Full  & \rankthree{19.83} & \rankthree{0.760} & \rankthree{0.167} \\
       & w/o Pl\"ucker  & 19.71 & 0.755 & 0.178 \\
       & w/o skeletons  & 18.03 & 0.736 & 0.214 \\
    \cmidrule(lr){1-5}
    \multirow{3}{*}{Ours A14B}
       & Full  & \rankone{21.98} & \rankone{0.802} & \rankone{0.128} \\
       & w/o Pl\"ucker  & \ranktwo{21.65} & \ranktwo{0.791} & \ranktwo{0.145} \\
       & w/o skeletons  & 18.17 & 0.744 & 0.208 \\
    \bottomrule
  \end{tabular}
  \endgroup
\end{table}

\begin{table}[!htbp]
  \centering
  \caption{\textbf{Effect of pretraining.} Our pretrained base vs.\ the same architecture from public video-generation weights (``official base''), post-trained on 1\%/10\%/100\% of the GR-1 data; best bold.}
  \label{tab:pretrain_ablation}
  \vspace{2pt}
  \begingroup
  \papertablestyle
  \renewcommand{\arraystretch}{0.95}
  \begin{tabular*}{\linewidth}{@{\extracolsep{\fill}}ll ccc@{\hspace{6pt}}ccc@{\hspace{6pt}}ccc@{}}
    \toprule
    \multirow{2}{*}{Data} & \multirow{2}{*}{Base} & \multicolumn{3}{c}{In-lab} & \multicolumn{3}{c}{EgoDex} & \multicolumn{3}{c}{DreamDojo-HV} \\
    \cmidrule(lr){3-5}\cmidrule(lr){6-8}\cmidrule(lr){9-11}
    & & PSNR$\uparrow$ & SSIM$\uparrow$ & LPIPS$\downarrow$
    & PSNR$\uparrow$ & SSIM$\uparrow$ & LPIPS$\downarrow$
    & PSNR$\uparrow$ & SSIM$\uparrow$ & LPIPS$\downarrow$ \\
    \midrule
    \multirow{2}{*}{1\%}
       & Official base  & 16.04 & 0.495 & 0.446 & 13.42 & 0.394 & 0.548 & 12.43 & 0.348 & 0.548 \\
       & \method{}      & \rankone{20.85} & \rankone{0.764} & \rankone{0.213} & \rankone{20.57} & \rankone{0.778} & \rankone{0.208} & \rankone{19.05} & \rankone{0.760} & \rankone{0.217} \\
    \cmidrule(lr){1-11}
    \multirow{2}{*}{10\%}
       & Official base  & 18.89 & 0.630 & 0.287 & 18.64 & 0.687 & 0.267 & 16.95 & 0.623 & 0.307 \\
       & \method{}      & \rankone{21.02} & \rankone{0.786} & \rankone{0.206} & \rankone{20.78} & \rankone{0.801} & \rankone{0.199} & \rankone{19.29} & \rankone{0.787} & \rankone{0.207} \\
    \cmidrule(lr){1-11}
    \multirow{2}{*}{100\%}
       & Official base  & 19.41 & 0.650 & 0.275 & 18.87 & 0.695 & 0.265 & 17.48 & 0.645 & 0.293 \\
       & \method{}      & \rankone{21.74} & \rankone{0.797} & \rankone{0.202} & \rankone{21.23} & \rankone{0.810} & \rankone{0.196} & \rankone{19.81} & \rankone{0.795} & \rankone{0.204} \\
    \bottomrule
  \end{tabular*}
  \endgroup
\end{table}

\FloatBarrier

\clearpage
\subsection{WorldArena Comparison against BWM}
\label{app:worldarena}

We compare \method{} (A14B) with the Boundless World Model (BWM)~\citep{bwm2026} on WorldArena~\citep{shang2026worldarena} Track~1 (RoboTwin~2.0~\citep{chen2025robotwin}, \texttt{fixed\_scene\_task} test set, 1000 episodes). Both models use their official code and public checkpoints, the same official evaluation pipeline at $640\times480$, and the same Real-ESRGAN~\citep{wang2021real} super-resolved first frame (``BWM\,+\,SR''). We report our original-length run.

\method{} leads on 12 of the 15 metrics and by $+5.9$ EWMScore over BWM\,+\,SR (Table~\ref{tab:worldarena}). The margin is dominated by \textbf{Trajectory Accuracy} ($+29.1$: the arm being where the commanded action says it should be), followed by background/subject consistency and instruction following. Fig.~\ref{fig:worldarena_qual} shows this qualitatively.

\begin{table}[!htbp]
  \centering
  \caption{\textbf{WorldArena~\citep{shang2026worldarena} Track~1 comparison against BWM~\citep{bwm2026}} (RoboTwin~2.0~\citep{chen2025robotwin}, 1000 episodes; identical official pipeline and inputs for both). EWMScore is the mean of the 15 metrics; best in bold.}
  \label{tab:worldarena}
  \vspace{2pt}
  \begingroup
  \papertablestyle
  \renewcommand{\arraystretch}{0.95}
  \begin{tabular*}{\linewidth}{@{\extracolsep{\fill}}l c c@{}}
    \toprule
    Metric & BWM\,+\,SR & \method{} (A14B) \\
    \midrule
    \textbf{EWMScore}          & 61.01 & \rankone{66.92} \\
    \midrule
    Image Quality              & 51.54 & \rankone{53.57} \\
    Aesthetic Quality          & \rankone{43.09} & 41.04 \\
    JEPA Similarity            & 90.93 & \rankone{96.00} \\
    Dynamic Degree             & 17.49 & \rankone{25.48} \\
    Flow Score                 & 8.49  & \rankone{15.75} \\
    Motion Smoothness          & 66.13 & \rankone{67.55} \\
    Subject Consistency        & 68.08 & \rankone{79.51} \\
    Background Consistency     & 71.78 & \rankone{84.17} \\
    Photometric Consistency    & \rankone{35.95} & 27.98 \\
    Interaction Quality        & 70.22 & \rankone{77.42} \\
    Trajectory Accuracy        & 34.64 & \rankone{63.78} \\
    Depth Accuracy             & 94.71 & \rankone{98.56} \\
    Perspectivity              & 96.04 & \rankone{97.72} \\
    Instruction Following      & 76.22 & \rankone{85.50} \\
    Semantic Alignment         & \rankone{89.79} & 89.73 \\
    \bottomrule
  \end{tabular*}
  \endgroup
\end{table}

\newcommand{\wahead}{\makebox[0.49\linewidth]{\small
  \makebox[0.163\linewidth][c]{GT}%
  \makebox[0.163\linewidth][c]{Ours}%
  \makebox[0.163\linewidth][c]{BWM}}}
\begin{figure}[!htbp]
  \centering
  \wahead\hfill\wahead\\[1.5pt]
  \includegraphics[width=0.49\linewidth]{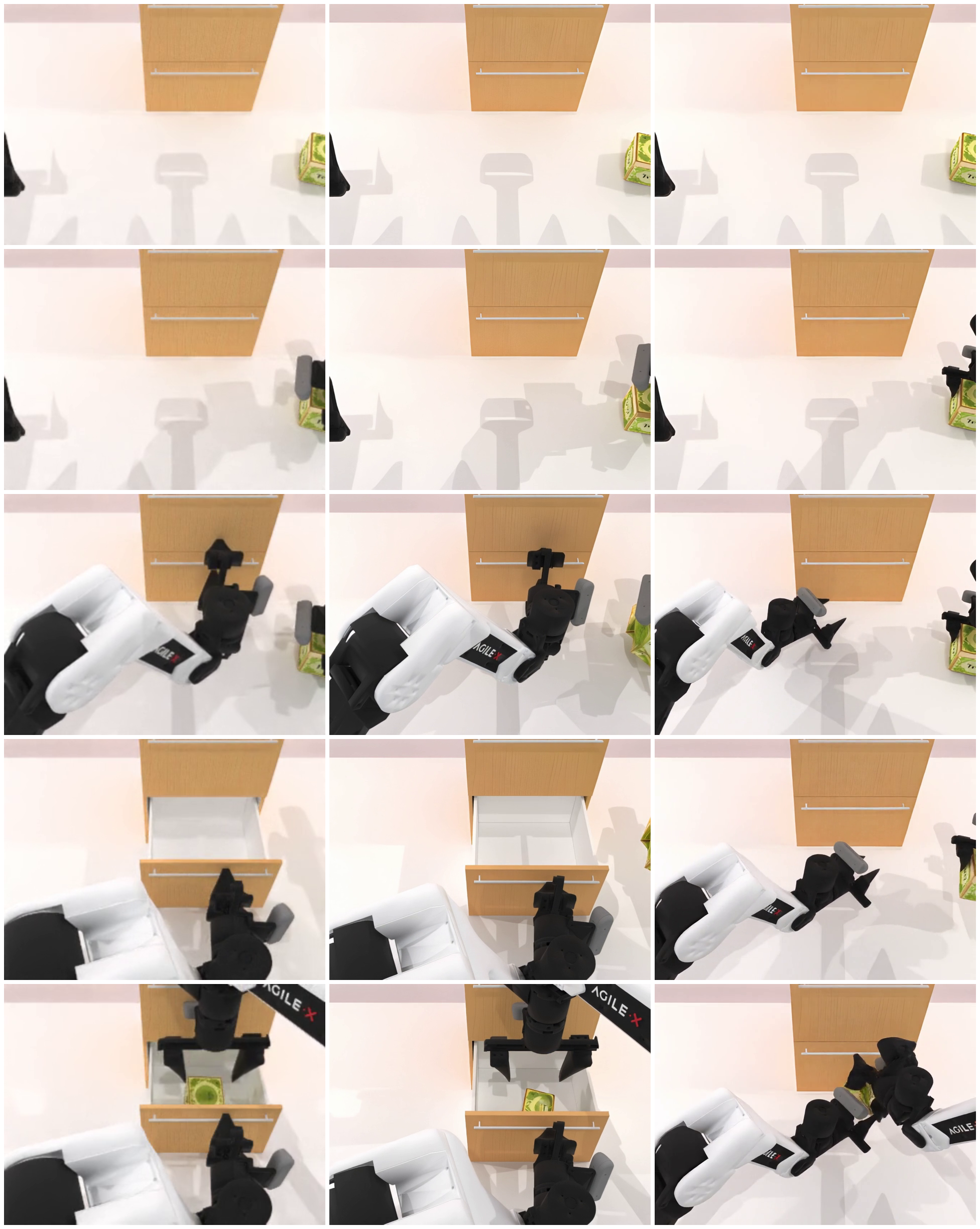}\hfill
  \includegraphics[width=0.49\linewidth]{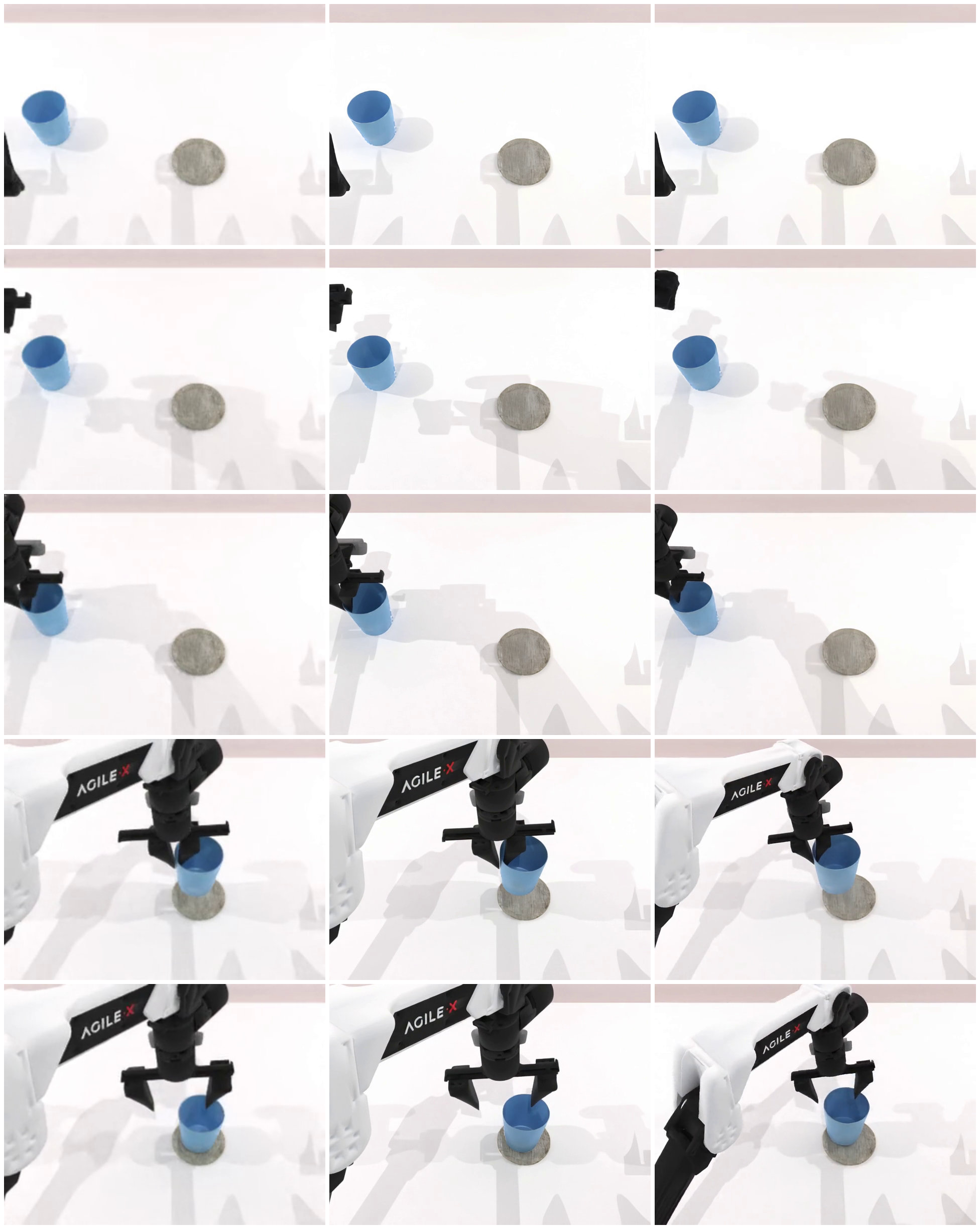}
  \caption{\textbf{Qualitative comparison on WorldArena} (left: drawer\,+\,box; right: cup onto coaster). Each panel: five chronological frames (rows); columns are \emph{GT} $\vert$ \emph{\method{} (A14B)} $\vert$ \emph{BWM} (Table~\ref{tab:worldarena}). \method{} follows the commanded arm trajectory and completes the task like the ground truth (opening the drawer and placing the box, precisely seating the cup), whereas BWM's arm drifts and under-moves (it never opens the drawer).}
  \label{fig:worldarena_qual}
\end{figure}

\FloatBarrier

\FloatBarrier
\section{Qualitative Comparisons}
\label{app:qualitative}

\label{app:additional_visualizations}

We group qualitative comparisons by evaluation setting. Compared methods receive the same initial frame and conditioning through their native interfaces. Ground-truth and predicted frames are aligned in time; figures from the \href{https://ropedia.github.io/ropedia-w0/}{project-page videos} use the same displayed frame indices across methods.

\subsection{Cross-Embodiment Robot Prediction}
\label{app:qual_robot}
\label{app:additional_robot_comparisons}

\begin{figure}[!htbp]
  \centering
  \includegraphics[width=0.88\linewidth]{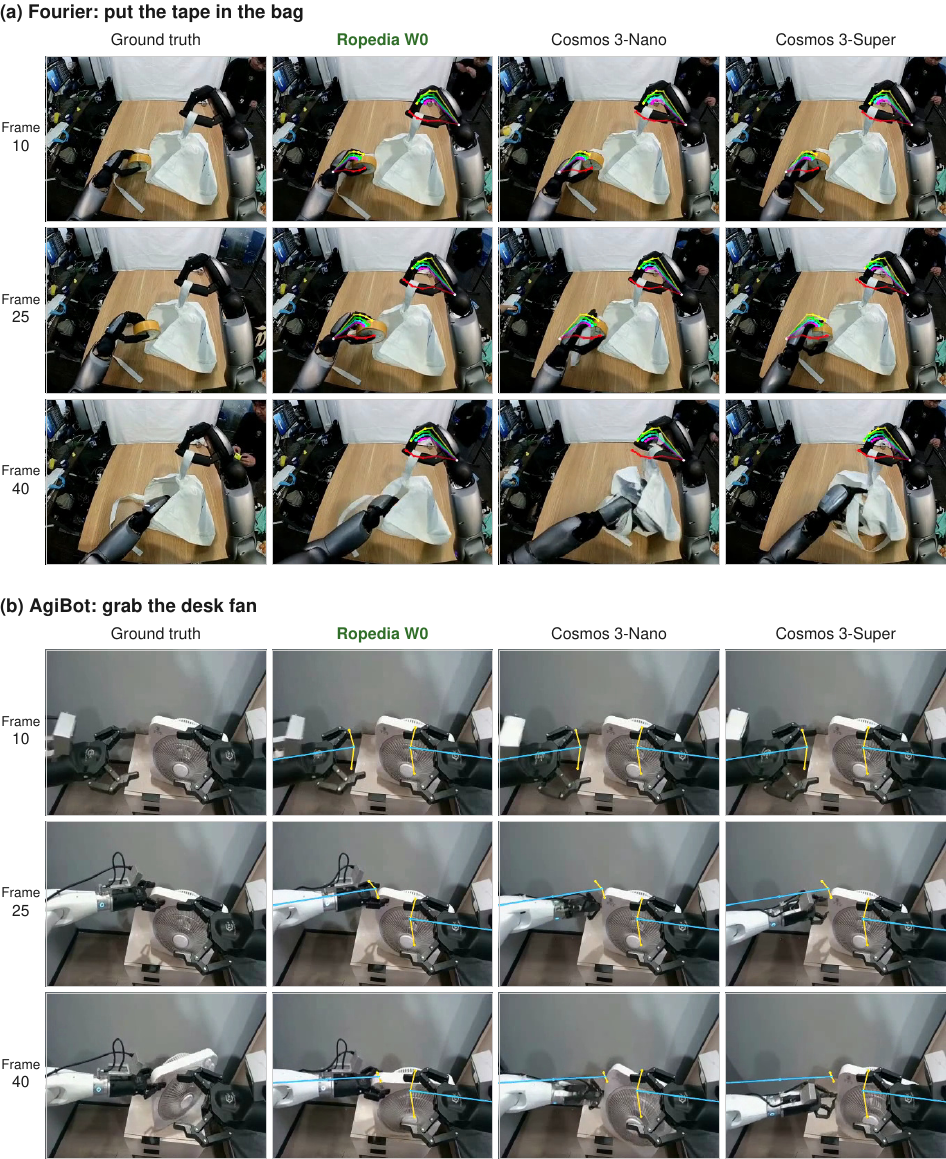}
  \caption{\textbf{Robot-dataset comparisons (1/3).}
  (a) Fourier: putting tape in a bag; (b) AgiBot: grasping a desk fan.
  These are held-out clips: \method{} uses mid-training and Cosmos3 uses
  fine-tuning on the robot mixture. Colored skeletons are supplied input
  conditions overlaid on the generated frames.}
  \label{fig:additional_robot_1}
\end{figure}

\begin{figure}[!htbp]
  \centering
  \includegraphics[width=0.94\linewidth]{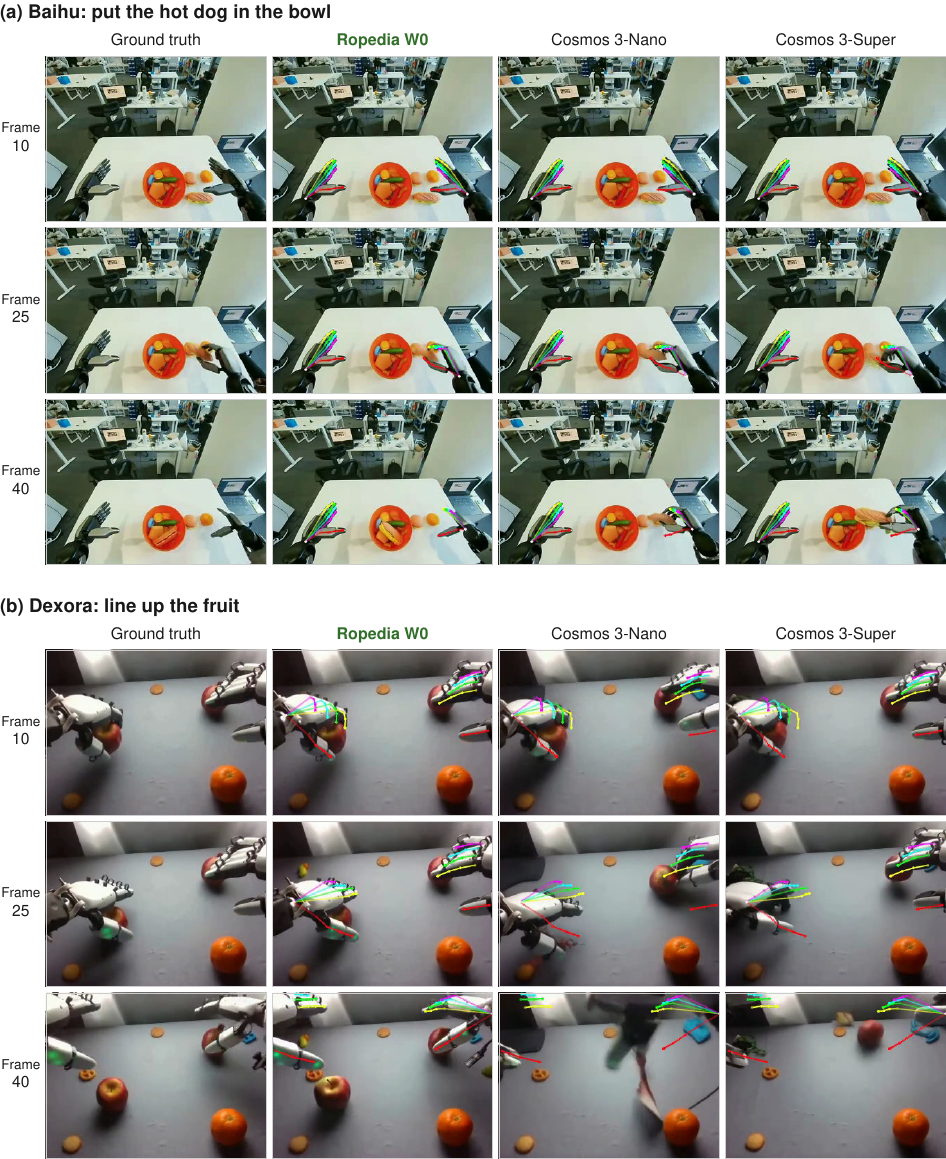}
  \caption{\textbf{Robot-dataset comparisons (2/3).}
  (a) Baihu: placing a hot dog in a bowl; (b) Dexora: arranging fruit.
  Columns compare the ground truth, \method{}, Cosmos3-Nano, and
  Cosmos3-Super at aligned frames. The held-out-clip setting and input
  skeleton overlays follow Fig.~\ref{fig:additional_robot_1}.}
  \label{fig:additional_robot_2}
\end{figure}

\begin{figure}[!htbp]
  \centering
  \includegraphics[width=0.94\linewidth]{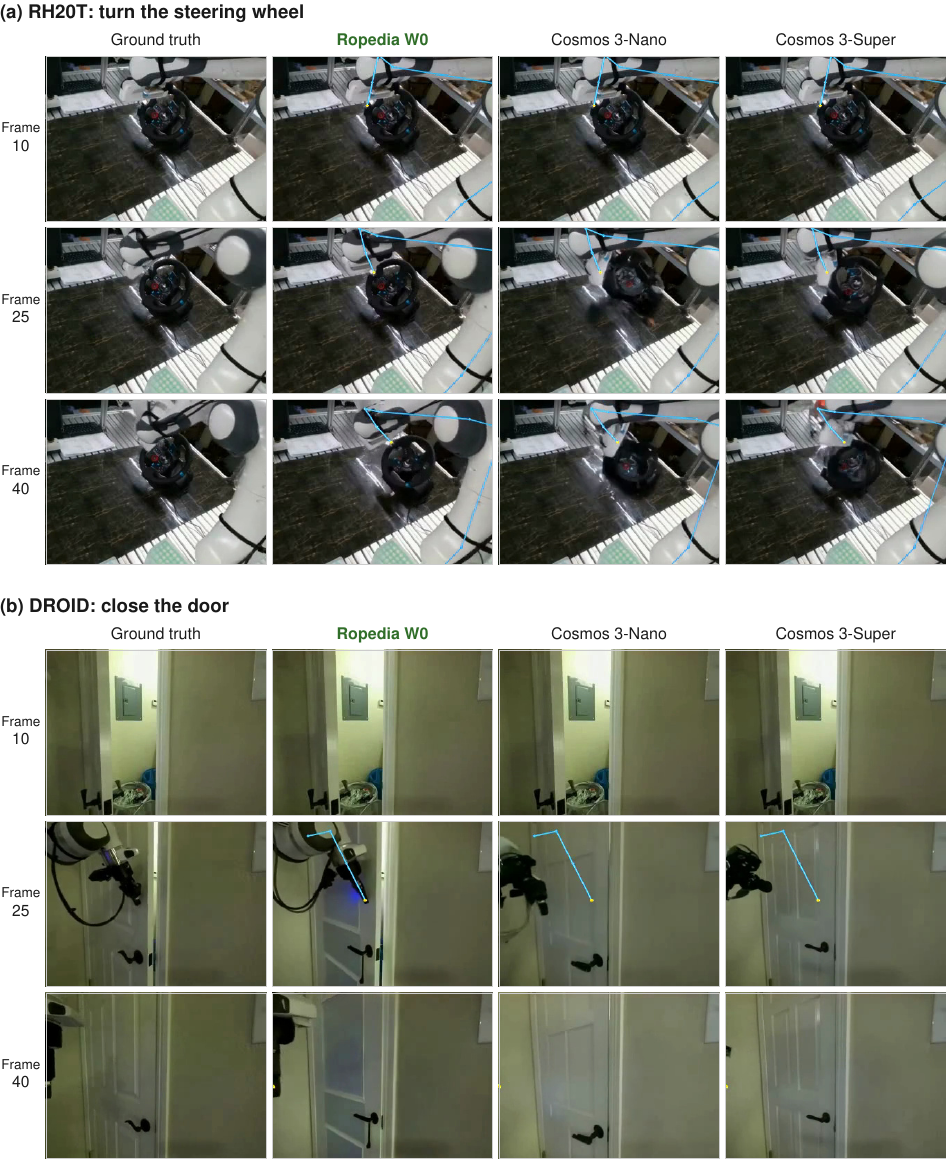}
  \caption{\textbf{Robot-dataset comparisons (3/3).}
  (a) RH20T: turning a steering wheel; (b) DROID: closing a door.
  These examples extend the comparison to gripper motion and interactions
  with articulated objects. The held-out-clip setting and input skeleton
  overlays follow Fig.~\ref{fig:additional_robot_1}.}
  \label{fig:additional_robot_3}
\end{figure}

Figures~\ref{fig:additional_robot_1}--\ref{fig:additional_robot_3} compare predictions on held-out clips from Fourier, AgiBot, Baihu, Dexora, RH20T, and DROID. These examples cover different manipulators, objects, and interaction types; the captions specify the adaptation setting.

\FloatBarrier

\subsection{Zero-Shot Loco-Manipulation}
\label{app:qual_locomotion}
\label{app:additional_omega_comparisons}

\begin{figure}[!htbp]
  \centering
  \includegraphics[width=0.94\linewidth]{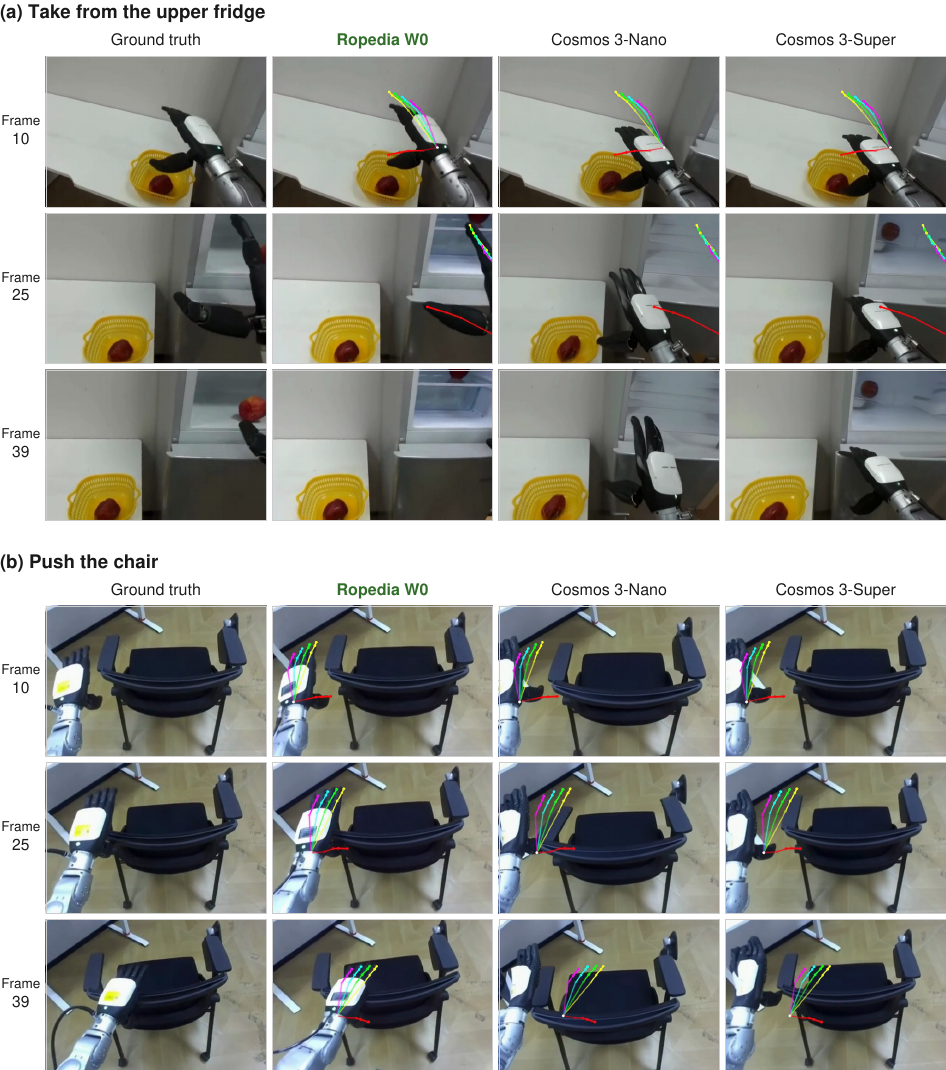}
  \caption{\textbf{Zero-shot comparisons on $\omega$-HOME (1/2).}
  (a) Reaching into the upper fridge; (b) pushing a chair.
  The compared models have not been trained on $\omega$-HOME.
  Rows show aligned frames from the same clip; colored skeletons are input
  conditions overlaid on each generated video.}
  \label{fig:additional_omega_1}
\end{figure}

\begin{figure}[!htbp]
  \centering
  \includegraphics[width=0.94\linewidth]{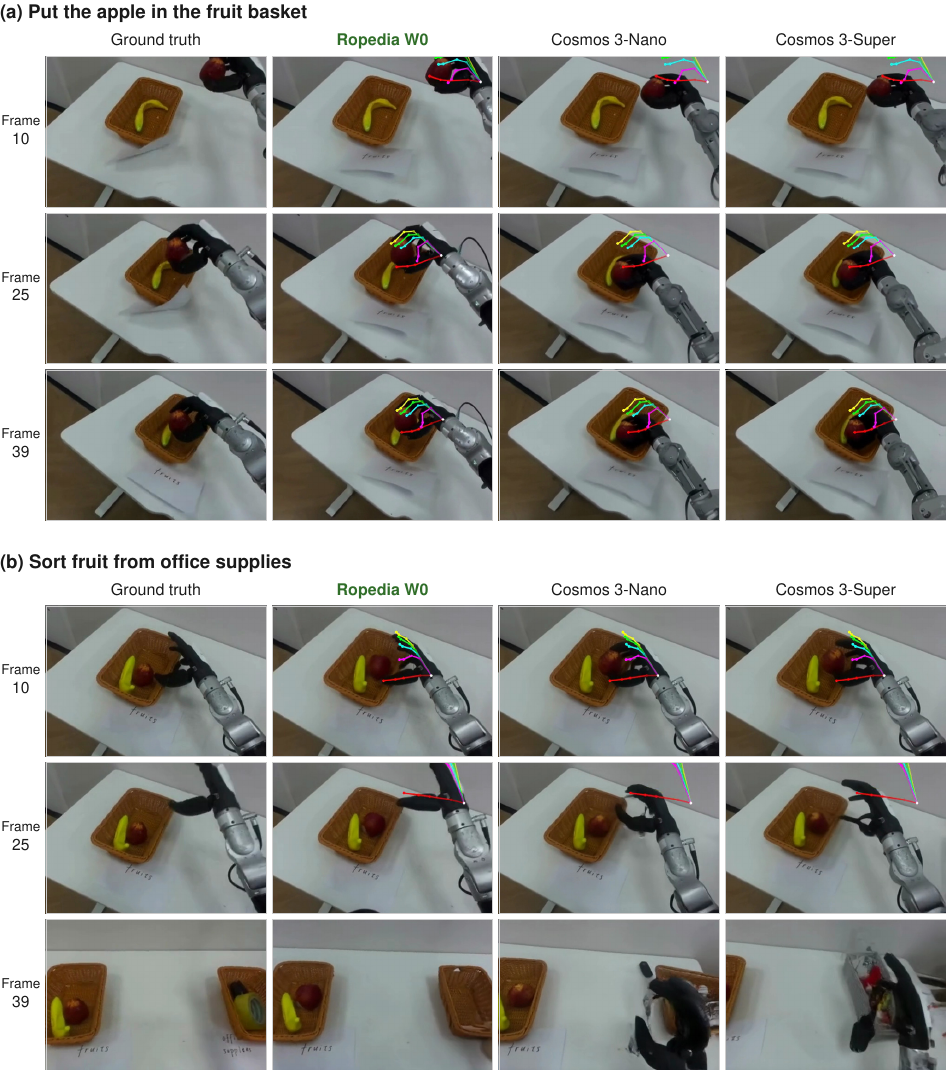}
  \caption{\textbf{Zero-shot comparisons on $\omega$-HOME (2/2).}
  (a) Putting an apple in a fruit basket; (b) sorting fruit from office supplies.
  Ground-truth frames provide a reference for the generated hand and object
  configurations. The setting and input skeleton overlays follow
  Fig.~\ref{fig:additional_omega_1}.}
  \label{fig:additional_omega_2}
\end{figure}

Figures~\ref{fig:additional_omega_1} and~\ref{fig:additional_omega_2} compare zero-shot predictions on household tasks from $\omega$-HOME~\citep{li2026omega0}, a held-out loco-manipulation dataset not used in training by the compared models. The examples show generated hand motion, object interactions, and scene appearance against time-aligned ground-truth frames.

\FloatBarrier

\subsection{Post-Training and Data Efficiency}
\label{app:qual_posttraining}
\label{app:additional_posttraining}

\begin{figure}[!htbp]
  \centering
  \includegraphics[width=\linewidth]{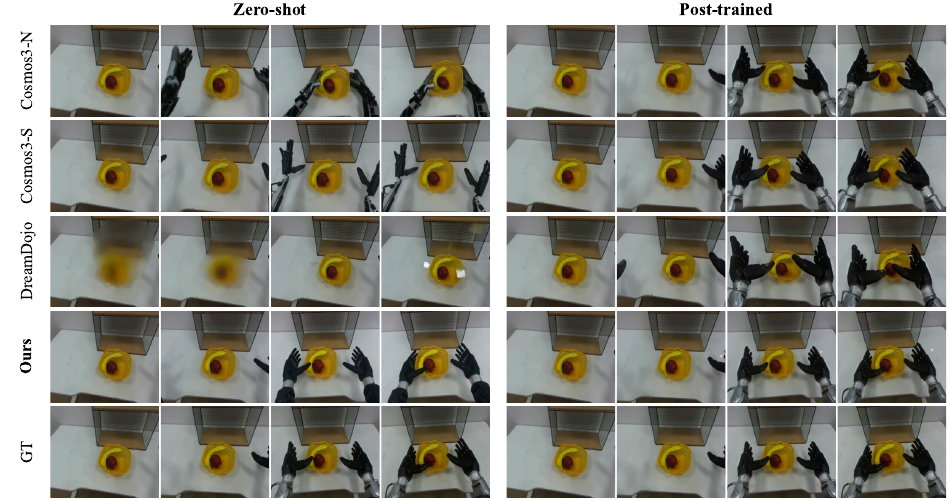}
  \caption{\textbf{$\omega$-HOME~\citep{li2026omega0} manipulation task} (fruit-bucket arrangement). Left: zero-shot; right: after post-training. \method{} follows the commanded hands and matches the ground truth in both settings, whereas the baselines lose the arm or drift.}
  \label{fig:qual_omega_seen}
\end{figure}

\paragraph{Adaptation on $\omega$-HOME.}
Figure~\ref{fig:qual_omega_seen} compares zero-shot and post-trained predictions against the baselines on a manipulation task.

\FloatBarrier

\begin{figure}[!htbp]
  \centering
  \includegraphics[width=\linewidth]{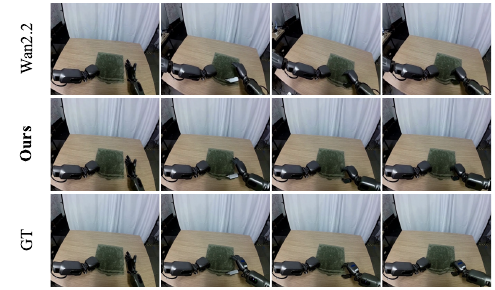}
  \caption{\textbf{Few-shot data efficiency on GR-1 (In-lab).} Both models are post-trained on 1\% of the data; \method{} initializes from our pretrained base and Wan2.2-I2V-A14B~\citep{wan2025} from public video-generation weights. Our base yields markedly sharper, more faithful predictions.}
  \label{fig:qual_gr1_inlab}
\end{figure}

\begin{figure}[!htbp]
  \centering
  \includegraphics[width=\linewidth]{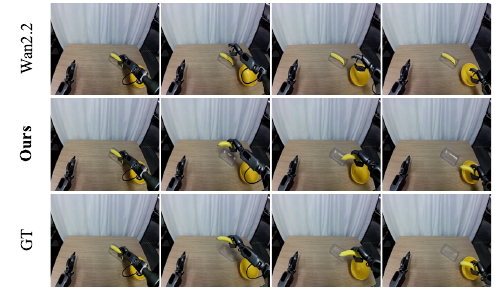}
  \caption{\textbf{Few-shot data efficiency on GR-1 (EgoDex~\citep{hoque2025egodex}).} Same 1\% budget; initializing from our pretrained base preserves object and hand detail that the public base loses.}
  \label{fig:qual_gr1_egodex}
\end{figure}

\paragraph{Pretrained initialization under limited data.}
We probe data efficiency by post-training on only 1\% of the GR-1 data from two initializations: our pretrained base versus public video-generation weights (Wan2.2-I2V-A14B). Across both the In-lab (Fig.~\ref{fig:qual_gr1_inlab}) and EgoDex (Fig.~\ref{fig:qual_gr1_egodex}) settings, initializing from our pretrained base yields markedly sharper predictions and preserves object and hand detail that the public base loses.

\FloatBarrier

\begin{figure}[!htbp]
  \centering
  \includegraphics[width=0.94\linewidth]{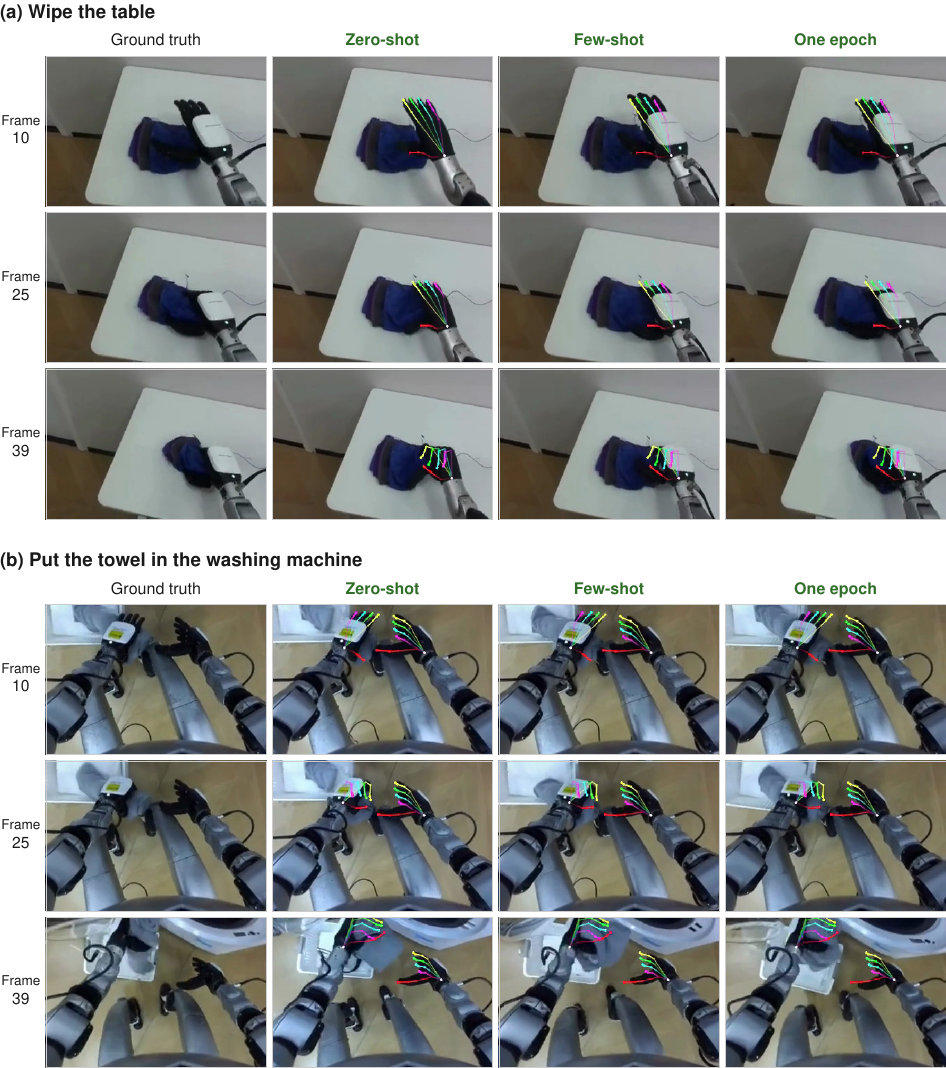}
  \caption{\textbf{Post-training examples on $\omega$-HOME.}
  (a) Wiping a table; (b) putting a towel in a washing machine.
  Columns show ground truth and \method{} under the three settings labeled
  in the project-page demonstrations: zero-shot, few-shot, and one epoch
  of target-data post-training. Colored skeletons are overlaid input
  conditions. Each row uses the same frame index across settings.}
  \label{fig:additional_posttraining}
\end{figure}

\paragraph{Post-training data budgets.}
Figure~\ref{fig:additional_posttraining} shows the same model under zero-shot, few-shot, and one-epoch target-data post-training on two further $\omega$-HOME tasks.

\FloatBarrier

\clearpage
\subsection{Additional Long-Horizon Rollouts}
\label{app:additional_long}

\begin{figure}[!htbp]
  \centering
  \includegraphics[width=0.94\linewidth]{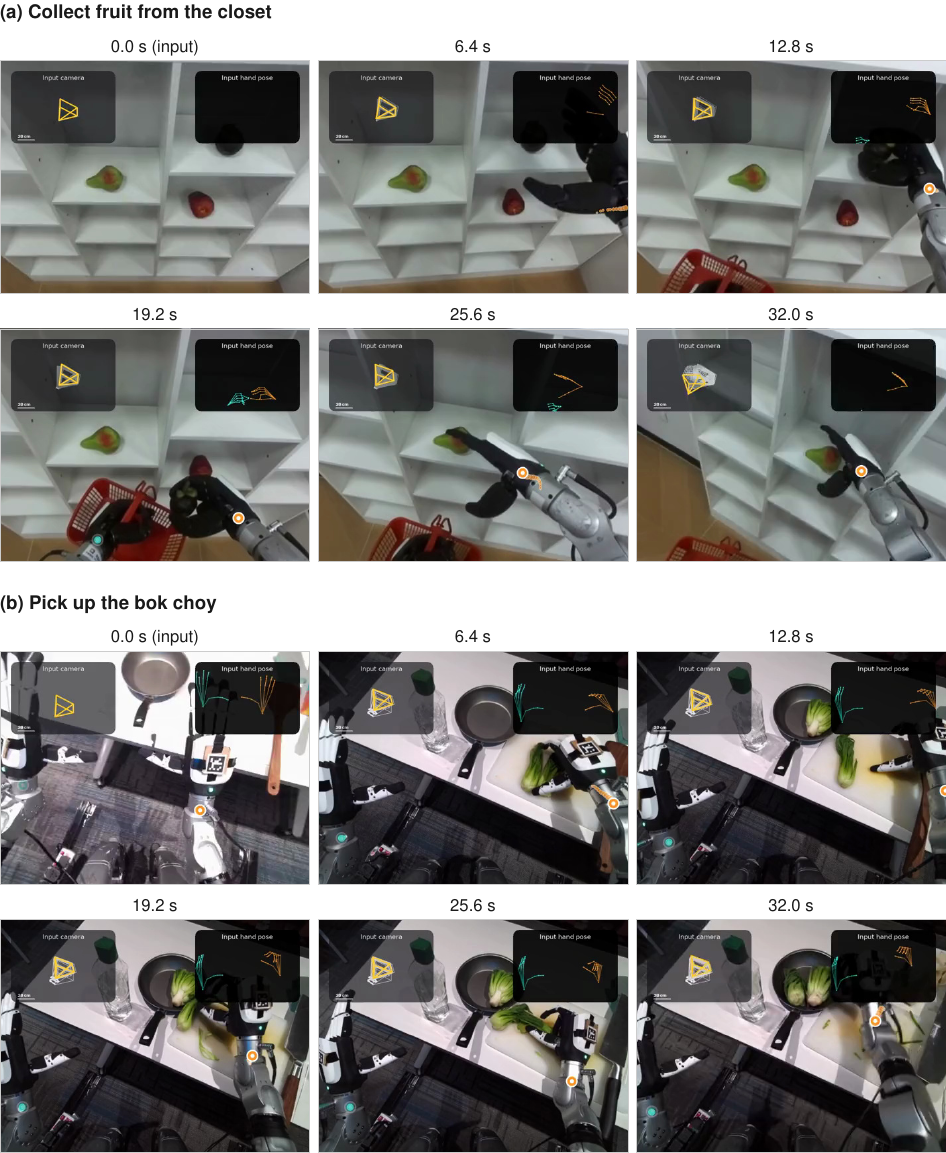}
  \caption{\textbf{Additional 32-second generated rollouts.}
  (a) Collecting fruit from a closet; (b) picking up bok choy.
  Each sequence starts with the input image at 0\,s, followed by generated
  frames sampled every 6.4\,s, read left to right and then top to bottom.
  The camera and hand-pose insets display supplied conditions.}
  \label{fig:additional_long}
\end{figure}

Figure~\ref{fig:additional_long} shows two additional 32-second generated sequences with their supplied camera and hand conditions. These examples visualize rollout continuity; the quantitative long-horizon comparisons are reported in Sec.~\ref{app:long_horizon_results}.

\FloatBarrier

\clearpage
\section{Limitations and Future Work}
\label{sec:limitations}
\label{app:limitations}

\begin{enumerate}
  \item The distilled student supports real-time, minute-long rollouts (Sec.~\ref{app:distillation}); stronger memory mechanisms or periodic re-grounding are promising directions for extending stable interactive simulation to longer horizons.
  \item We study up to a 14B-active model and scale the amount of human-video pre-training data while holding the data mixture fixed; prediction quality improves monotonically with the amount of data (Sec.~\ref{app:scaling}), suggesting that scaling data and model size further, and broadening embodiment and task coverage, are promising directions toward more capable and reliable embodied world models.
  \item Our kinematic interface captures camera and manipulator motion but not physical contact or object dynamics; coupling the world model with simulation-ready, physically grounded 3D assets and scenes~\citep{cao2026physx3d,cao2026physxanything,cao2026physxomni} is a promising direction toward modeling the underlying dynamics rather than only their visual consequences.
\end{enumerate}

\end{document}